\pdfoutput=1
\documentclass[11pt]{article}
\usepackage[preprint]{acl}
\usepackage{array}
\usepackage{booktabs}
\usepackage[T1]{fontenc}
\usepackage{graphicx}
\usepackage{inconsolata}
\usepackage[utf8]{inputenc}
\usepackage{latexsym}
\usepackage{longtable}
\usepackage{makecell}
\usepackage{microtype}
\usepackage{pifont}
\usepackage{ragged2e}
\usepackage{subcaption}
\usepackage{tabularx}
\usepackage{tcolorbox}
\usepackage{times}
\usepackage{xcolor}
\newcommand{\xmark}{\textcolor{red}{\ding{55}}}
\newcommand{\cmark}{\textcolor{green}{\ding{51}}}
\title{EconWebArena: Benchmarking Autonomous Agents on Economic Tasks in Realistic Web Environments}
\author{%
    Zefang Liu\thanks{These authors contributed equally to this work.} \\
    Capital One \\
    San Jose, CA, USA \\
    \texttt{zefang.liu@capitalone.com} \\
    \And
    Yinzhu Quan$^*$ \\
    Georgia Institute of Technology \\
    Atlanta, GA, USA \\
    \texttt{yquan9@gatech.edu} \\
}
\begin{document}
\maketitle
\begin{abstract}
We introduce EconWebArena, a benchmark for evaluating autonomous agents on complex, multimodal economic tasks in realistic web environments. The benchmark comprises 360 curated tasks from 82 authoritative websites spanning domains such as macroeconomics, labor, finance, trade, and public policy. Each task challenges agents to navigate live websites, interpret structured and visual content, interact with real interfaces, and extract precise, time-sensitive data through multi-step workflows. We construct the benchmark by prompting multiple large language models (LLMs) to generate candidate tasks, followed by rigorous human curation to ensure clarity, feasibility, and source reliability. Unlike prior work, EconWebArena emphasizes fidelity to authoritative data sources and the need for grounded web-based economic reasoning. We evaluate a diverse set of state-of-the-art multimodal LLMs as web agents, analyze failure cases, and conduct ablation studies to assess the impact of visual grounding, plan-based reasoning, and interaction design. Our results reveal substantial performance gaps and highlight persistent challenges in grounding, navigation, and multimodal understanding, positioning EconWebArena as a rigorous testbed for economic web intelligence.
\end{abstract}
\section{Introduction}

Accurate and timely access to economic data \citep{einav2014data,einav2014economics} is essential for research, policy analysis, and financial decision-making. Such data are typically published by government agencies, central banks, international organizations, and financial institutions through structured web portals. Retrieving this information often requires navigating dynamic websites, interpreting charts and tables, and interacting with elements like filters, dropdowns, and forms \citep{edelman2012using,ferrara2014web}. Although some platforms offer application programming interfaces (APIs) for direct data access, such interfaces are not consistently available. Many official sources do not support APIs, and those that do often differ significantly in format, coverage, and usability across countries and institutions. In practice, user-facing websites are more common and standardized, making them a more accessible and reliable target for autonomous agents. Nevertheless, many existing approaches rely on general-purpose search engines or pre-collected datasets, which often point to secondary or less reliable sources. This indirect access can introduce errors in precision, units, or interpretation, reducing the reliability of downstream analysis.

Direct interaction with authoritative online sources is critical for high-fidelity economic data acquisition, but it poses unique challenges for autonomous agents. These agents must operate in live web environments, reason over both structured and visual content, and execute multi-step procedures to obtain specific, verifiable information. In practice, such tasks mirror typical workflows in applied economics, for example retrieving official Consumer Price Index (CPI) releases for inflation analysis, collecting central bank interest rate data for policy evaluation, or accessing trade and labor statistics for empirical research. However, existing benchmarks rarely reflect these demands. Most web agent benchmarks \citep{zhou2024webarena,drouin2024workarena,yoran2024assistantbench} focus on general-purpose tasks such as shopping, email handling, or navigating productivity tools. These emphasize routine interactions but overlook the structured reasoning, domain expertise, and precision needed in economic contexts.

\begin{figure*}[!h]
    \centering
    \includegraphics[width=\linewidth]{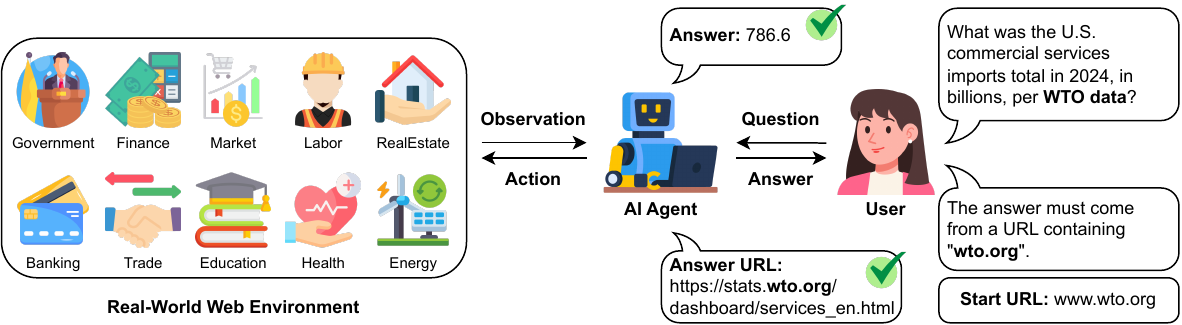}
    \caption{Overview of EconWebArena where agents solve realistic economic tasks by navigating real websites, interpreting content, and extracting grounded numeric answers.}
    \label{fig:benchmark}
\end{figure*}

\begin{figure*}[!h]
    \centering
    \setlength\fboxsep{1pt}
    \setlength\fboxrule{0.5pt}

    \begin{subfigure}[t]{0.32\textwidth}
        \centering
        \fbox{\includegraphics[width=\linewidth]{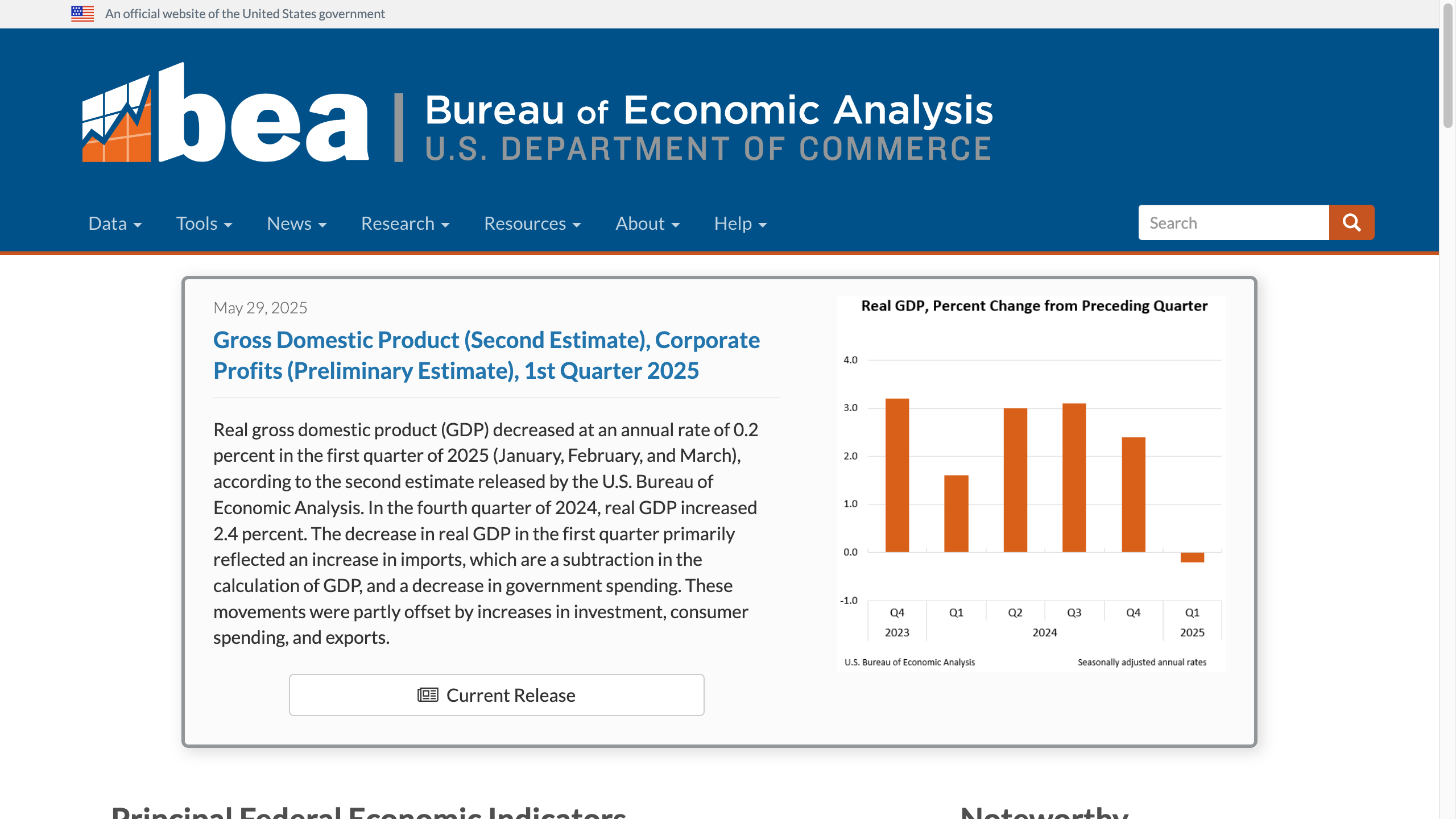}}
        \caption{Open an economic website}
    \end{subfigure}
    \hfill
    \begin{subfigure}[t]{0.32\textwidth}
        \centering
        \fbox{\includegraphics[width=\linewidth]{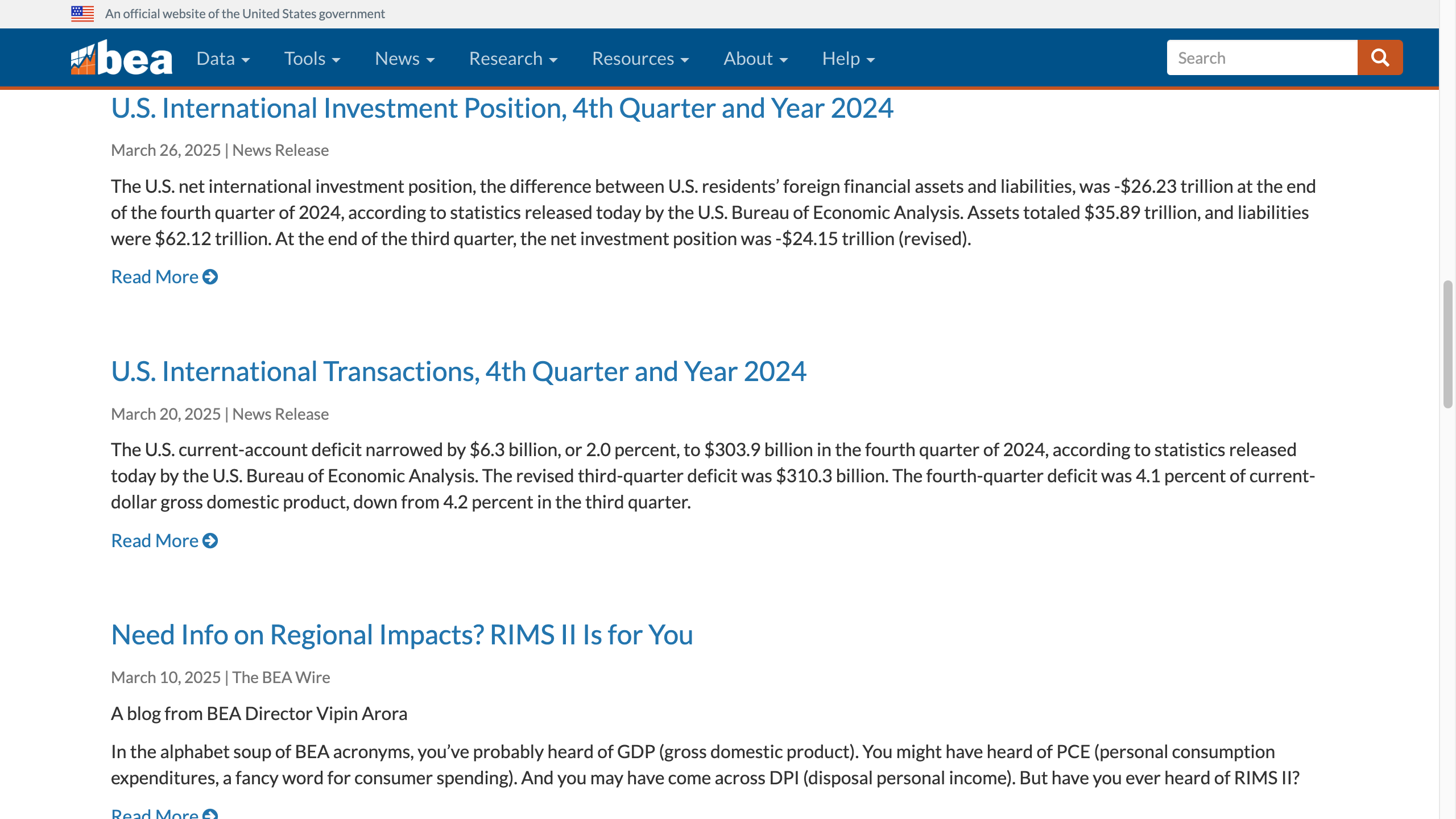}}
        \caption{Navigate to a relevant report}
    \end{subfigure}
    \hfill
    \begin{subfigure}[t]{0.32\textwidth}
        \centering
        \fbox{\includegraphics[width=\linewidth]{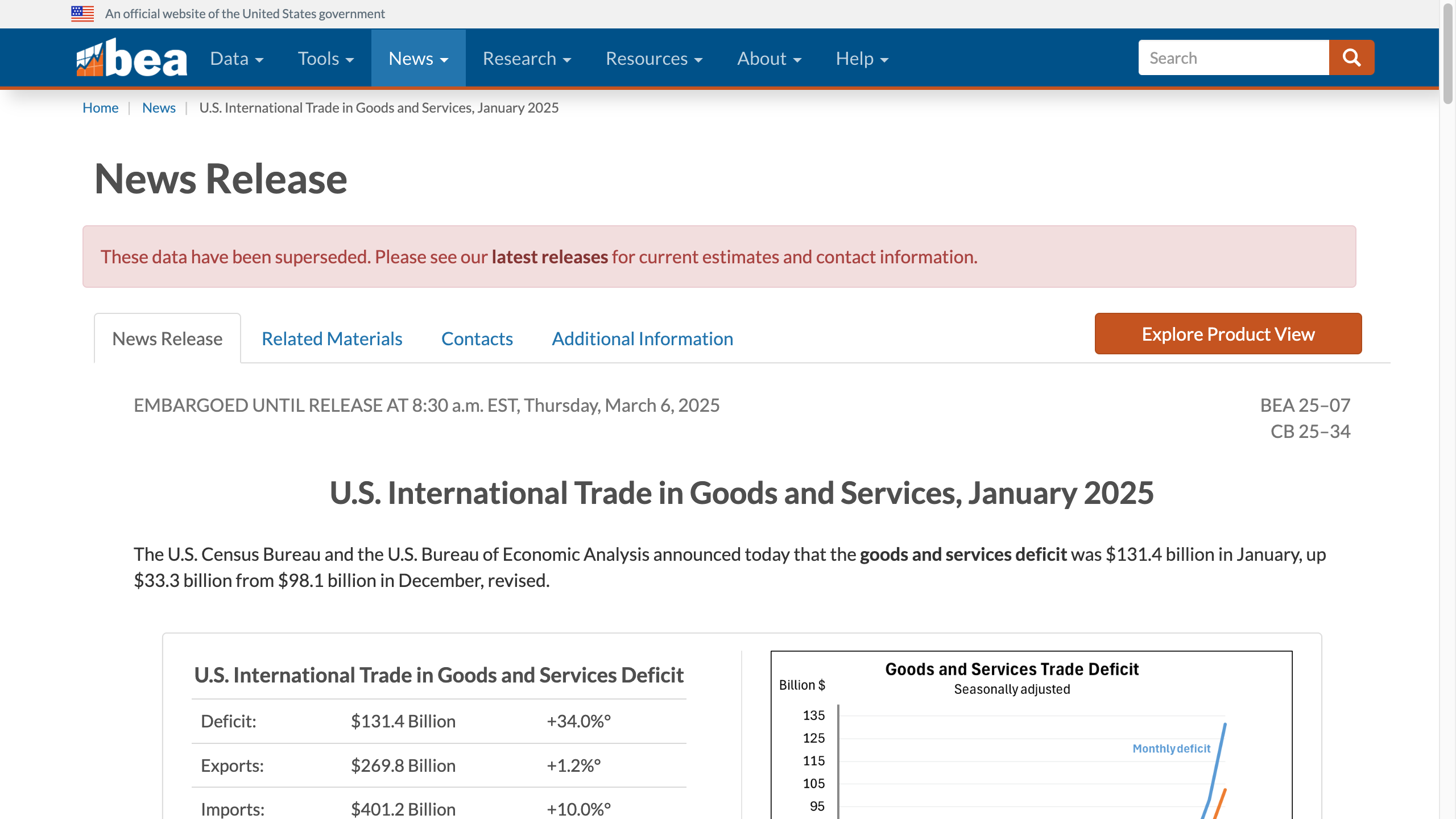}}
        \caption{Extract the required value}
    \end{subfigure}

    \caption{Illustrative example of a typical EconWebArena task. Solving a task involves multi-step navigation across economic data websites, accessing the appropriate report or database, and extracting a target numeric value from tables, diagrams, or databases.}
    \label{fig:success-235}
\end{figure*}

To address this gap, we introduce EconWebArena\footnote{\url{https://econwebarena.github.io/}}, a benchmark for evaluating autonomous agents on realistic economic tasks embedded in real-world websites. EconWebArena comprises 360 tasks based on 82 high-quality data sources across domains such as macroeconomics, labor, trade, and public policy. Each task is designed to test an agent's ability to navigate, interpret, and extract accurate economic data through live browser interactions. As illustrated in Figure~\ref{fig:benchmark} and Figure~\ref{fig:success-235}, EconWebArena covers a wide range of economic domains and requires agents to navigate authoritative websites, interpret diverse content types, and follow multi-step workflows grounded in real data. By combining domain-specific reasoning with realistic web workflows, EconWebArena offers a rigorous evaluation setting for next-generation multimodal large language model (MLLM) \citep{zhang2024mm} agents in applied economic contexts. Our contributions are threefold: 1) we propose a novel benchmark that captures the real-world demands of web-based economic reasoning and data retrieval, 2) we evaluate a diverse set of state-of-the-art MLLM agents and analyze their failure modes, and 3) we conduct ablation studies to assess the effect of visual grounding, reasoning strategies, and interaction design on task performance.

\section{Related Work}

Recent benchmarks for autonomous agents have primarily focused on general-purpose tasks in simulated or real web environments, with growing attention to multimodal grounding, interface interaction, and domain-specific reasoning.

\subsection{Web-Based Benchmarks}

Web-based agent benchmarks have primarily focused on general-purpose tasks such as navigation, data entry, and e-commerce, with foundational platforms like WebArena \citep{zhou2024webarena} and BrowserGym \citep{chezelles2024browsergym} enabling large-scale or modular evaluation. Specialized benchmarks include WebShop \citep{yao2022webshop} for online shopping and WorkArena \citep{drouin2024workarena, boisvert2024workarena++} for productivity tasks. More recent efforts such as AssistantBench \citep{yoran2024assistantbench}, Mind2Web \citep{deng2023mind2web}, WebCanvas \citep{pan2024webcanvas}, RealWebAssist \citep{ye2026realwebassist}, and WebLINX \citep{lu2024weblinx} aim to test broader agent capabilities in cross-task generalization, long-horizon reasoning, and interactive assistance. To support multimodal grounding, benchmarks like VisualWebArena \citep{koh2024visualwebarena}, VideoWebArena \citep{jang2025videowebarena}, and MMINA \citep{tian2025mmina} incorporate image and video inputs, while VisualAgentBench \citep{liu2025visualagentbench} and VisualWebBench \citep{liu2024visualwebbench} assess agents' ability to align language with UI elements. However, none of these settings focus on the specialized challenges of economic data retrieval and domain-specific reasoning. While they could in principle be adapted to economics, such modifications would not ensure grounding in authoritative sources or the numeric precision required in financial contexts, which EconWebArena is designed to support. Unlike prior work that emphasizes scale or multimodal grounding, our benchmark targets domain-specific workflows that require careful handling of reporting periods and reliable numeric accuracy.

\subsection{Economic Benchmarks}

Benchmarks in economics and finance largely emphasize static inputs such as question answering, document analysis, or language modeling. EconQA \citep{van2023evaluating}, FinanceBench \citep{islam2023financebench}, and EconLogicQA \citep{quan2024econlogicqa} evaluate LLMs on domain-specific factual and reasoning questions, while FLUE/FLANG \citep{shah2022flue} target financial language modeling. Other efforts explore tool-augmented agents in finance \citep{zhang2024multimodal}, CRM tasks \citep{huang2025crmarena}, or multi-agent simulations such as InvAgent \citep{quan2024invagent} and EconArena \citep{guo2024economics}. Several benchmarks also extend to non-English financial domains, including BBT-Fin \citep{lu2023bbt} and FinEval \citep{guo2025fineval}. However, these settings mostly rely on static content and do not capture the web-based workflows necessary for real-world economic data retrieval.

\section{EconWebArena}

\begin{figure*}[!h]
    \centering
    \includegraphics[width=.9\linewidth]{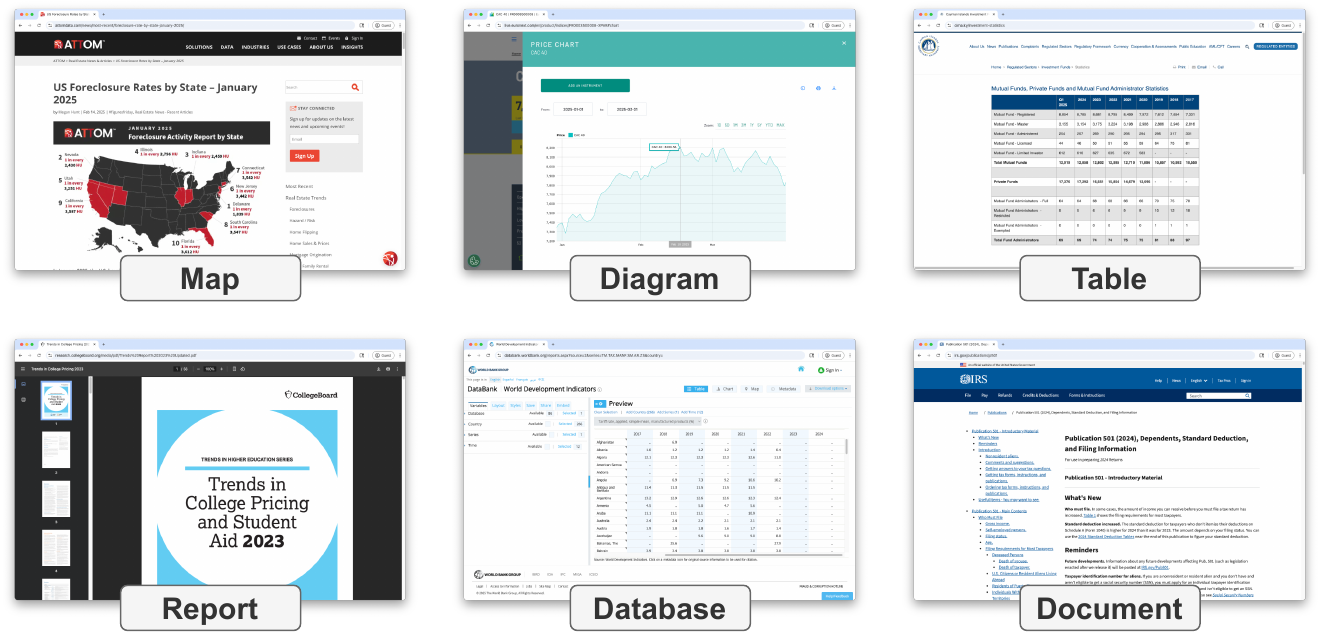}
    \caption{Representative EconWebArena tasks involve interacting with diverse web elements, including maps, diagrams, tables, reports, databases, documents, and so on.}
    \label{fig:examples}
\end{figure*}

In this section, we describe the construction of EconWebArena, a benchmark for evaluating agents on realistic economic tasks in web environments. We outline the process of task generation, manual curation and annotation, and the criteria used for answer evaluation.

\subsection{Task Generation}

To construct the benchmark, we prompt four state-of-the-art large language models (LLMs): GPT-4o \citep{openai2024gpt4o}, Claude-3.7-Sonnet \citep{anthropic2025claude37}, DeepSeek-V3 \citep{liu2024deepseek}, and Gemini-2.0-Flash \citep{deepmind2024gemini2}. Each model generates 50 candidate tasks, resulting in 200 initial questions. The prompt is carefully designed to elicit realistic, high-level tasks that autonomous agents can perform on real-world economic websites. It enforces strict guidelines for clarity, specificity, and feasibility, requiring each task to be a concise question with a single verifiable answer, using only fixed date ranges and objective language. Tasks must involve meaningful economic content, such as finance, labor, trade, or macroeconomic indicators, and be executable via realistic web interactions like navigating, filtering, or completing forms. To ensure diversity and coverage, each task must target a different named website. The full prompt is provided in Figure~\ref{fig:prompt} in Appendix~\ref{sec:generation-prompt}. As illustrated in Figure~\ref{fig:examples}, the resulting tasks span a wide range of interactive web elements, including maps, tables, diagrams, documents, and databases.

\subsection{Task Curation and Annotation}

Following task generation, we manually review, filter, and revise all 200 candidate tasks to ensure clarity, reliability, and feasibility. We retain only those that point to verifiable economic data from reputable public sources, discarding any that are vague, redundant, rely on subjective interpretation, or require access to subscription-based or non-English websites. For each accepted task, we identify the exact webpage containing the correct answer and record the corresponding numeric value. This process results in 120 high-quality seed tasks, each clearly stated, yielding a single numeric answer, and grounded in authoritative data.

Each task is assigned to one of ten high-level categories based on the domain of its source website: government, finance, markets, labor, banking, energy, trade, real estate, education, or health. These categories capture a wide spectrum of economic subject areas, ranging from macroeconomic indicators and central banking to employment, housing, and healthcare. The benchmark draws from 82 authoritative websites with the complete list detailed in Appendix~\ref{sec:website-list} Table~\ref{tab:websites}. Definitions for each category are provided in Appendix~\ref{sec:website-categories}.

\begin{figure}[!h]
    \centering
    \includegraphics[width=.9\linewidth]{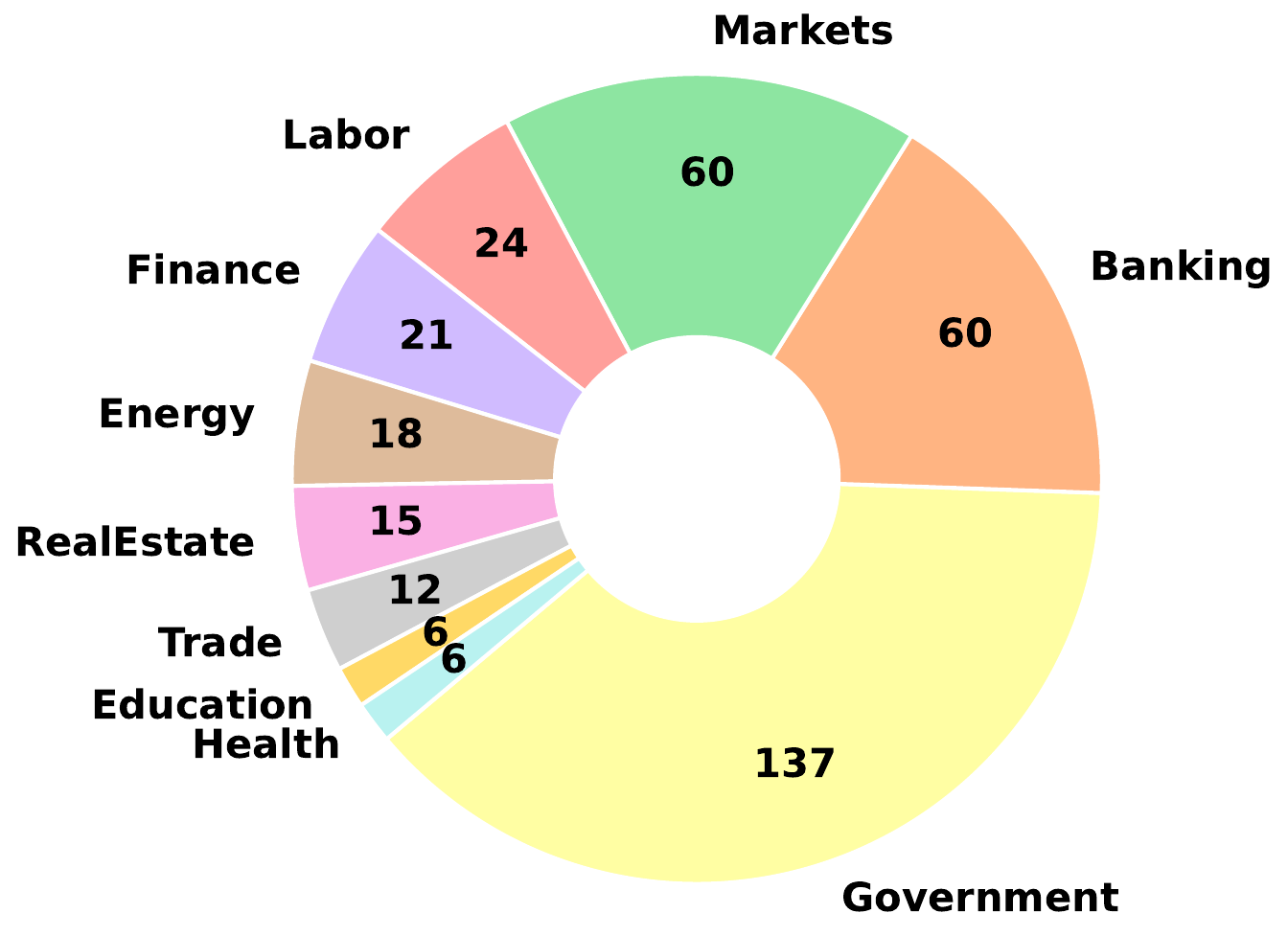}
    \caption{Distribution of task categories in EconWebArena, reflecting the diversity of economic domains represented in the benchmark.}
    \label{fig:pie}
\end{figure}

To expand the dataset, we generate two variants per seed task by modifying elements such as time range, country, or indicator. Tasks involving daily or monthly data are limited to early 2025, while quarterly and annual queries span 2022 to 2025 to ensure temporal relevance. We enforce strict output constraints so that each answer is a single numeric value, enabling consistent automated evaluation. Each finalized task includes a question with a start URL, an expected answer format, and a required domain in the answer URL. For multilingual websites, we use the English-language version as the start URL. This process yields 360 tasks in total, with category distribution shown in Figure~\ref{fig:pie}. Task examples are shown in Table~\ref{tab:examples} and the full dataset is publicly available\footnote{\url{https://huggingface.co/datasets/EconWebArena/EconWebArena}}.

\begin{table*}[!h]
\small
\centering
\begin{tabular}{>{\RaggedRight\arraybackslash}p{4em} >{\RaggedRight\arraybackslash}p{22em} >{\RaggedRight\arraybackslash}p{8em} > {\RaggedRight\arraybackslash}p{3em} > {\RaggedRight\arraybackslash}p{6.5em}}
\toprule
\textbf{Category} & \textbf{Task Description} & \textbf{Start URL} & \textbf{Answer} & \textbf{Domain} \\
\midrule
Government & As published by the Office for National Statistics, what was the CPIH annual inflation rate for all items (2015=100) in the United Kingdom in March 2025? Provide only the number as a decimal with one digit after the decimal point, without percent symbols or other units. & \url{https://www.ons.gov.uk/} & 3.4 & ons.gov.uk \\
Energy & As reported by the U.S. Energy Information Administration, what was the average retail price of regular gasoline in California during the week of March 24, 2025, in dollars per gallon? Provide only the number as a decimal with three digits after the decimal point, without currency symbols, commas, or other units. & \url{https://www.eia.gov/} & 4.418 & eia.gov \\
Markets & As reported by Cox Automotive, what was the total number of unsold used vehicles in the United States as of March 31, 2025? Provide only the number as a decimal with two digits, in millions, without commas or other units. & \url{https://www.coxautoinc.com/} & 2.14 & coxautoinc.com \\
Banking & As reported by the Federal Reserve Bank of New York, what was the effective federal funds rate on January 10, 2025? Provide only the number as a decimal with two digits, without percent symbols or other units. & \url{https://www.newyorkfed.org/} & 4.33 & newyorkfed.org \\
\bottomrule
\end{tabular}
\caption{Representative EconWebArena tasks illustrating diverse economic domains, answer formats, and real-world data sources.}
\label{tab:examples}
\end{table*}

\subsection{Observation and Action Spaces}

EconWebArena extends the BrowserGym\footnote{\url{https://github.com/ServiceNow/BrowserGym}} \citep{chezelles2024browsergym} framework to support economic tasks in realistic web environments. Each task provides the agent with structured observations, including the accessibility tree (AXTree), webpage screenshots, and contextual metadata such as focused elements, error logs, and action history. These inputs enable both language-based and multimodal reasoning. The action space includes fine-grained browser control primitives at both the element and coordinate levels, supporting operations like mouse movement, text input, tab switching, and form submission. This setup allows agents to perform the low-level interactions required for complex economic data retrieval. Further details are provided in Section~\ref{sec:experimental-setup}.

\subsection{Answer Evaluation}

We evaluate each agent response using two criteria. First, the predicted answer must contain the correct numeric value, typically an integer or decimal that represents an economic measure such as a price, rate, or index. The value must match the gold annotation exactly, even if it appears within a longer sentence. Second, the response must include a valid URL that contains the required domain name, confirming the information was retrieved from the intended authoritative source. A task is considered correct only if both conditions are met. This evaluation supports automated scoring while ensuring factual accuracy and source reliability. We report success rate as the percentage of correctly answered tasks within each category and also include the average number of steps taken on successful tasks.

\section{Experiments}

In this section, we evaluate a range of agents on EconWebArena, examining their overall performance, common failure patterns, and the effects of different design choices through ablation studies.

\subsection{Experimental Setup}
\label{sec:experimental-setup}

We implement the EconWebArena tasks using BrowserGym\footnote{\url{https://github.com/ServiceNow/BrowserGym}} \citep{chezelles2024browsergym}, a lightweight browser simulation environment that enables fine-grained agent control in web-based settings. The action space includes a broad range of primitives for common browser operations, such as clicking, typing, filling input fields, hovering, and scrolling. Agents can also manage tabs and navigate across pages. Interactions are supported at both the element level (via structured identifiers) and the coordinate level (via pixel positions), allowing flexible strategies for handling diverse web layouts. Additional actions include sending messages to the user and performing no-op delays. A full list of supported actions is shown in Table~\ref{tab:actions}.

\begin{table}[!h]
\small
\centering
\begin{tabular}{ll}
\toprule
\textbf{Primitive} & \textbf{Description} \\
\midrule
\multicolumn{2}{c}{\textbf{\texttt{bid}}} \\
\midrule
\texttt{fill} & Input text \\
\texttt{click} & Click element \\
\texttt{hover} & Hover on element \\
\texttt{press} & Press key combination \\
\texttt{focus} & Focus element \\
\texttt{clear} & Clear input \\
\texttt{select\_option} & Select dropdown option \\
\midrule
\multicolumn{2}{c}{\textbf{\texttt{coord}}} \\
\midrule
\texttt{mouse\_move} & Move mouse \\
\texttt{mouse\_click} & Click by position \\
\texttt{mouse\_drag\_and\_drop} & Drag and drop \\
\texttt{keyboard\_press} & Press key(s) \\
\texttt{keyboard\_type} & Type with keyboard \\
\midrule
\multicolumn{2}{c}{\textbf{\texttt{tab}}} \\
\midrule
\texttt{new\_tab} & Open new tab \\
\texttt{tab\_close} & Close current tab \\
\texttt{tab\_focus} & Focus on tab \\
\midrule
\multicolumn{2}{c}{\textbf{\texttt{nav}}} \\
\midrule
\texttt{go\_back} & Navigate back \\
\texttt{go\_forward} & Navigate forward \\
\texttt{goto} & Go to URL \\
\midrule
\multicolumn{2}{c}{\textbf{\texttt{misc}}} \\
\midrule
\texttt{send\_msg\_to\_user} & Send message \\
\texttt{scroll} & Scroll page \\
\texttt{noop} & Wait without action \\
\bottomrule
\end{tabular}
\caption{Action space in EconWebArena, organized by primitive category based on BrowserGym.}
\label{tab:actions}
\end{table}

The web agents are built using AgentLab\footnote{\url{https://github.com/ServiceNow/AgentLab}} \citep{chezelles2024browsergym}, a modular framework for prompting and executing LLM-based agents. Each agent receives structured observations, including the AXTree, a webpage screenshot, metadata on the focused element, recent errors, and the full history of prior actions and thoughts. We enable key prompt features such as chain-of-thought reasoning, contextual hints, and real-world examples, while disabling multi-action execution and memory retrieval. Additional settings control element extraction, coordinate formatting, and response style. These configurations are tuned to balance informativeness and efficiency within token limits. The full setup is listed in Appendix~\ref{sec:more-settings} Table~\ref{tab:settings}.

For evaluation, we select a range of multimodal large language models (MLLMs), covering both proprietary and open-weight models: GPT-4o \citep{achiam2023gpt,openai2024gpt4o}, GPT-4.1 \citep{openai2025gpt41}, o4-mini \citep{openai2025o4mini}, Claude Sonnet 4 \citep{anthropic2025claude}, Gemini 2.5 Flash \citep{team2023gemini,deepmind2025geminiflash}, and Llama 4 Maverick \citep{grattafiori2024llama,meta2025llama4}. Each model is allowed up to 30 steps per task to prevent looping or excessive action chains. Due to API credit constraints, we run a single trial per model. Since each seed task has two variants, the reported results reflect an average over three instances per seed task. In addition, we conduct human evaluation by dividing the benchmark among the authors (graduate-level researchers with general knowledge of economics), ensuring that each annotator reviews only tasks they had not previously seen. All experiments were conducted during the final week of May 2025 to ensure consistency across models and reduce variance due to changes in live website content.

\subsection{Experimental Results}

\begin{table*}[!h]
\small
\centering
\begin{tabular}{lr|rrrrrr|r}
\toprule
\textbf{Category} & \textbf{Tasks} & \textbf{o4-mini} & \textbf{GPT-4.1} & \textbf{GPT-4o} & \textbf{Claude-4} & \textbf{Gemini-2.5} & \textbf{Llama-4} & \textbf{Human} \\
\midrule
Banking & 60 & 41.7\% & 23.3\% & 18.3\% & 38.3\% & 28.3\% & 21.7\% & 95.0\% \\
Finance & 21 & 33.3\% & 14.3\% & 14.3\% & 23.8\% & 33.3\% & 9.5\% & 95.2\% \\
Government & 138 & 57.2\% & 45.7\% & 35.5\% & 47.1\% & 39.1\% & 26.1\% & 91.3\% \\
Labor & 24 & 20.8\% & 0.0\% & 8.3\% & 12.5\% & 4.2\% & 4.2\% & 91.7\% \\
Markets & 60 & 48.3\% & 35.0\% & 33.3\% & 41.7\% & 33.3\% & 15.0\% & 96.7\% \\
Other & 57 & 42.1\% & 24.6\% & 21.1\% & 31.6\% & 22.8\% & 12.3\% & 93.0\% \\
\midrule
All SR ($\uparrow$) & 360 & \textbf{46.9\%} & 31.9\% & 26.9\% & 38.6\% & 31.1\% & 18.9\% & 93.3\% \\
\midrule
Steps ($\downarrow$) & - & 8.99 & \textbf{7.23} & 7.77 & 11.77 & 9.29 & 9.54 & - \\
\bottomrule
\end{tabular}
\caption{Average task success rates (SR) and average steps (on successful tasks) on EconWebArena by major category for models (o4-mini, GPT-4.1, GPT-4o, Claude Sonnet 4, Gemini 2.5 Flash, Llama 4 Maverick) and human. (Other: Energy, RealEstate, Trade, Education, and Health.)}
\label{tab:results}
\end{table*}

To better understand model capabilities on EconWebArena, we organize our analysis around three core research questions (RQs) concerning performance, failure patterns, and potential improvements. Table~\ref{tab:results} reports overall accuracy across major economic categories, while Table~\ref{tab:more-results} in Appendix~\ref{sec:more-results} provides more detailed, category-specific breakdowns.

\subsubsection{RQ1: How well can LLM agents solve economic tasks in realistic web environments?}

As shown in Table~\ref{tab:results}, current LLM agents achieve only partial success on EconWebArena. The best-performing model, o4-mini, reaches an average success rate of 46.9\%, while other proprietary models like GPT-4.1 (31.9\%), Claude Sonnet 4 (38.6\%), and Gemini 2.5 Flash (31.1\%) show mixed results across categories. Open-weight model Llama 4 Maverick performs considerably lower, with an overall success rate of 18.9\%. Government and market domains yield higher scores across most models, whereas labor, finance, and other specialized categories remain particularly challenging. Human performance remains consistently high at 93.3\%, underscoring the gap between current agents and expert-level competence. In terms of efficiency, GPT-4.1 completes successful tasks with the fewest steps on average, while Claude Sonnet 4 tends to follow longer action sequences. These results highlight the limitations of existing agents in handling complex, real-world economic workflows. Representative success cases are provided in Appendix~\ref{sec:success-cases}.

\subsubsection{RQ2: What causes LLM agents to fail on EconWebArena tasks?}

\begin{table}[!h]
\small
\centering
\begin{tabular}{lrr}
\toprule
\textbf{Error Type} & \textbf{Count} & \textbf{Percentage} \\
\midrule
Access Issues & 16 & 25.0\% \\
Data Extraction Errors & 16 & 25.0\% \\
Interaction Failures & 8 & 12.5\% \\
Navigation Failures & 15 & 23.4\% \\
Visual Understanding Failures & 9 & 14.1\% \\
\midrule
All & 64 & 100.0\%\\
\bottomrule
\end{tabular}
\caption{Distribution of failure cases by error type in o4-mini's performance on EconWebArena seed tasks.}
\label{tab:errors}
\end{table}

To understand why LLM agents fail on EconWebArena tasks, we perform a detailed error analysis on the o4-mini model. Out of 360 benchmark tasks, o4-mini fails 193, among which 64 are seed tasks. We manually review these 64 failures to identify common error patterns and categorize them into five distinct types: access issues, data extraction errors, navigation failures, visual understanding failures, and interaction failures. Table~\ref{tab:errors} summarizes the distribution of these error types, and Figure~\ref{fig:errors} illustrates representative examples for each category (see Appendix~\ref{sec:error-cases} for full task visualizations).

Access issues (Figure~\ref{fig:error-access}) include problems such as blocked or restricted websites, loading errors, or access denials that prevent task execution entirely. Data extraction errors (Figure~\ref{fig:error-extraction}) occur when the agent navigates to the correct page but extracts the wrong value, often due to confusion among dense statistics or approximate retrieval. Navigation failures (Figure~\ref{fig:error-navigation}) reflect difficulty locating the correct webpage or section, including getting stuck or choosing incorrect paths. Visual understanding failures (Figure~\ref{fig:error-visual}) are due to an inability to interpret charts, diagrams, or other non-textual representations. Interaction failures (Figure~\ref{fig:error-interaction}) stem from execution errors in UI manipulation, such as failing to trigger dropdowns or input fields. These results suggest that agent failure often stems from combined weaknesses in reasoning, perception, and interface fluency when dealing with complex economic content online.

\begin{figure*}[!h]
    \centering
    \setlength\fboxsep{1pt}
    \setlength\fboxrule{0.5pt}

    \begin{subfigure}[t]{0.32\textwidth}
        \centering
        \fbox{\includegraphics[width=\linewidth]{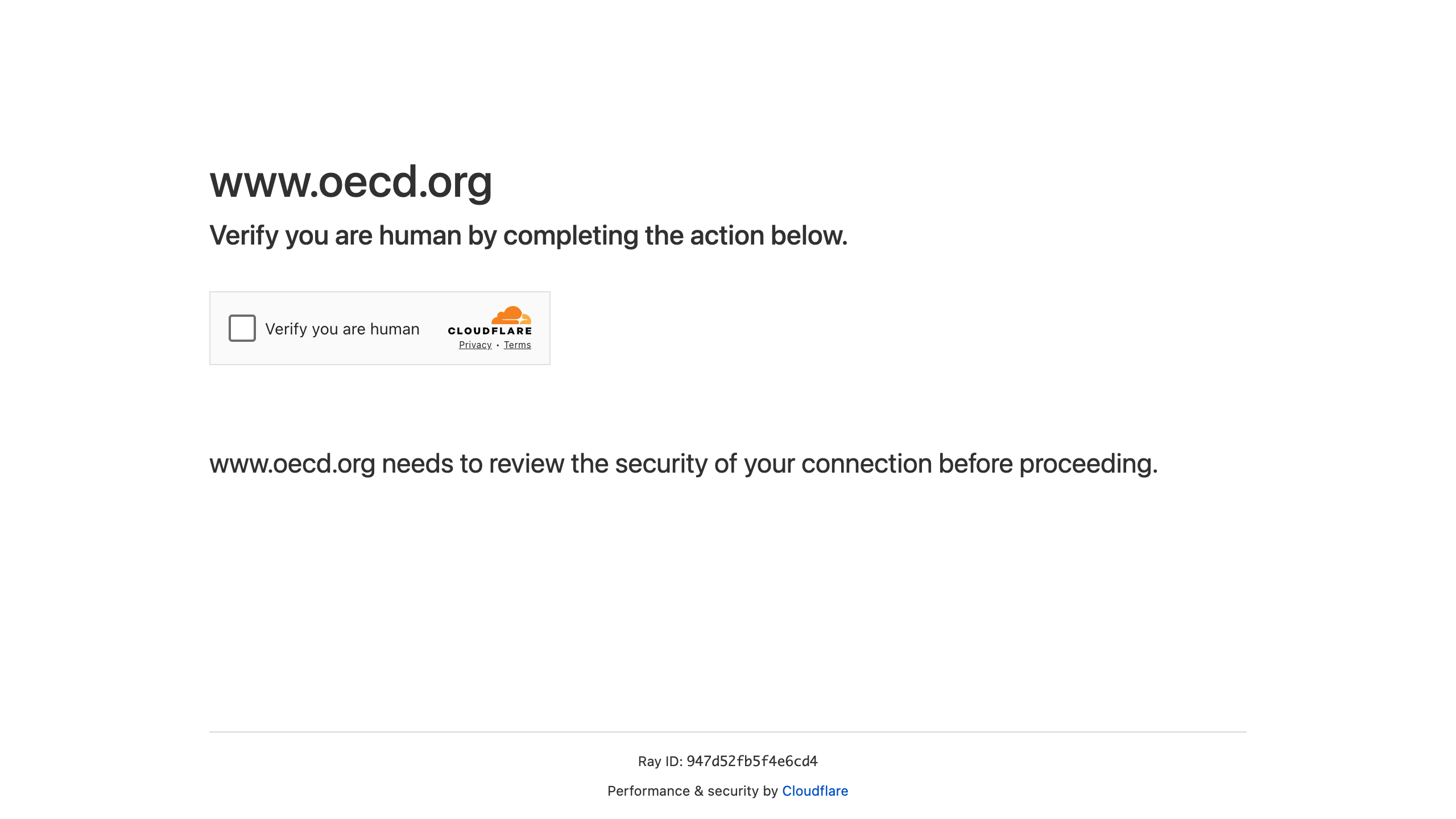}}
        \caption{Access Issue}
        \label{fig:error-access}
    \end{subfigure}
    \hfill
    \begin{subfigure}[t]{0.32\textwidth}
        \centering
        \fbox{\includegraphics[width=\linewidth]{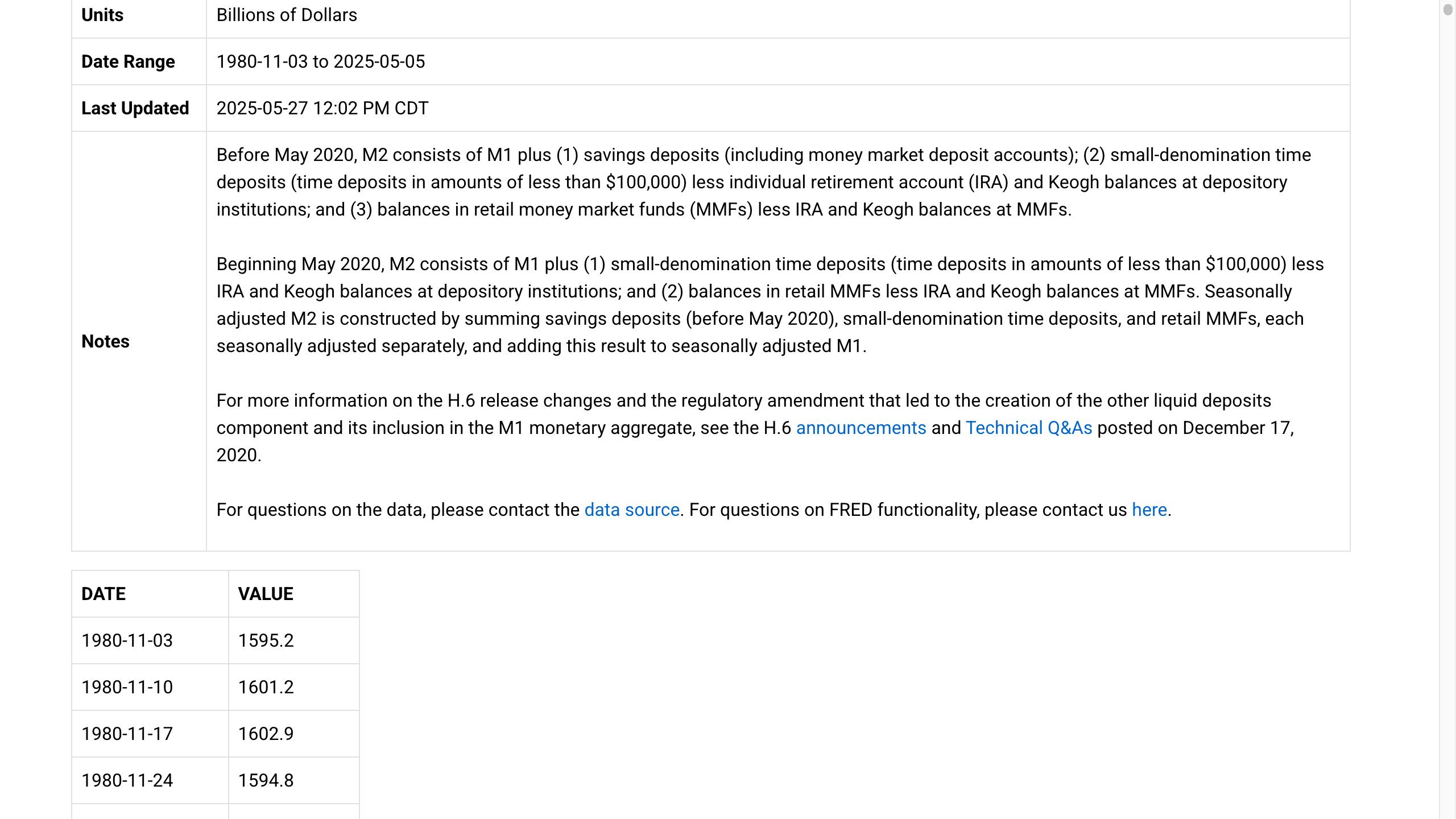}}
        \caption{Data Extraction Error}
        \label{fig:error-extraction}
    \end{subfigure}
    \hfill
    \begin{subfigure}[t]{0.32\textwidth}
        \centering
        \fbox{\includegraphics[width=\linewidth]{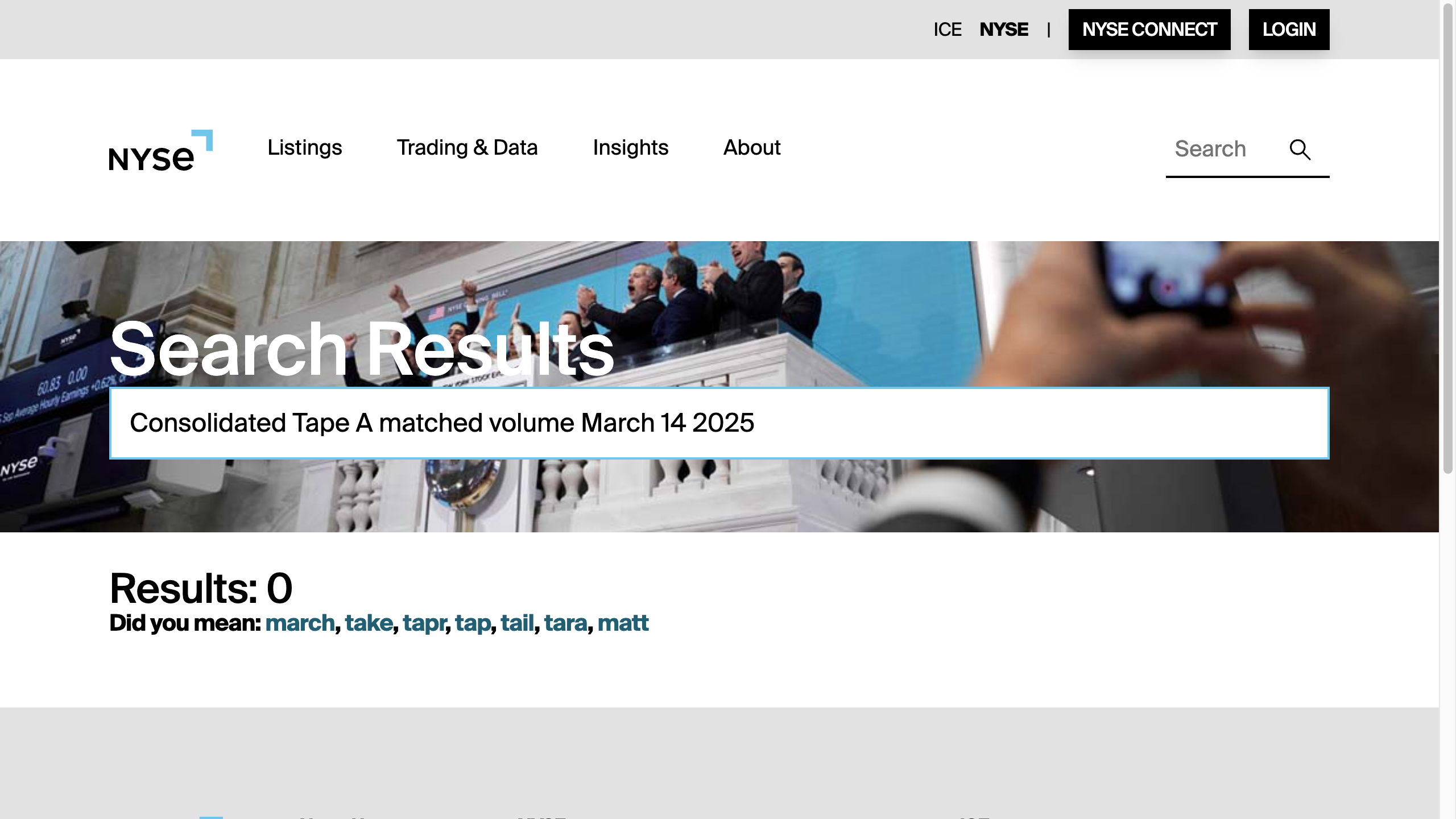}}
        \caption{Navigation Failure}
        \label{fig:error-navigation}
    \end{subfigure}

    \vspace{1em}

    \begin{subfigure}[t]{0.32\textwidth}
        \centering
        \fbox{\includegraphics[width=\linewidth]{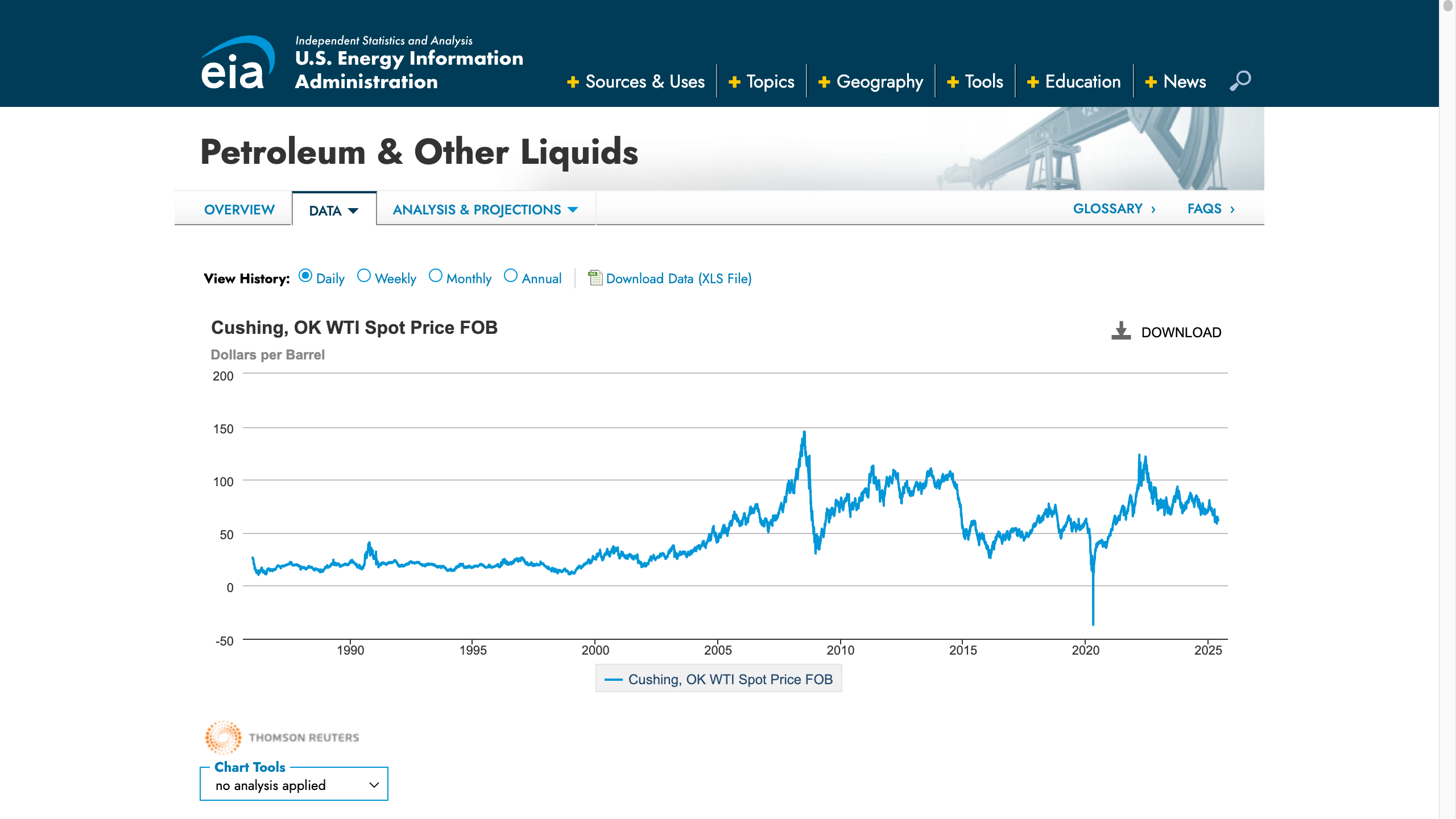}}
        \caption{Visual Understanding Failure}
        \label{fig:error-visual}
    \end{subfigure}
    \hspace{0.4em}
    \begin{subfigure}[t]{0.32\textwidth}
        \centering
        \fbox{\includegraphics[width=\linewidth]{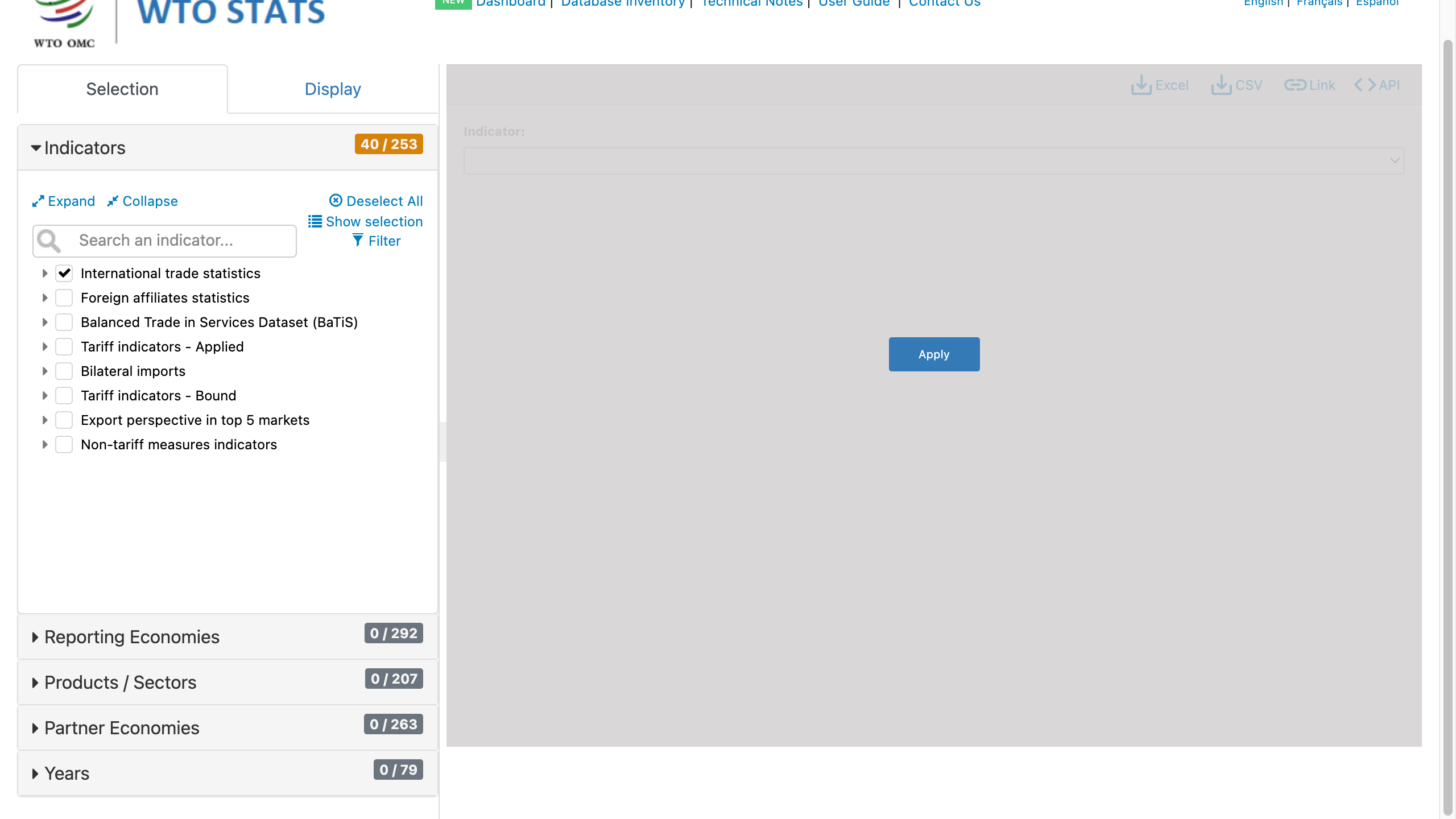}}
        \caption{Interaction Failure}
        \label{fig:error-interaction}
    \end{subfigure}

    \caption{Representative examples of common error types from o4-mini in EconWebArena: access issues, data extraction errors, navigation failures, visual understanding failures, and interaction failures.}
    \label{fig:errors}
\end{figure*}

\subsubsection{RQ3: What directions can improve LLM agent performance on these tasks?}

Our error analysis suggests several directions to improve LLM agents on EconWebArena tasks. First, robust visual grounding and multimodal reasoning are essential for interpreting economic charts, tables, and diagrams, which frequently appear in official data portals. Structured input representations, such as combining AXTree with screenshots and coordinates, can better support perception and interaction. Planning and self-criticism prompting also improve performance in multi-step navigation tasks. In addition, hybrid methods that integrate GUI-based exploration with structured API calls, when available, can enhance both reliability and efficiency. Moreover, access issues such as blocked or unstable sites remain a challenge for LLM agents, though approaches like caching snapshots or API fallbacks can help reduce their impact. Finally, domain-adaptive tuning and retrieval-augmented prompting using realistic economic examples may further strengthen agent accuracy in complex cases.

\subsection{Ablation Studies}

To assess the contribution of each prompt configuration, we conduct ablation experiments on the o4-mini agent, varying components related to observation, action, and reasoning. The initial setup is described in Appendix~\ref{sec:more-settings} Table~\ref{tab:settings}, and results are reported in Table~\ref{tab:ablation}. Each variant alters a single feature based on AgentLab conventions \citep{chezelles2024browsergym}.

\begin{table}[!h]
\small
\centering
\begin{tabular}{lrr}
\toprule
\textbf{Configuration} & \textbf{SR} ($\uparrow$) & \textbf{Steps} ($\downarrow$) \\
\midrule
Initial configuration & 46.9\% & 8.99 \\
\midrule
\multicolumn{3}{c}{\textbf{\texttt{observation}}} \\
\midrule
\texttt{- use\_ax\_tree + use\_html} & 36.7\% & 9.61 \\
\texttt{- use\_screenshot} & 44.7\% & 8.88 \\
\texttt{+ use\_som} & 47.2\% & 8.71 \\
\texttt{- use\_history} & 45.8\% & 7.76 \\
\texttt{- extract\_coords} & 43.6\% & 8.61 \\
\midrule
\multicolumn{3}{c}{\textbf{\texttt{action}}} \\
\midrule
\texttt{+ multiaction} & 41.9\% & 7.37 \\
\midrule
\multicolumn{3}{c}{\textbf{\texttt{reasoning}}} \\
\midrule
\texttt{- use\_thinking} & 46.9\% & 8.93 \\
\texttt{+ use\_plan} & 49.4\% & 9.86 \\
\texttt{+ use\_criticise} & 47.5\% & 8.61 \\
\bottomrule
\end{tabular}
\caption{Ablation results for the o4-mini model, showing the impact of prompt configuration options on EconWebArena success rate (SR).}
\label{tab:ablation}
\end{table}

Removing the structured accessibility tree and using only raw HTML (\texttt{-use\_ax\_tree +use\_html}) leads to a large performance drop (from 46.9\% to 36.7\%), confirming the critical role of clean DOM representations. Excluding screenshots (\texttt{-use\_screenshot}) causes a modest decline (44.7\%), but visual information remains essential for tasks requiring interpretation of charts and diagrams. Removing history context (\texttt{-use\_history}) slightly reduces success rate (45.8\%) but leads to noticeably fewer steps (7.76), suggesting that brief context improves efficiency but is not always necessary for correctness. Turning off coordinate extraction (\texttt{-extract\_coords}) also hurts performance (43.6\%), as many UI components require position-based interaction that AXTree alone may not cover. Enabling set-of-mark prompting (\texttt{+use\_som}) \citep{yang2023set} slightly improves accuracy (47.2\%), indicating its benefit for spatial referencing.

In the action section, enabling multiple action execution (\texttt{+multiaction}) leads to a lower average step count (7.37) but reduces accuracy to 41.9\%, indicating a trade-off between efficiency and reliability. For reasoning, removing explicit thinking traces (\texttt{-use\_thinking}) has no measurable effect on success rate (46.9\%), suggesting that o4-mini is already capable of internal step-by-step reasoning. Adding high-level planning (\texttt{+use\_plan}) yields the largest improvement in success rate (49.4\%) but also increases the average number of steps, reflecting its benefit for long-horizon tasks. Enabling self-critique (\texttt{+use\_criticise}) slightly improves success rate (47.5\%) and reduces steps (8.61), suggesting it helps refine decision-making. Overall, these results emphasize the importance of structured observations, visual grounding, coordinate awareness, and global planning in complex economic web tasks.

\section{Conclusion}

We present EconWebArena, a benchmark for evaluating autonomous agents on complex economic tasks grounded in real-world web environments. The benchmark includes 360 manually curated tasks across 82 authoritative websites and ten economic domains, requiring agents to navigate live webpages, interpret structured and visual content, and extract precise, time-sensitive data through multi-step interactions. Our evaluation of state-of-the-art multimodal large language models reveals significant limitations in grounding, navigation, and multimodal reasoning. Error analysis highlights persistent failure modes, while ablation studies identify key configuration choices that impact performance. EconWebArena provides a challenging and realistic testbed for advancing domain-aware, interaction-capable agents for economic data retrieval and reasoning.

\section*{Limitations}

EconWebArena uses a strict evaluation metric based on exact numeric matching and source URL verification, which ensures high precision but may not account for partially correct answers or valid intermediate reasoning steps. While the benchmark operates on live web content, it is limited to publicly accessible pages and excludes websites requiring login, subscription, or region-specific access. All tasks are presented in English, which restricts the benchmark's applicability in multilingual or localized settings. Moreover, EconWebArena focuses on single-turn, goal-oriented tasks and does not capture longer-term planning or interactive dialogue scenarios.

\section*{Ethical Considerations}

This work follows principles of transparency, reproducibility, and responsible AI development. All tasks are sourced from publicly available, authoritative economic websites, with no use of personal or sensitive data. A thorough human review process filters out ambiguous, unverifiable, or inappropriate content to ensure quality and integrity. Since the benchmark does not involve human subjects or private information, it presents minimal ethical risk. We openly release the dataset and tools to support further research, with an emphasis on factual accuracy and source accountability.

\section*{Acknowledgments}

This research was supported in part by API credits provided through the OpenAI Researcher Access Program.

\bibliography{refs}
\appendix
\section{Benchmark Details}
\label{sec:benchmark-details}

This appendix provides additional details on the task generation process, prompt design, and the categorization of websites used in EconWebArena.

\subsection{Task Generation Prompt}
\label{sec:generation-prompt}

To ensure the quality, diversity, and feasibility of the benchmark tasks, we use a carefully designed prompt to guide task generation by large language models. The prompt instructs the models to produce concise, goal-driven questions grounded in real economic scenarios. It specifies formatting rules, data constraints, and task requirements to guarantee that each example can be realistically completed by an autonomous web agent. The full prompt is shown in Figure~\ref{fig:prompt}.

\begin{figure*}[!h]
  \centering
  \begin{tcolorbox}[title=\textbf{Prompt for Economic Task Generation}, colback=gray!5, colframe=gray!40!black]

    {\small

    Generate a list of 50 high-level, realistic, and unambiguous tasks designed to evaluate the performance of autonomous web-based agents in economics-related scenarios. Each task must be phrased as a concise question and presented as a plain-text bullet point.\\

    All tasks must follow these criteria:\\

    $\bullet$ Task format:\\

      \hspace*{2mm} $\circ$ Each task must be phrased as a concise, clearly worded question.\\
      \hspace*{2mm} $\circ$ Each question must have a single, unique, and verifiable answer.\\
      \hspace*{2mm} $\circ$ The answer to each task must be expressible in plain text.\\
      \hspace*{2mm} $\circ$ Present the list of tasks as plain-text bullet points using a dash (``-'') for each item. Do not include numbering or additional formatting.\\

    $\bullet$ Clarity and specificity:\\

      \hspace*{2mm} $\circ$ Use only absolute dates or fixed date ranges. Do not use vague or relative time expressions such as ``current'', ``latest'', ``recent'', or ``as of now''.\\
      \hspace*{2mm} $\circ$ Do not include any dates beyond April 2025.\\
      \hspace*{2mm} $\circ$ Avoid vague or subjective language. Each task must be clearly defined and objectively measurable.\\
      \hspace*{2mm} $\circ$ Do not include URLs, domain names, or any form of web address in the task descriptions.\\

    $\bullet$ Relevance:\\

      \hspace*{2mm} $\circ$ Each task must relate to meaningful economic content, such as markets, trade, labor, finance, regulation, pricing, taxation, or macroeconomic indicators.\\
      \hspace*{2mm} $\circ$ Refer only to specific, named websites where the task should be performed. Do not list alternatives or refer to general categories of websites.\\

    $\bullet$ Web-based execution:\\

      \hspace*{2mm} $\circ$ Each task must be realistically executable on a real-world website using standard browser interactions.\\
      \hspace*{2mm} $\circ$ Tasks must involve interactive actions such as navigating, clicking, typing, selecting, filtering, or filling forms.\\
      \hspace*{2mm} $\circ$ Tasks should generally require multiple user actions and may involve visiting multiple webpages within or across websites.\\

    $\bullet$ Diversity and complexity:\\

      \hspace*{2mm} $\circ$ Avoid simple fact lookups (e.g., a price or rate on a date) unless they are part of a broader, practical objective.\\
      \hspace*{2mm} $\circ$ Include a range of task types and difficulty levels, emphasizing realistic, multi-step, goal-oriented scenarios that reflect real-world economic research, analysis, or decision-making.\\
      \hspace*{2mm} $\circ$ Ensure that each task involves a different website to increase coverage and variety.\\

    Generate exactly 50 tasks that meet all of the above requirements.
    }
  \end{tcolorbox}
  \caption{Prompt used to generate a diverse and rigorous task set for evaluating autonomous web-based agents.}
  \label{fig:prompt}
\end{figure*}

\subsection{Website Categories}
\label{sec:website-categories}

The benchmark tasks are grouped into ten economic categories, each reflecting a distinct type of data source. These categories span a broad range of content, from macroeconomic indicators to labor statistics and market data. Each task is assigned a single category based on its data source, as listed below.

\begin{itemize}
  \item \textbf{Government:} National or international public-sector institutions that publish official statistics, policy data, tax rules, economic indicators, or demographic reports across sectors (e.g., World Bank, U.S. Census Bureau, OECD, Statistics Canada, IRS, IMF).

  \item \textbf{Finance:} Organizations or entities focused on investment products, corporate financials, fund performance, insurance costs, or regulatory financial filings (e.g., Vanguard, JPMorgan Chase, Amazon, ATTOM Data, SEC).

  \item \textbf{Markets:} Sources that report real-time or historical prices of financial instruments, commodities, indices, or cryptocurrencies, typically from exchanges or market aggregators (e.g., NASDAQ, Yahoo Finance, CoinMarketCap, London Stock Exchange, USAGOLD).

  \item \textbf{Labor:} Institutions that publish data on employment, unemployment, wages, workforce demographics, or job benefits at national or regional levels (e.g., U.S. Bureau of Labor Statistics, ILO, Texas Workforce Commission, U.S. Department of Labor).

  \item \textbf{Banking:} Central banks and monetary authorities that control interest rates, currency policy, money supply, and other macro-financial instruments (e.g., Federal Reserve, European Central Bank, Bank of England, Central Bank of Brazil).

  \item \textbf{Energy:} Organizations reporting on production, pricing, or consumption of energy sources such as oil, gas, or electricity (e.g., U.S. Energy Information Administration, OPEC).

  \item \textbf{Trade:} Sources focused on international commerce including import/export data, tariffs, and trade balances between nations (e.g., World Trade Organization, UN Comtrade, Observatory of Economic Complexity).

  \item \textbf{RealEstate:} Entities that report on housing prices, rents, real estate markets, or commercial property metrics (e.g., Zillow, CBRE, Cushman \& Wakefield, National Association of Realtors).

  \item \textbf{Education:} Sources providing data on tuition, school fees, academic costs, or university-affiliated expenses (e.g., College Board, Harvard Business School).

  \item \textbf{Health:} Agencies or organizations reporting health-related economic data such as premiums, public healthcare costs, or Medicare rates (e.g., U.S. Medicare, Kaiser Family Foundation).
\end{itemize}

\subsection{Website List}
\label{sec:website-list}

This appendix lists all websites included in the EconWebArena benchmark. For each site, we provide its URL and the associated category, as shown in Table~\ref{tab:websites}.

\begin{table*}[!h]
\centering
\scriptsize
\begin{tabular}{lll}
\toprule
\textbf{Website} &                         \textbf{URL} & \textbf{Category} \\
\midrule
ATTOM Data &            https://www.attomdata.com &           Finance \\
Amazon &           https://ir.aboutamazon.com &           Finance \\
Australian Bureau of Statistics &               https://www.abs.gov.au &        Government \\
Bank of England &      https://www.bankofengland.co.uk &           Banking \\
Brazilian Institute of Geography and Statistics &              https://www.ibge.gov.br &        Government \\
California Franchise Tax Board &               https://www.ftb.ca.gov &        Government \\
Canadian Real Estate Association &                https://stats.crea.ca &        RealEstate \\
Cayman Islands Monetary Authority &                  https://www.cima.ky &           Finance \\
Central Bank of Argentina &              https://www.bcra.gob.ar &           Banking \\
Central Bank of Brazil &               https://www.bcb.gov.br &           Banking \\
Central Bank of Egypt &               https://www.cbe.org.eg &           Banking \\
Central Bank of Mexico &           https://www.banxico.org.mx &           Banking \\
Central Bank of Nigeria &               https://www.cbn.gov.ng &           Banking \\
Central Bank of the Russian Federation &                   https://www.cbr.ru &           Banking \\
China National Bureau of Statistics &             https://www.stats.gov.cn &        Government \\
CoinMarketCap &            https://coinmarketcap.com &           Markets \\
Coldwell Banker Richard Ellis &                 https://www.cbre.com &        RealEstate \\
College Board &         https://www.collegeboard.org &         Education \\
Cox Automotive &           https://www.coxautoinc.com &           Markets \\
Cushman \& Wakefield &     https://www.cushmanwakefield.com &        RealEstate \\
Euronext &            https://live.euronext.com &           Markets \\
European Central Bank &            https://www.ecb.europa.eu &           Banking \\
Eurostat &                 https://ec.europa.eu &        Government \\
Federal Reserve &       https://www.federalreserve.gov &           Banking \\
Federal Reserve Bank of New York &           https://www.newyorkfed.org &           Banking \\
Federal Reserve Economic Data &          https://fred.stlouisfed.org &           Banking \\
France National Institute of Statistics and Economic Studies &                 https://www.insee.fr &        Government \\
Frankfurt Stock Exchange &      https://www.boerse-frankfurt.de &           Markets \\
German Federal Statistical Office &              https://www.destatis.de &        Government \\
Harvard Business School &                  https://www.hbs.edu &         Education \\
Hellenic Statistical Authority &            https://www.statistics.gr &        Government \\
India Brand Equity Foundation &                 https://www.ibef.org &        Government \\
Inland Revenue Authority of Singapore &              https://www.iras.gov.sg &        Government \\
International Labour Organization &                  https://www.ilo.org &             Labor \\
International Monetary Fund &                  https://www.imf.org &        Government \\
JPMorgan Chase &        https://www.jpmorganchase.com &           Finance \\
Kaiser Family Foundation &                  https://www.kff.org &            Health \\
London Bullion Market Association &              https://www.lbma.org.uk &           Markets \\
London Stock Exchange &  https://www.londonstockexchange.com &           Markets \\
NASDAQ &               https://www.nasdaq.com &           Markets \\
National Association of Realtors &              https://www.nar.realtor &        RealEstate \\
New York Department of Labor &                   https://dol.ny.gov &             Labor \\
New York Stock Exchange &                 https://www.nyse.com &           Markets \\
Observatory of Economic Complexity &                    https://oec.world &             Trade \\
Organization for Economic Co-operation and Development &                 https://www.oecd.org &        Government \\
Organization of the Petroleum Exporting Countries &                 https://www.opec.org &            Energy \\
Pew Research Center &          https://www.pewresearch.org &        Government \\
Philippine Statistics Authority &                   https://psa.gov.ph &        Government \\
Reserve Bank of New Zealand &             https://www.rbnz.govt.nz &           Banking \\
Saudi Arabia General Authority for Statistics &             https://www.stats.gov.sa &        Government \\
Saudi Central Bank &              https://www.sama.gov.sa &           Banking \\
South African Reserve Bank &            https://www.resbank.co.za &           Banking \\
Spanish Statistical Office &                   https://www.ine.es &        Government \\
State Bank of Pakistan &               https://www.sbp.org.pk &           Banking \\
Statistics Canada &            https://www.statcan.gc.ca &        Government \\
Statistics Korea &                 https://kostat.go.kr &        Government \\
Statistics Sweden &                   https://www.scb.se &        Government \\
The Wall Street Journal &                  https://www.wsj.com &           Markets \\
Turkish Statistical Institute &              https://www.tuik.gov.tr &        Government \\
U.K. Office for National Statistics &               https://www.ons.gov.uk &        Government \\
U.S. Bureau of Economic Analysis &                  https://www.bea.gov &        Government \\
U.S. Bureau of Labor Statistics &                  https://www.bls.gov &             Labor \\
U.S. Census Bureau &               https://www.census.gov &        Government \\
U.S. Citizenship and Immigration Services &                https://www.uscis.gov &        Government \\
U.S. Department of Housing and Urban Development &              https://www.huduser.gov &        Government \\
U.S. Department of Labor &                  https://www.dol.gov &             Labor \\
U.S. Department of the Treasury &            https://home.treasury.gov &        Government \\
U.S. Energy Information Administration &                  https://www.eia.gov &            Energy \\
U.S. Internal Revenue Service &                  https://www.irs.gov &        Government \\
U.S. Medicare &             https://www.medicare.gov &            Health \\
U.S. Securities and Exchange Commission &                  https://www.sec.gov &           Finance \\
U.S. Small Business Administration &                  https://www.sba.gov &           Finance \\
U.S. Social Security Administration &                  https://www.ssa.gov &        Government \\
USAGOLD &              https://www.usagold.com &           Markets \\
United Nations Comtrade Database &          https://comtradeplus.un.org &             Trade \\
Vanguard &        https://investor.vanguard.com &           Finance \\
Westmetall &           https://www.westmetall.com &           Markets \\
World Bank &            https://www.worldbank.org &        Government \\
World Inequality Database &                    https://wid.world &        Government \\
World Trade Organization &                  https://www.wto.org &             Trade \\
Yahoo Finance &            https://finance.yahoo.com &           Markets \\
Zillow &               https://www.zillow.com &        RealEstate \\
\bottomrule
\end{tabular}
\caption{List of websites used in EconWebArena, along with their URLs and categories.}
\label{tab:websites}
\end{table*}

\section{Experimental Details}

This appendix provides supplementary details on the experimental setup, evaluation configurations, and result breakdowns for EconWebArena.

\subsection{More Experimental Settings}
\label{sec:more-settings}

In this section, we provide additional configuration details for all experiments conducted on EconWebArena. Table~\ref{tab:settings} summarizes the prompt-level settings used by agents, adapted from the AgentLab \citep{chezelles2024browsergym} interface. Unless otherwise specified, these settings were fixed across all model evaluations. We evaluate a comprehensive set of LLMs, including GPT-4o\footnote{\texttt{gpt-4o-2024-08-06}: \url{https://platform.openai.com/docs/models/gpt-4o}} \citep{achiam2023gpt,openai2024gpt4o}, GPT-4o mini\footnote{\texttt{gpt-4o-mini-2024-07-18}: \url{https://platform.openai.com/docs/models/gpt-4o-mini}} \citep{openai2024gpt4omini}, GPT-4.1\footnote{\texttt{gpt-4.1-2025-04-14}: \url{https://platform.openai.com/docs/models/gpt-4.1}} \citep{openai2025gpt41}, GPT-4.1 mini\footnote{\texttt{gpt-4.1-mini-2025-04-14}: \url{https://platform.openai.com/docs/models/gpt-4.1-mini}} \citep{openai2025gpt41}, o4-mini\footnote{\texttt{o4-mini-2025-04-16}: \url{https://platform.openai.com/docs/models/o4-mini}} \citep{openai2025o4mini}, Claude Sonnet 4\footnote{\texttt{claude-sonnet-4-20250514}: \url{https://openrouter.ai/anthropic/claude-sonnet-4}} \citep{anthropic2025claude}, Gemini 2.5 Flash\footnote{\texttt{gemini-2.5-flash-preview-05-20}: \url{https://openrouter.ai/google/gemini-2.5-flash-preview-05-20}} \citep{team2023gemini,deepmind2025geminiflash}, and Llama 4 Maverick\footnote{\texttt{llama-4-maverick-17b-128e-instruct-fp8}: \url{https://openrouter.ai/meta-llama/llama-4-maverick}} \citep{grattafiori2024llama,meta2025llama4}.

\begin{table*}[!h]
\centering
\small
\begin{tabular}{>{\RaggedRight\arraybackslash}p{6em} >{\RaggedRight\arraybackslash}p{16.5em} >{\RaggedRight\arraybackslash}p{2.5em} >{\RaggedRight\arraybackslash}p{20em}}
\toprule
\textbf{Category} & \textbf{Flag} & \textbf{Setting} & \textbf{Description} \\
\midrule
& \texttt{use\_html} & \xmark & Include raw HTML content in the input prompt \\
& \texttt{use\_ax\_tree} & \cmark & Incorporate AXTree structure into the prompt \\
& \texttt{use\_focused\_element} & \cmark & Indicate the currently focused element \\
& \texttt{use\_error\_logs} & \cmark & Attach the last encountered error \\
& \texttt{use\_past\_error\_logs} & \xmark & Append historical error messages \\
& \texttt{use\_history} & \cmark & Add contextual history from the past \\
\texttt{observation} & \texttt{use\_action\_history} & \cmark & Provide a timeline of previous actions taken \\
& \texttt{use\_think\_history} & \cmark & Show prior reasoning steps (chain-of-thought) \\
& \texttt{use\_diff} & \xmark & Use image-based differences to represent changes \\
& \texttt{use\_screenshot} & \cmark & Use visual input by including a screenshot \\
& \texttt{use\_som} & \xmark & Replace screenshots with a Set-of-Marks format \\
& \texttt{extract\_visible\_tag} & \cmark & Tag elements that are visually present \\
& \texttt{extract\_clickable\_tag} & \cmark & Mark elements that are clickable \\
& \texttt{extract\_coords} & \cmark & Provide location data for each element \\
& \texttt{filter\_visible\_elements\_only} & \xmark & Restrict to only visible elements \\
\midrule
& \texttt{multiaction} & \xmark & Enable execution of multiple actions at once \\
\texttt{action} & \texttt{long\_description} & \xmark & Include full documentation for each action \\
& \texttt{individual\_examples} & \xmark & Add individual usage examples per action \\ 
\midrule
& \texttt{use\_thinking} & \cmark & Enable chain-of-thought reasoning \\
\texttt{reasoning} & \texttt{use\_plan} & \xmark & Let the agent draft and refine a plan at each step \\
& \texttt{use\_criticise} & \xmark & Let the agent to critique its own action \\
& \texttt{use\_memory} & \xmark & Retrieve and apply long-term memory \\
& \texttt{use\_concrete\_example} & \cmark & Utilize specific real-world examples for guidance \\
& \texttt{use\_abstract\_example} & \cmark & Use generalized, descriptive examples\\
& \texttt{use\_hints} & \cmark & Provide additional contextual hints to the model \\
& \texttt{enable\_chat} & \xmark & Allow multi-turn conversational interactions \\
& \texttt{be\_cautious} & \cmark & Promote a more conservative response style \\
& \texttt{extra\_instructions} & \xmark & Supplement with extra task-specific guidance \\
\bottomrule
\end{tabular}
\caption{Prompt configuration options used in EconWebArena, based on available flags and settings from AgentLab.}
\label{tab:settings}
\end{table*}

\subsection{More Experimental Results}
\label{sec:more-results}

We provide detailed performance metrics by category and model in Table~\ref{tab:more-results}, complementing the summary results reported in the main paper.

\begin{table*}[!h]
\small
\centering
\begin{tabular}{lr|rrrrrrrr|r}
\toprule
\textbf{Category} & \textbf{Tasks} & \textbf{o4-mini} & \textbf{GPT-4.1} & \textbf{-mini} & \textbf{GPT-4o} & \textbf{-mini} & \textbf{Claude} & \textbf{Gemini} & \textbf{Llama} & \textbf{Human} \\
\midrule
Banking & 60 & 41.7\% & 23.3\% & 15.0\% & 18.3\% & 15.0\% & 38.3\% & 28.3\% & 21.7\% & 95.0\% \\
Education & 6 & 50.0\% & 50.0\% & 50.0\% & 50.0\% & 33.3\% & 50.0\% & 50.0\% & 0.0\% & 100.0\% \\
Energy & 18 & 27.8\% & 5.6\% & 27.8\% & 5.6\% & 0.0\% & 44.4\% & 11.1\% & 11.1\% & 100.0\% \\
Finance & 21 & 33.3\% & 14.3\% & 19.0\% & 14.3\% & 19.0\% & 23.8\% & 33.3\% & 9.5\% & 95.2\% \\
Government & 138 & 57.2\% & 45.7\% & 34.1\% & 35.5\% & 12.3\% & 47.1\% & 39.1\% & 26.1\% & 91.3\% \\
Health & 6 & 100.0\% & 100.0\% & 66.7\% & 100.0\% & 16.7\% & 16.7\% & 50.0\% & 33.3\% & 100.0\% \\
Labor & 24 & 20.8\% & 0.0\% & 4.2\% & 8.3\% & 0.0\% & 12.5\% & 4.2\% & 4.2\% & 91.7\% \\
Markets & 60 & 48.3\% & 35.0\% & 35.0\% & 33.3\% & 6.7\% & 41.7\% & 33.3\% & 15.0\% & 96.7\% \\
RealEstate & 15 & 13.3\% & 0.0\% & 6.7\% & 0.0\% & 0.0\% & 0.0\% & 0.0\% & 0.0\% & 93.3\% \\
Trade & 12 & 66.7\% & 33.3\% & 33.3\% & 16.7\% & 0.0\% & 50.0\% & 41.7\% & 25.0\% & 75.0\% \\
\midrule
All SR ($\uparrow$) & 360 & 46.9\% & 31.9\% & 27.5\% & 26.9\% & 10.3\% & 38.6\% & 31.1\% & 18.9\% & 93.3\% \\
\midrule
Steps ($\downarrow$) & - & 8.99 & 7.23 & 8.7 & 7.77 & 8.7 & 11.77 & 9.29 & 9.54 & - \\
\bottomrule
\end{tabular}
\caption{Detailed success rates (SR) and average steps (on successful tasks) on EconWebArena by category for models (o4-mini, GPT-4.1, GPT-4.1 mini, GPT-4o, GPT-4o mini, Claude Sonnet 4, Gemini 2.5 Flash, Llama 4 Maverick) and human.}
\label{tab:more-results}
\end{table*}

\subsection{Successful Cases}
\label{sec:success-cases}

\begin{figure*}[!h]
    \centering
    \setlength\fboxsep{1pt}
    \setlength\fboxrule{0.5pt}

    \begin{subfigure}[t]{0.32\textwidth}
        \centering
        \fbox{\includegraphics[width=\linewidth]{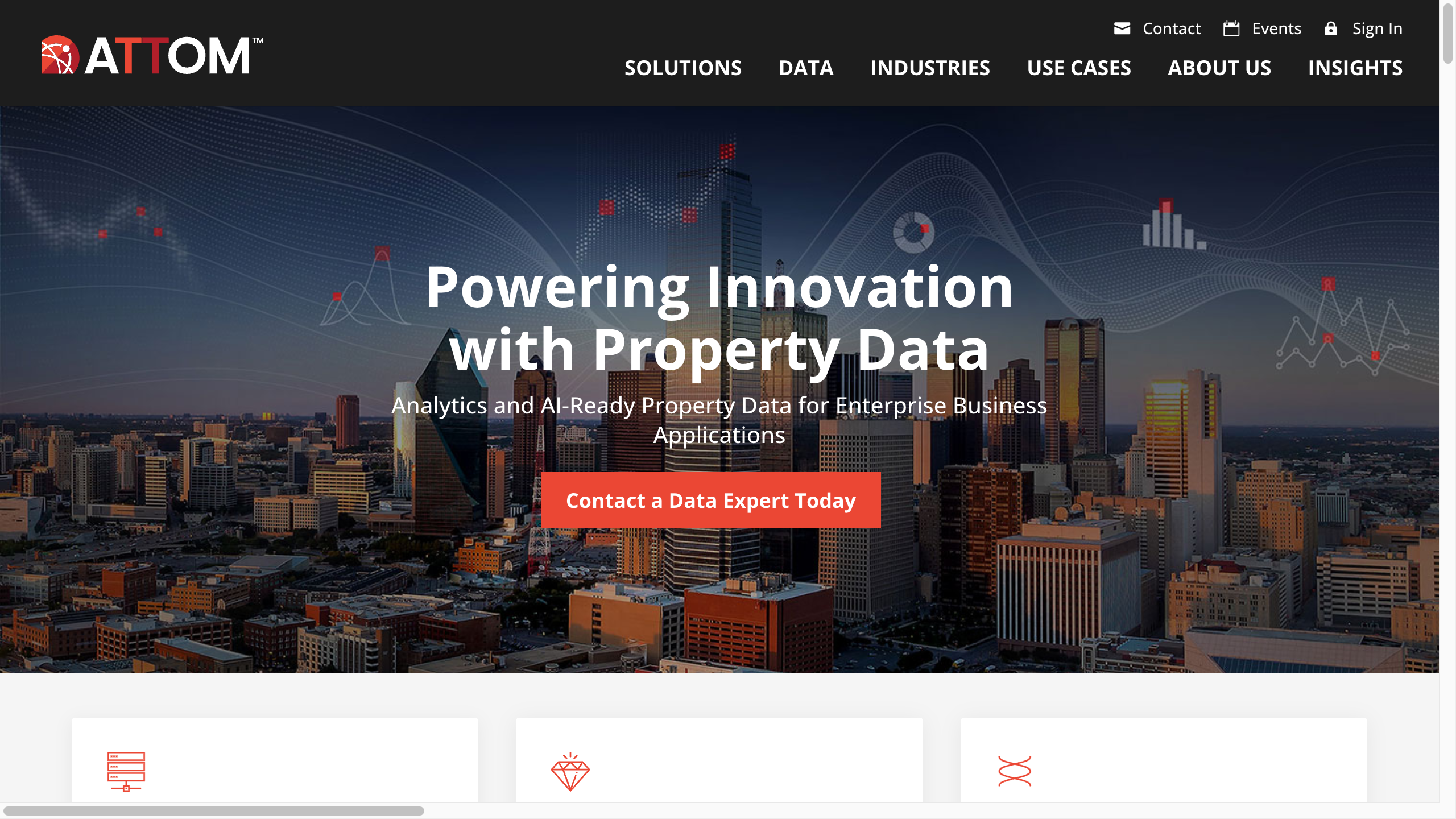}}
        \caption{Step 0: Open ATTOM website}
    \end{subfigure}
    \hfill
    \begin{subfigure}[t]{0.32\textwidth}
        \centering
        \fbox{\includegraphics[width=\linewidth]{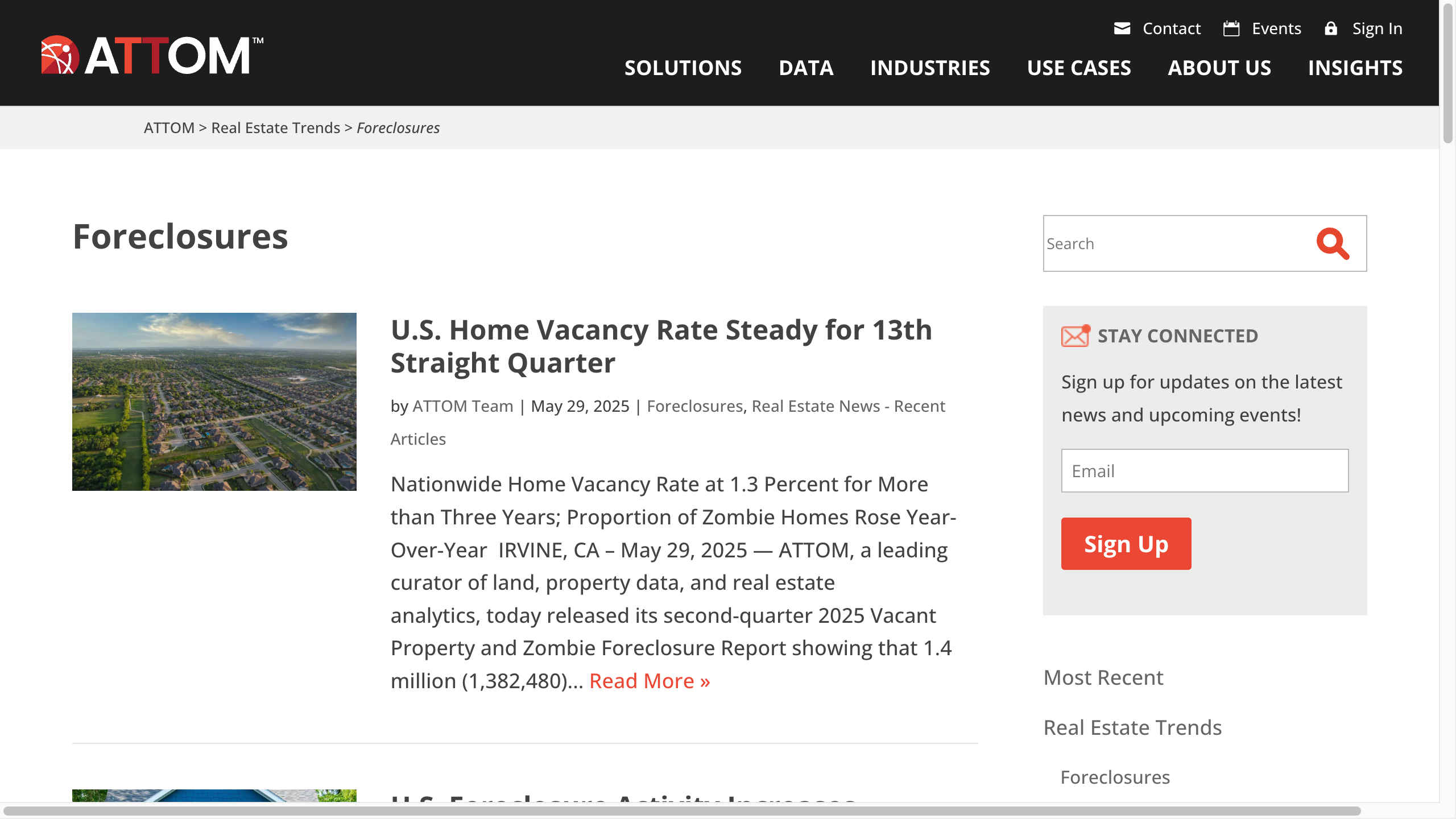}}
        \caption{Step 4: Navigate foreclosure reports}
    \end{subfigure}
    \hfill
    \begin{subfigure}[t]{0.32\textwidth}
        \centering
        \fbox{\includegraphics[width=\linewidth]{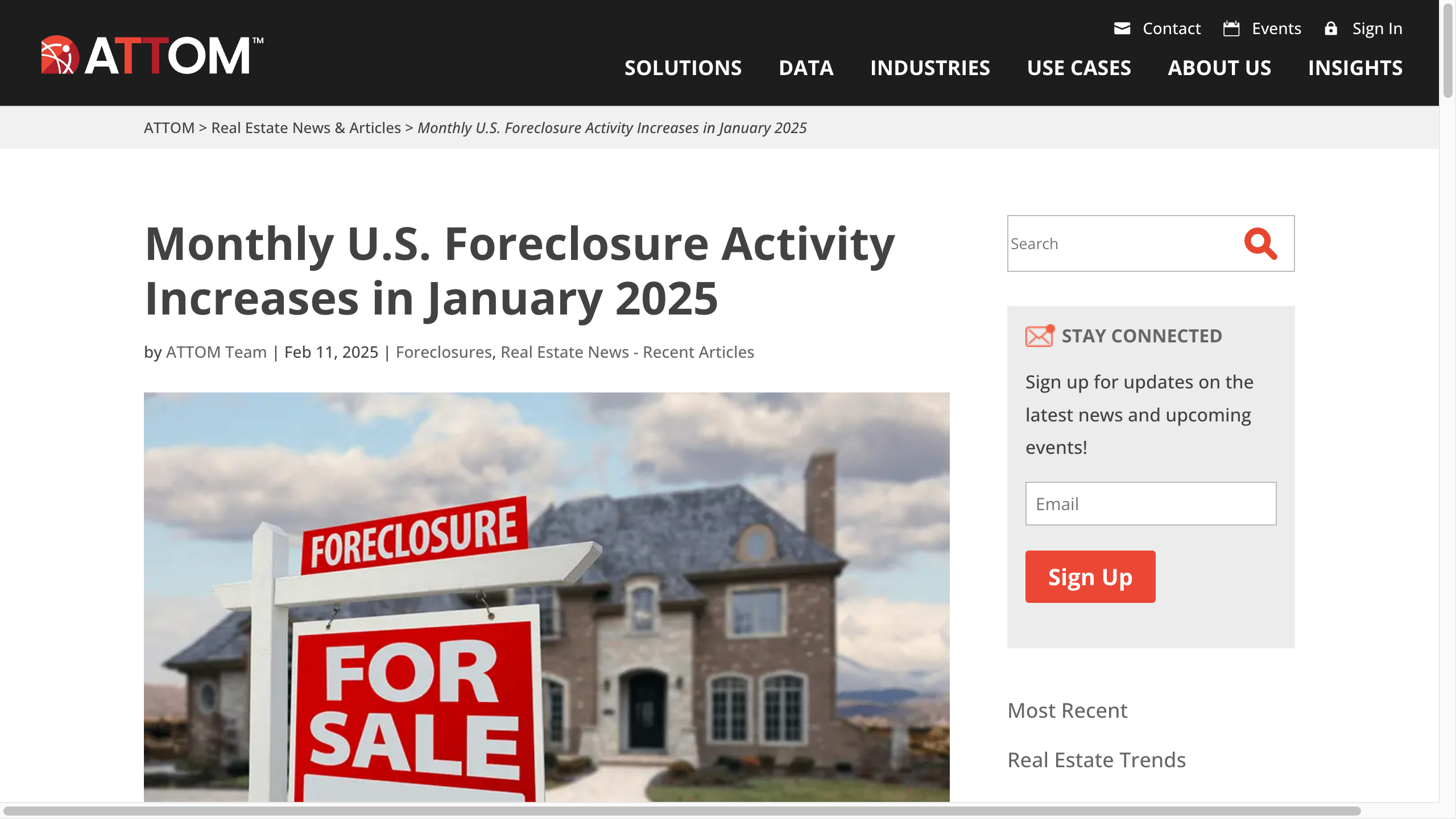}}
        \caption{Step 5: Extract the answer}
    \end{subfigure}

    \caption{Successful example of Task 5 with o4-mini, which retrieves Delaware's foreclosure rate from ATTOM Data Solutions for January 2025. The agent uses search, navigates through blog categories, and correctly locates the relevant report.}
    \label{fig:success-5}
\end{figure*}

\begin{figure*}[!h]
    \centering
    \setlength\fboxsep{1pt}
    \setlength\fboxrule{0.5pt}

    \begin{subfigure}[t]{0.32\textwidth}
        \centering
        \fbox{\includegraphics[width=\linewidth]{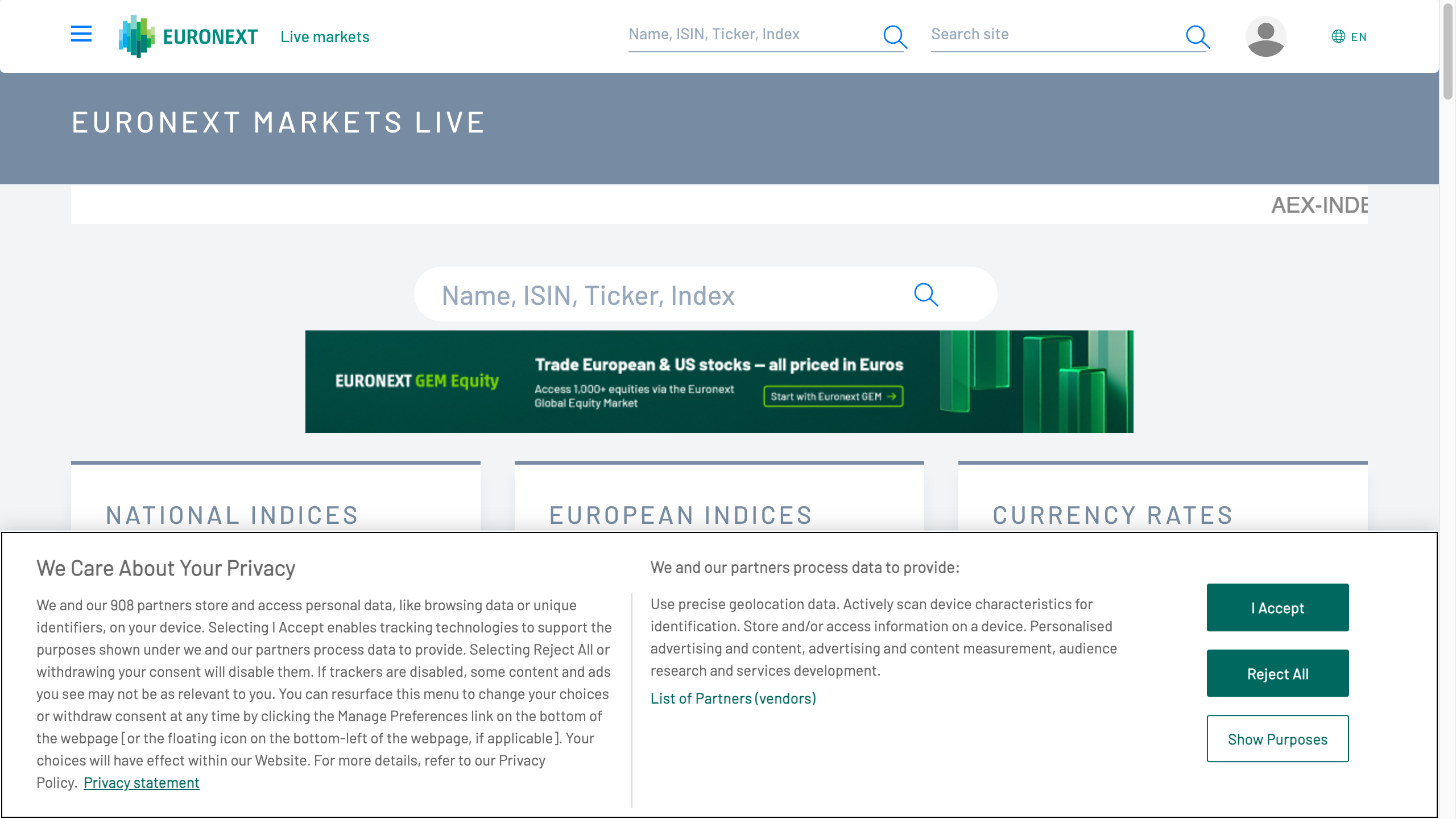}}
        \caption{Step 0: Open Euronext website}
    \end{subfigure}
    \hfill
    \begin{subfigure}[t]{0.32\textwidth}
        \centering
        \fbox{\includegraphics[width=\linewidth]{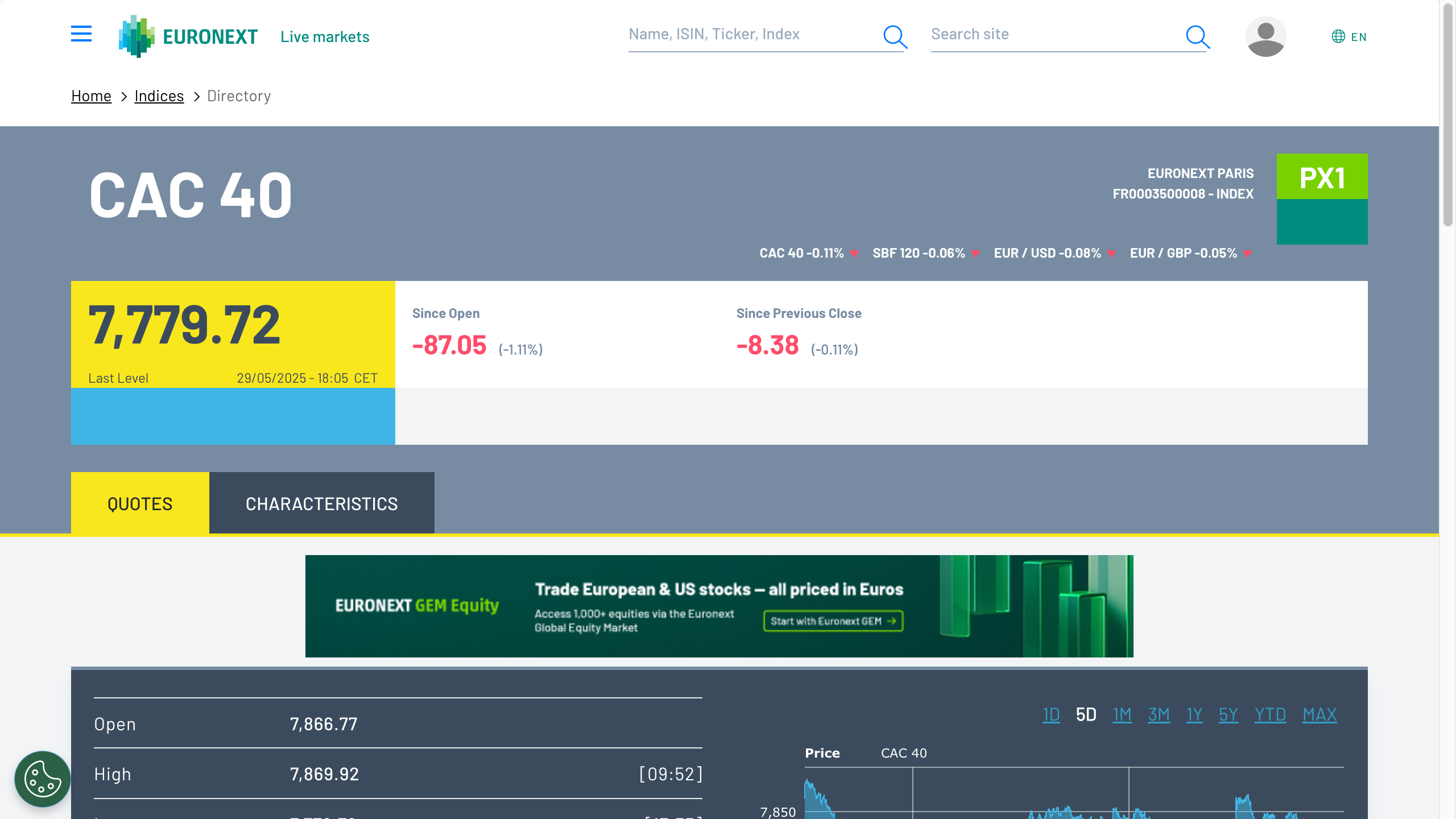}}
        \caption{Step 2: Navigate to CAC 40 page}
    \end{subfigure}
    \hfill
    \begin{subfigure}[t]{0.32\textwidth}
        \centering
        \fbox{\includegraphics[width=\linewidth]{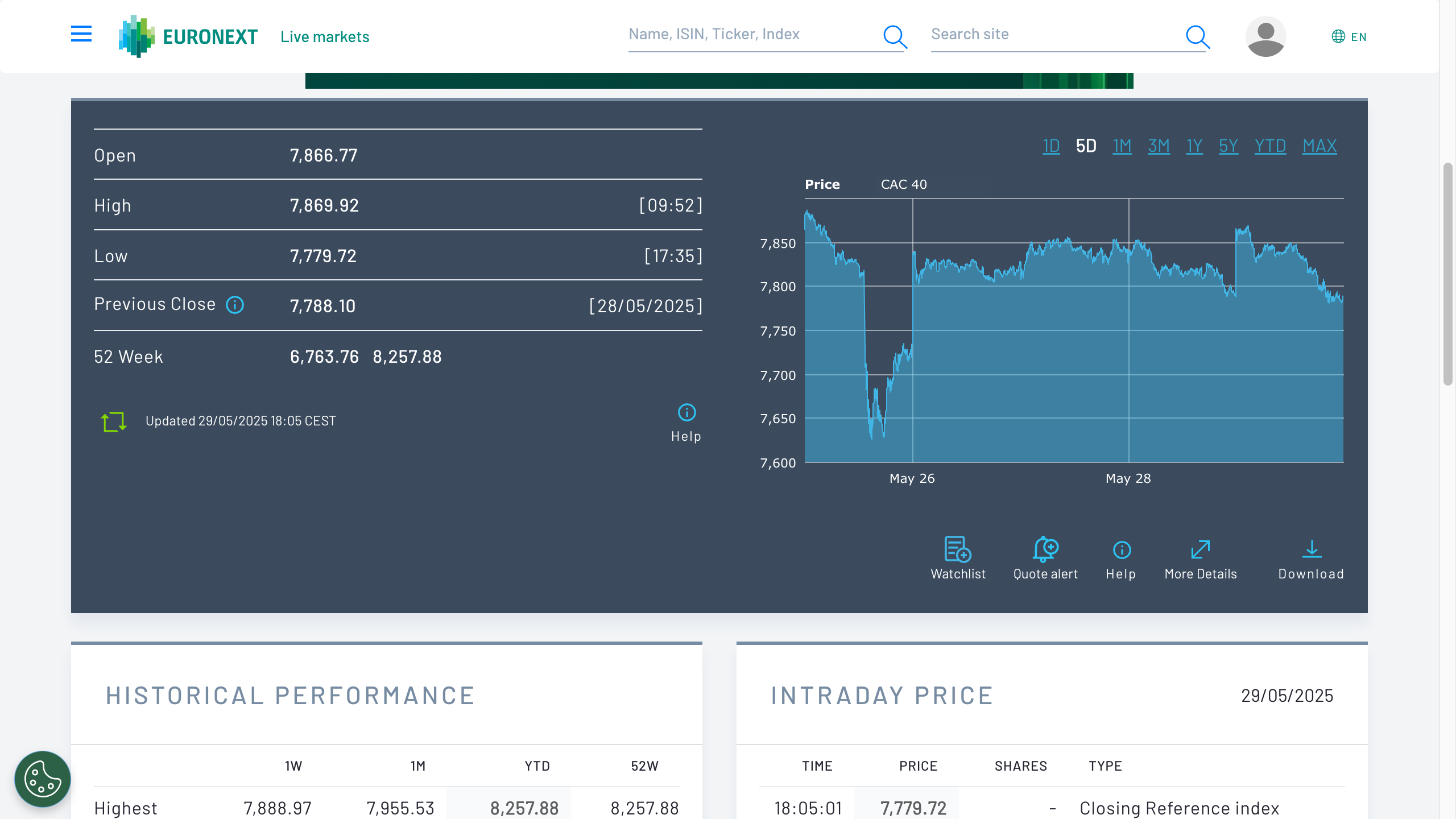}}
        \caption{Step 3: Browse the index page}
    \end{subfigure}

    \vspace{1em}

    \begin{subfigure}[t]{0.32\textwidth}
        \centering
        \fbox{\includegraphics[width=\linewidth]{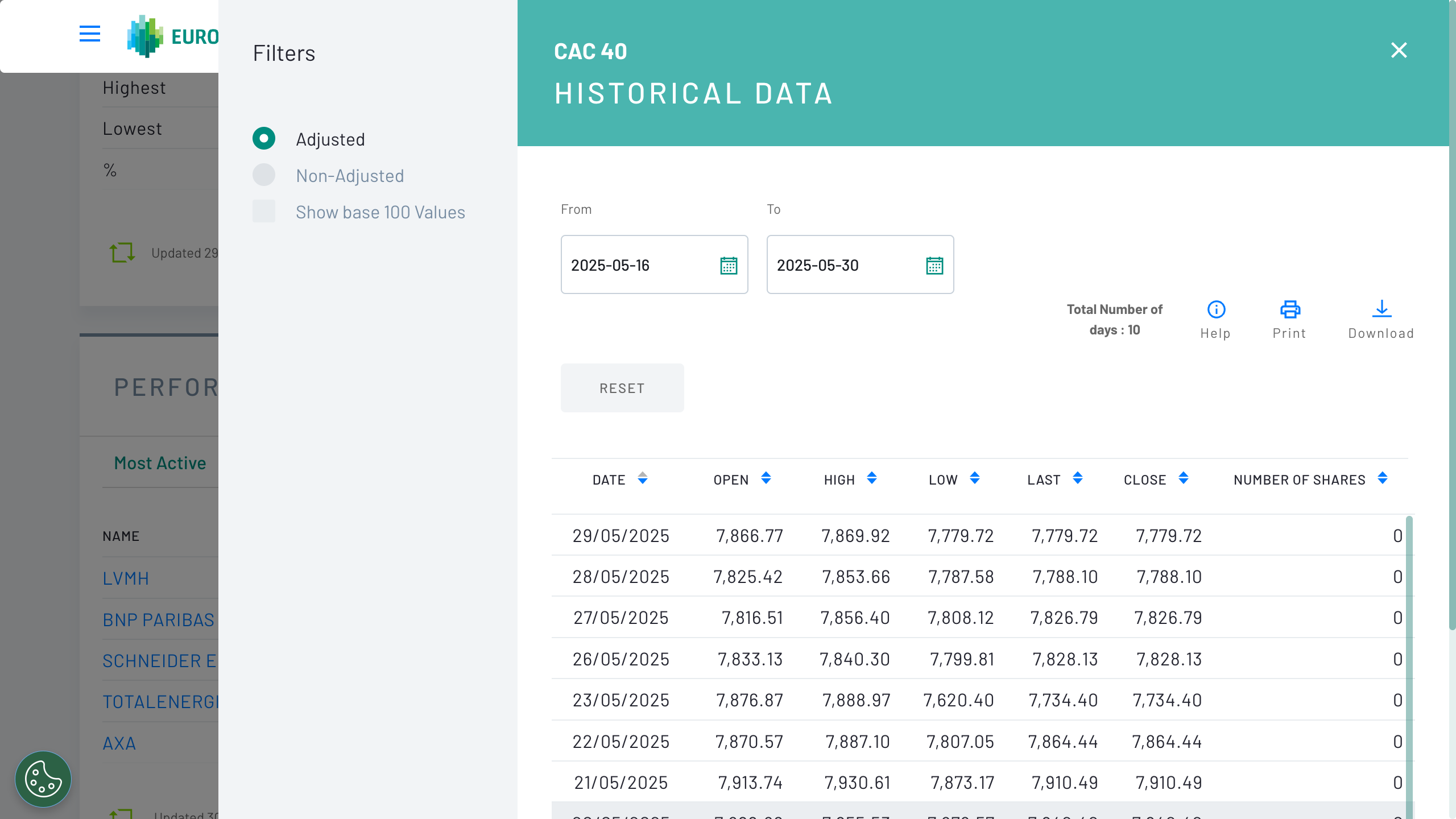}}
        \caption{Step 20: Locate historical data section}
    \end{subfigure}
    \hfill
    \begin{subfigure}[t]{0.32\textwidth}
        \centering
        \fbox{\includegraphics[width=\linewidth]{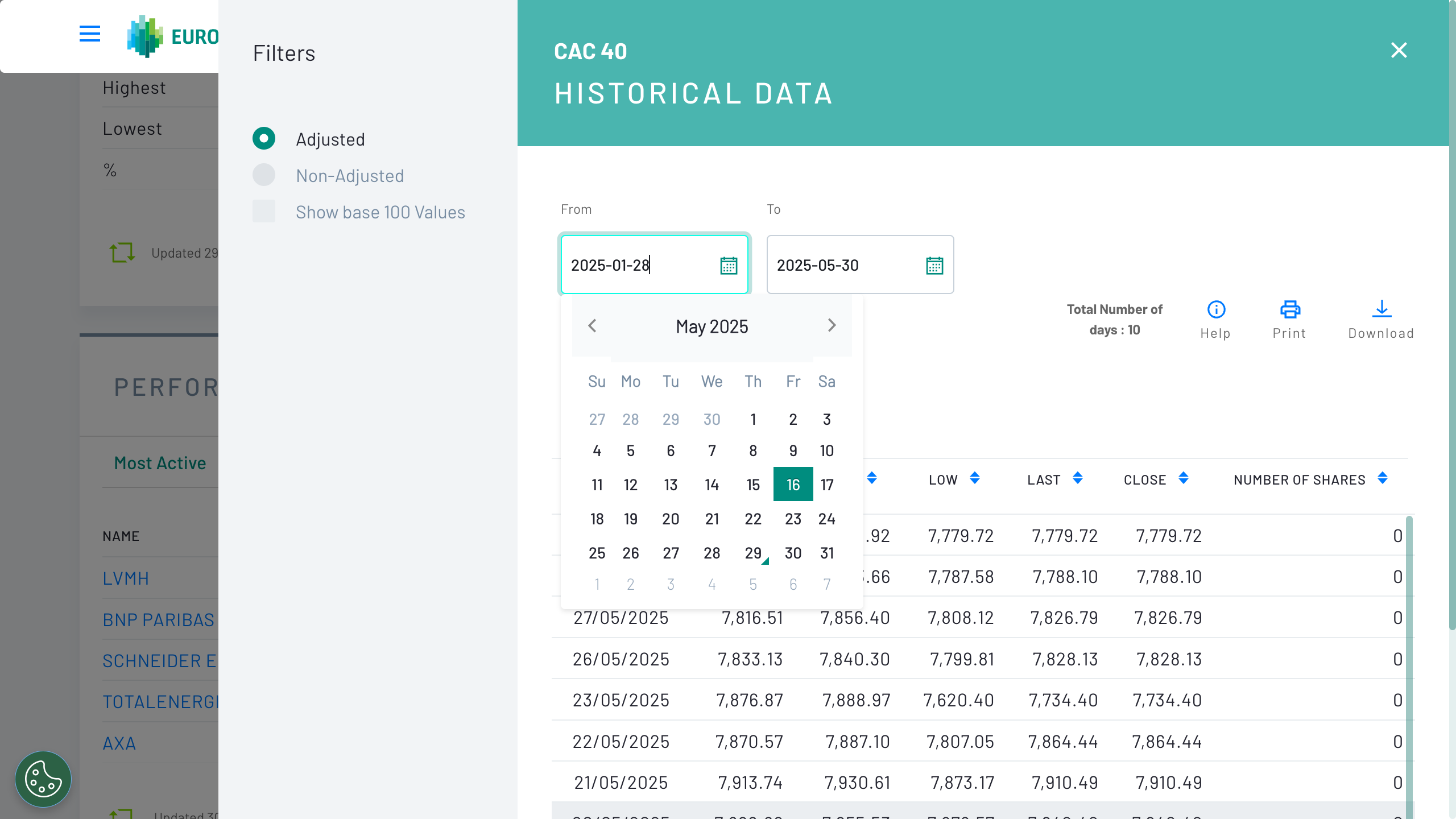}}
        \caption{Step 21: Set the target date}
    \end{subfigure}
    \hfill
    \begin{subfigure}[t]{0.32\textwidth}
        \centering
        \fbox{\includegraphics[width=\linewidth]{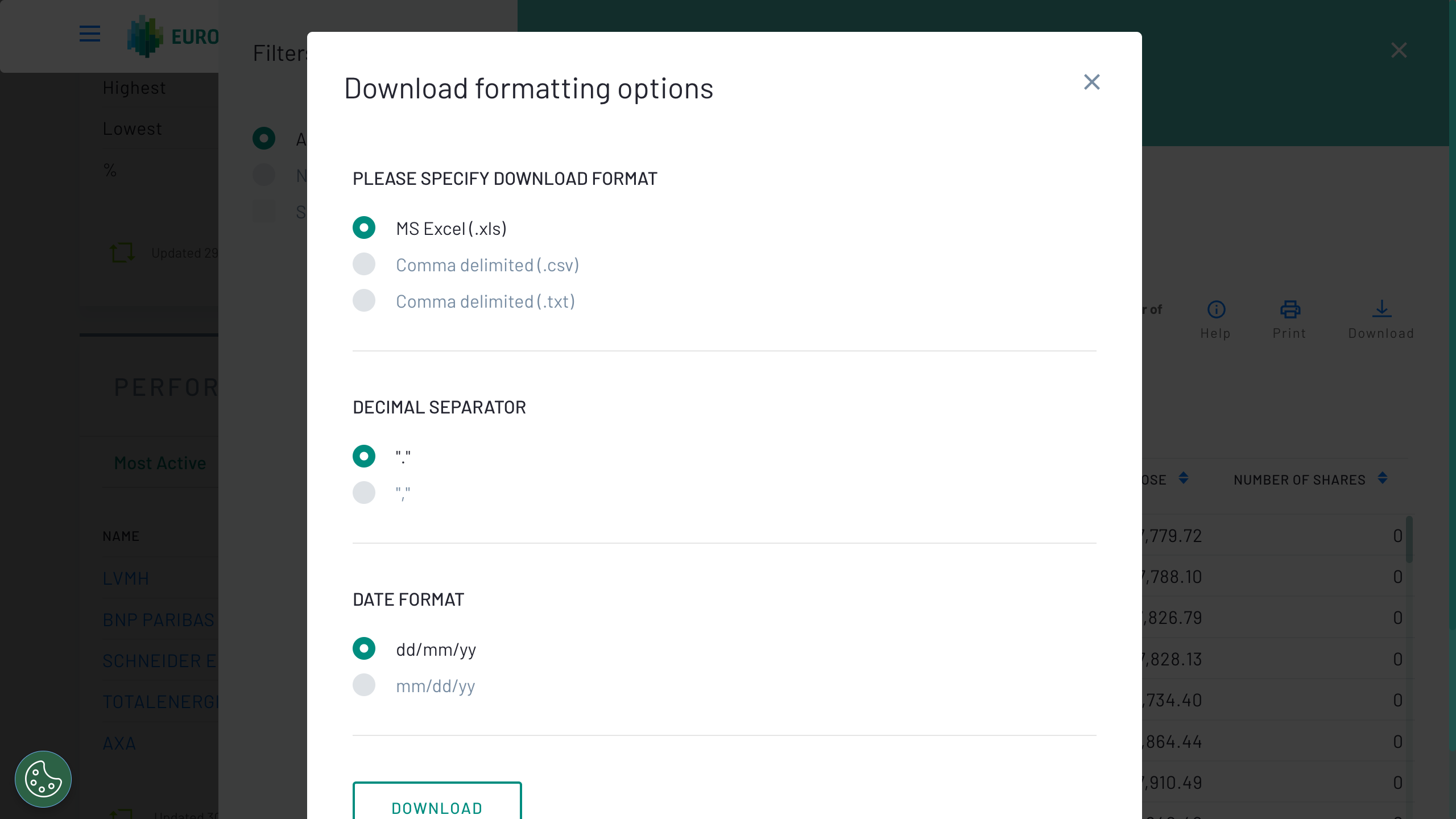}}
        \caption{Step 23: Output the closing value}
    \end{subfigure}

    \caption{Successful example of Task 66 with o4-mini, retrieving the CAC 40 closing value from Euronext on January 28, 2025. The agent correctly navigates to the index page, accesses historical data, inputs the date, and extracts the required number.}
    \label{fig:success-66}
\end{figure*}

\begin{figure*}[!h]
    \centering
    \setlength\fboxsep{1pt}
    \setlength\fboxrule{0.5pt}

    \begin{subfigure}[t]{0.32\textwidth}
        \centering
        \fbox{\includegraphics[width=\linewidth]{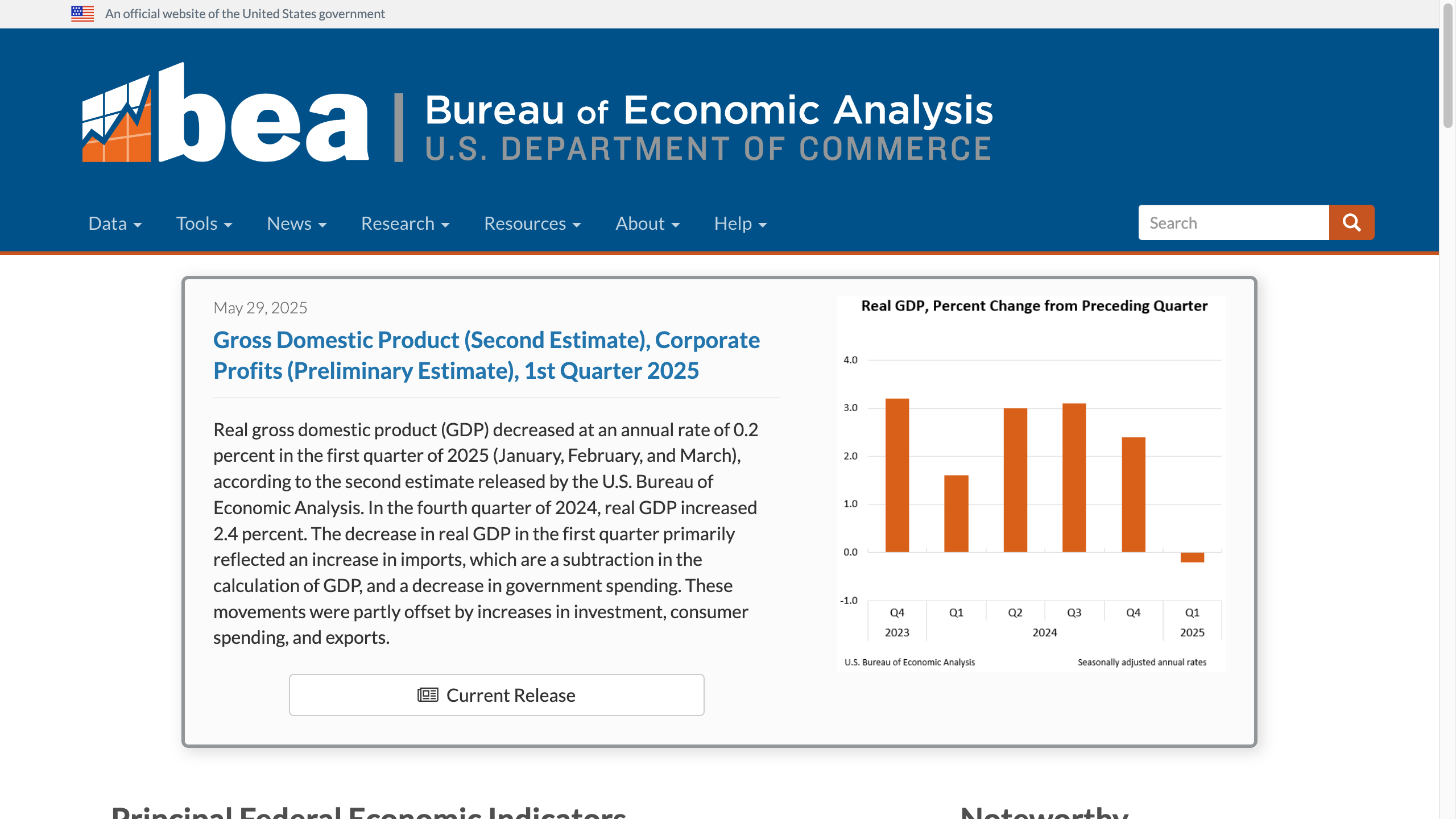}}
        \caption{Step 0: Navigate to recent releases}
    \end{subfigure}
    \hfill
    \begin{subfigure}[t]{0.32\textwidth}
        \centering
        \fbox{\includegraphics[width=\linewidth]{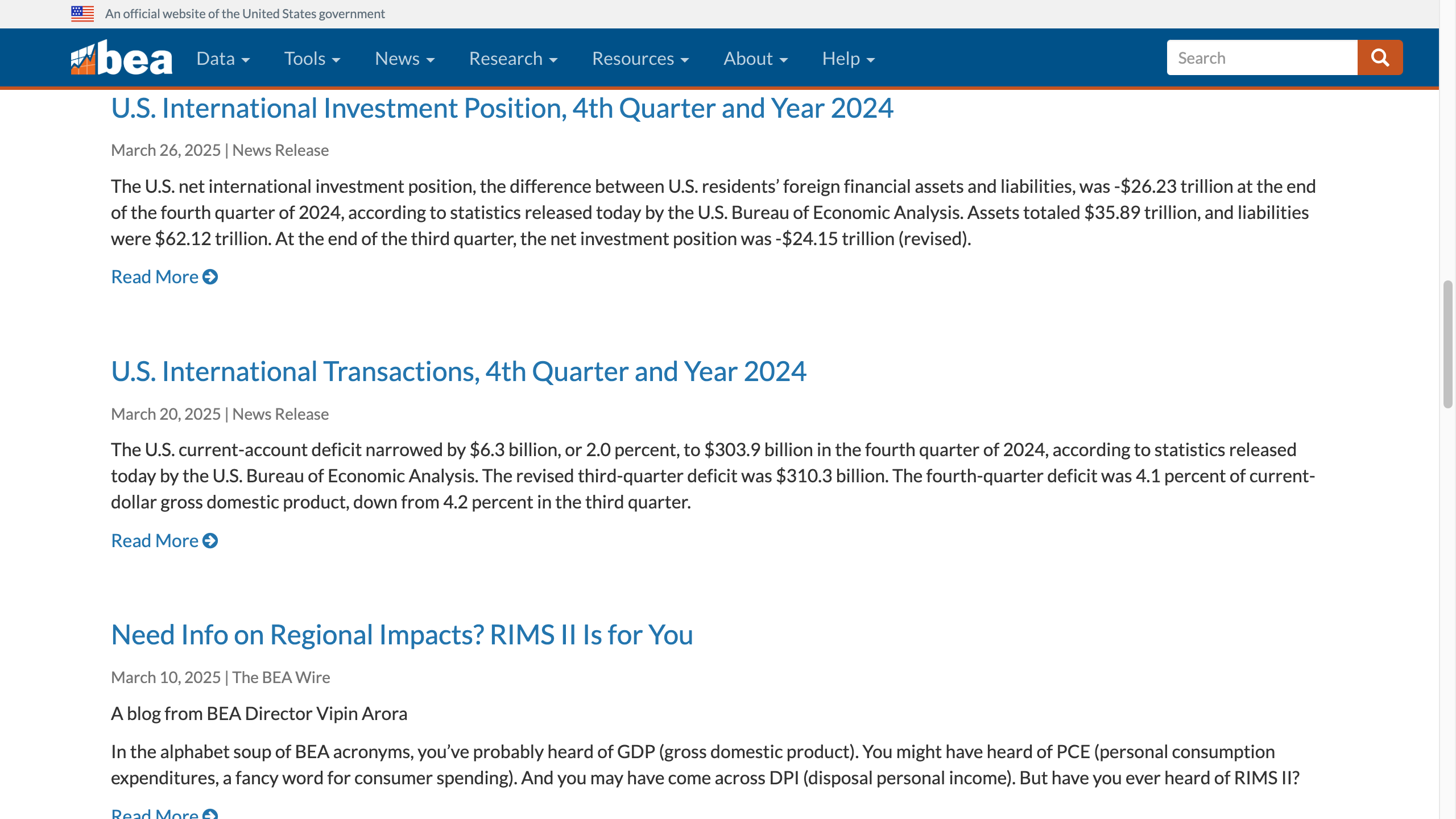}}
        \caption{Step 1: Open January trade report}
    \end{subfigure}
    \hfill
    \begin{subfigure}[t]{0.32\textwidth}
        \centering
        \fbox{\includegraphics[width=\linewidth]{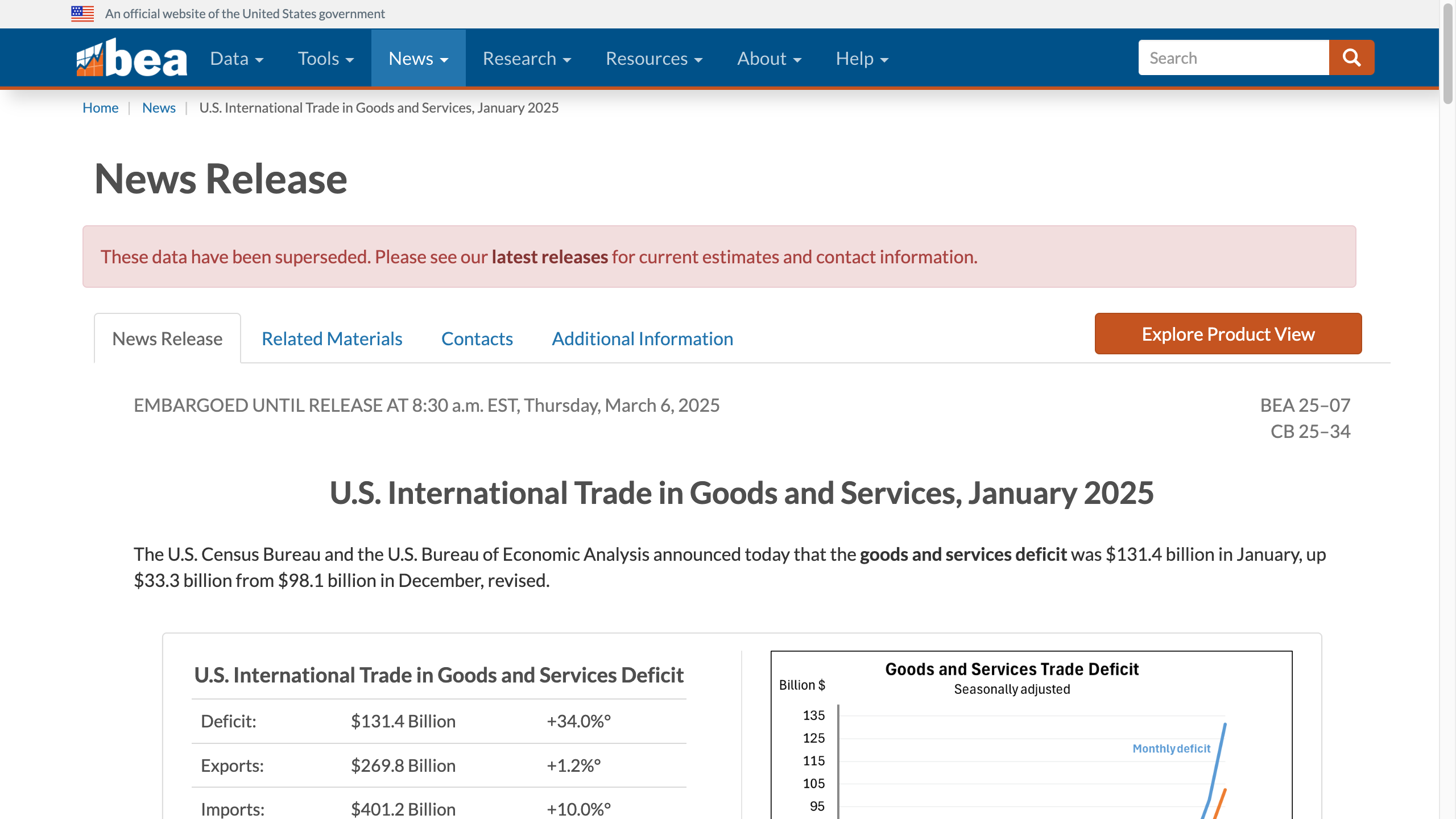}}
        \caption{Step 2: Extract the import value}
    \end{subfigure}

    \caption{Successful example of Task 237 with o4-mini, retrieving U.S. imports for January 2025 from the Bureau of Economic Analysis. The agent correctly navigates to older releases, opens the January report, and extracts the reported trade imports value.}
    \label{fig:success-237}
\end{figure*}

\begin{figure*}[!h]
    \centering
    \setlength\fboxsep{1pt}
    \setlength\fboxrule{0.5pt}

    \begin{subfigure}[t]{0.32\textwidth}
        \centering
        \fbox{\includegraphics[width=\linewidth]{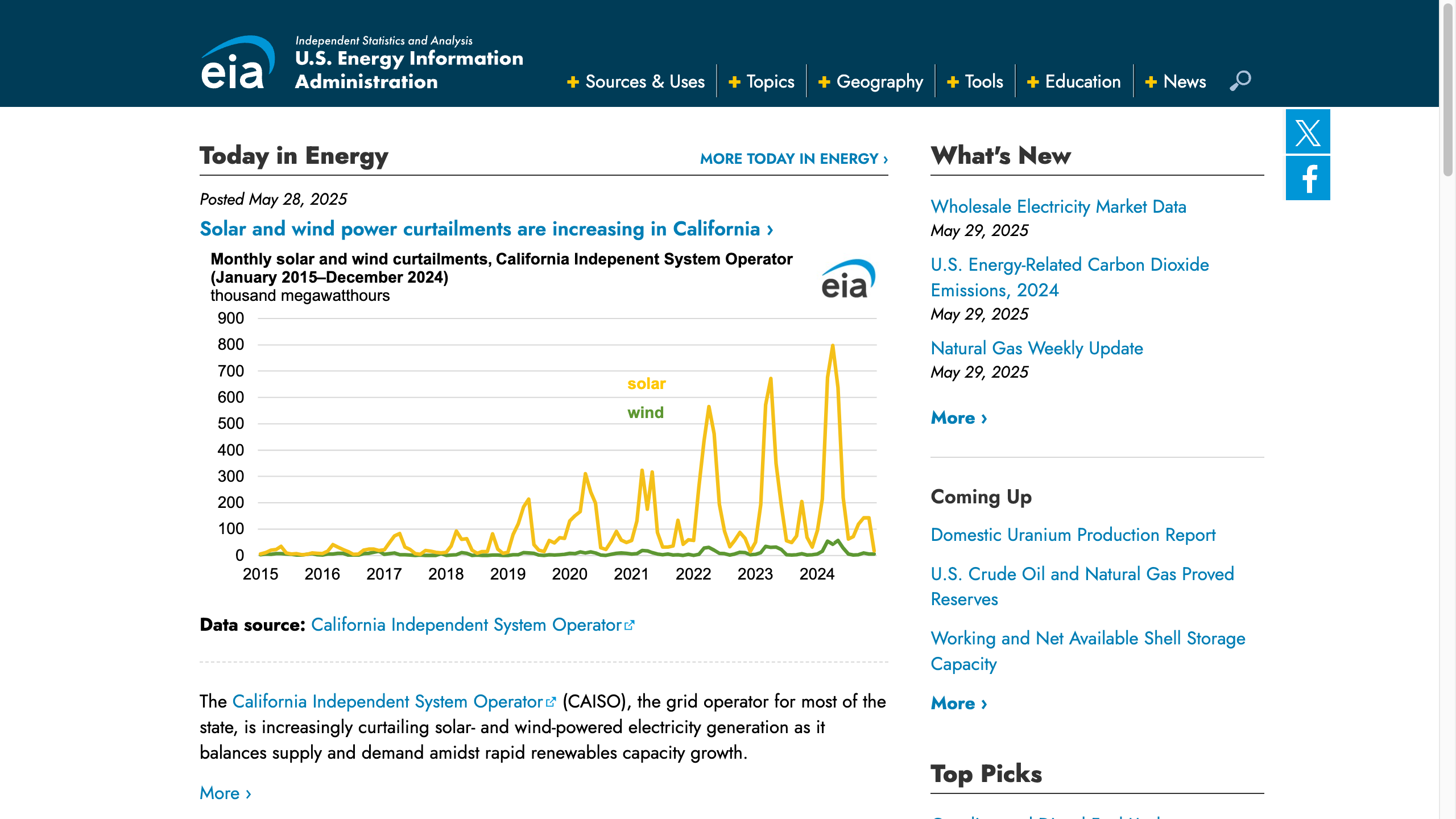}}
        \caption{Step 0: Open EIA homepage}
    \end{subfigure}
    \hfill
    \begin{subfigure}[t]{0.32\textwidth}
        \centering
        \fbox{\includegraphics[width=\linewidth]{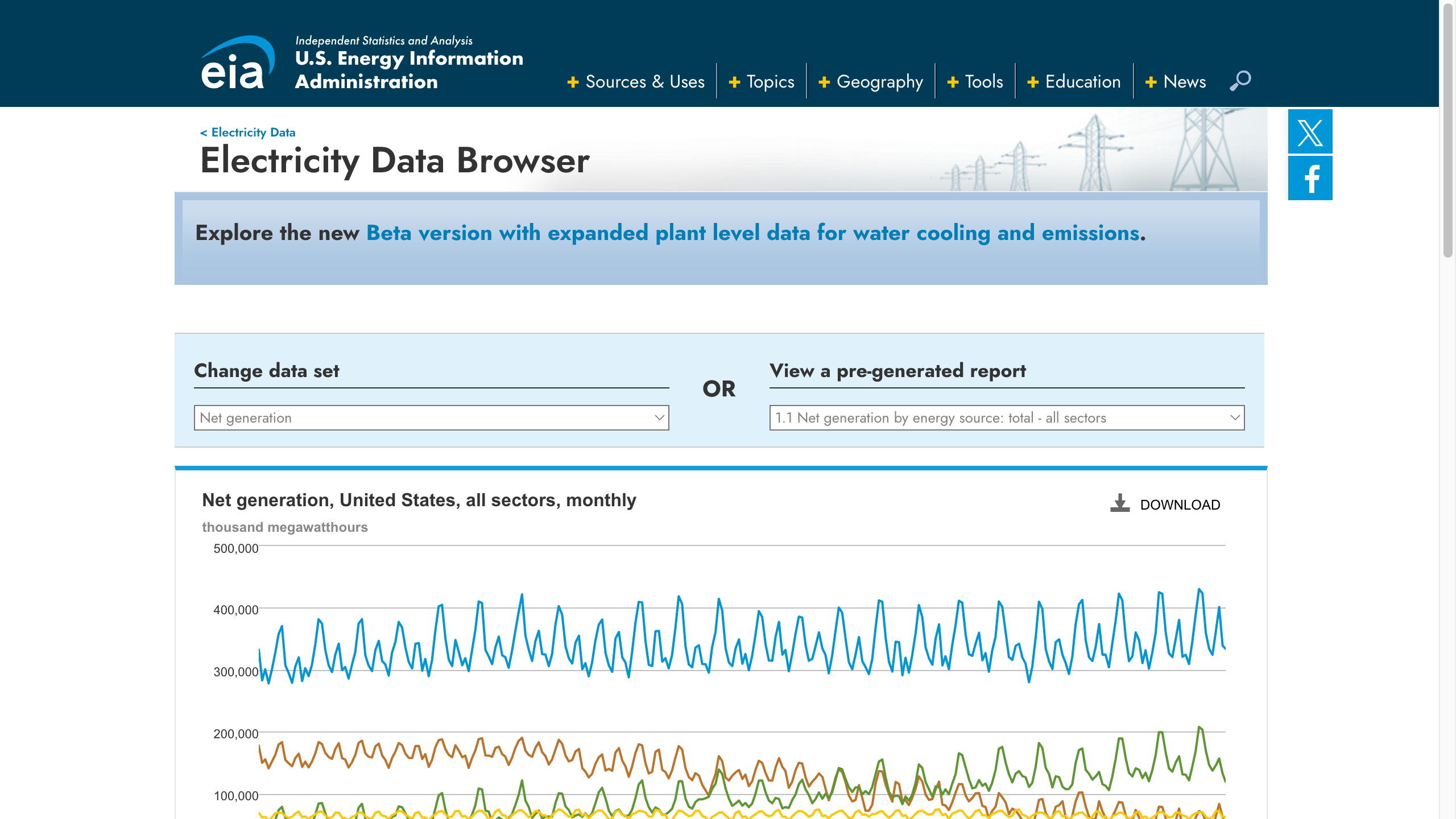}}
        \caption{Step 4: Navigate to electricity data}
    \end{subfigure}
    \hfill
    \begin{subfigure}[t]{0.32\textwidth}
        \centering
        \fbox{\includegraphics[width=\linewidth]{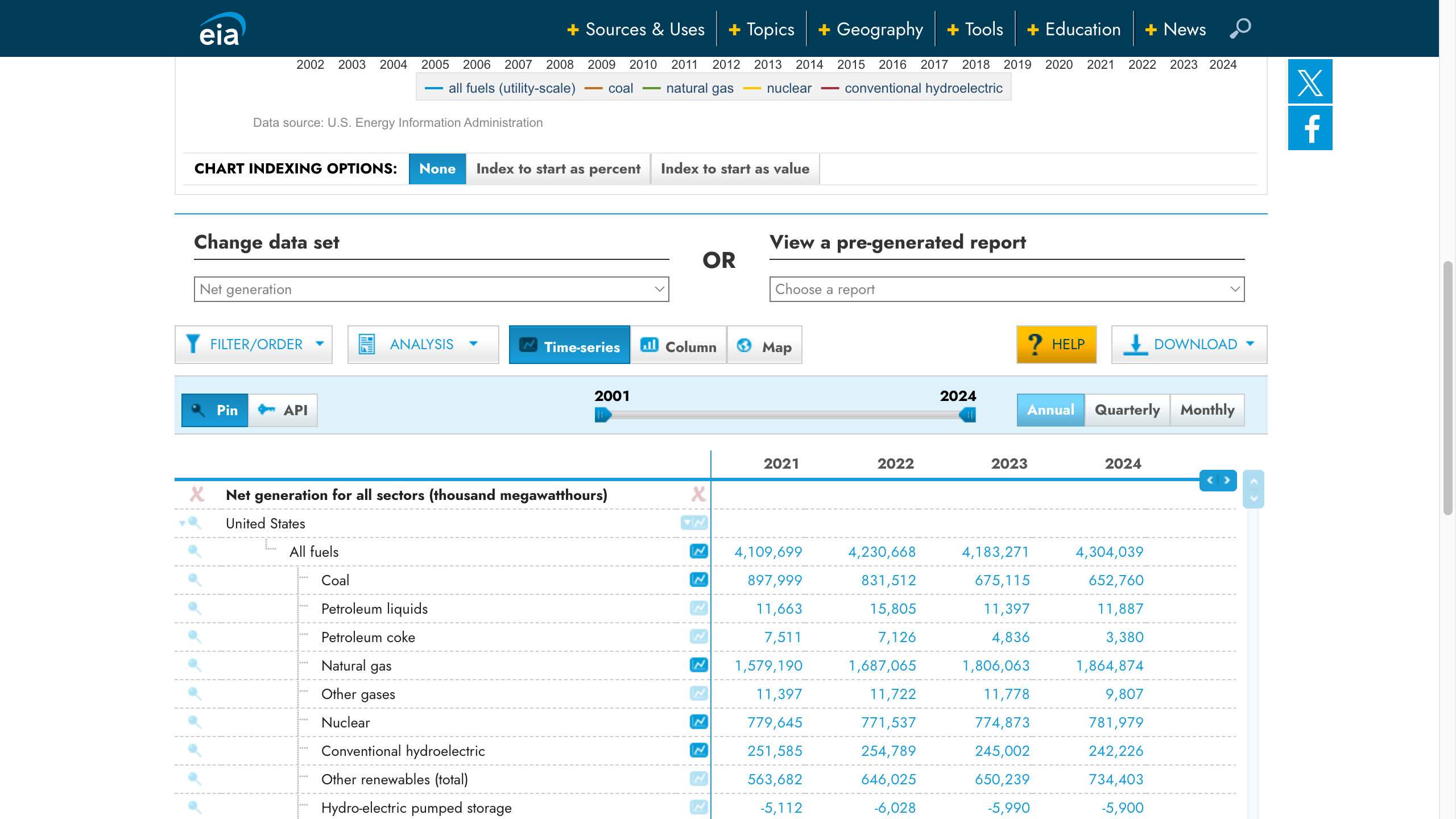}}
        \caption{Step 5: Extract the 2024 value}
    \end{subfigure}

    \caption{Successful example of Task 280 with o4-mini, which retrieves the total net electricity generation from natural gas in the United States for 2024 using the EIA Electricity Data Browser.}
    \label{fig:success-280}
\end{figure*}

To illustrate how LLM agents can succeed on complex economic web tasks, we present four representative success cases completed by the o4-mini model. These examples span diverse categories and interaction styles, including document navigation, structured table retrieval, search-based report access, and chart-driven queries. In each case, the agent accurately interprets the task, locates the appropriate webpage, handles user interface components such as dropdowns and filters, and extracts the correct numeric value. Figures~\ref{fig:success-5}-\ref{fig:success-280} demonstrate that while the benchmark poses many challenges, properly configured agents are capable of executing realistic workflows across a wide variety of sources.

\subsection{Error Cases}
\label{sec:error-cases}

\begin{figure*}[!h]
    \centering
    \setlength\fboxsep{1pt}
    \setlength\fboxrule{0.5pt}

    \begin{subfigure}[t]{0.32\textwidth}
        \centering
        \fbox{\includegraphics[width=\linewidth]{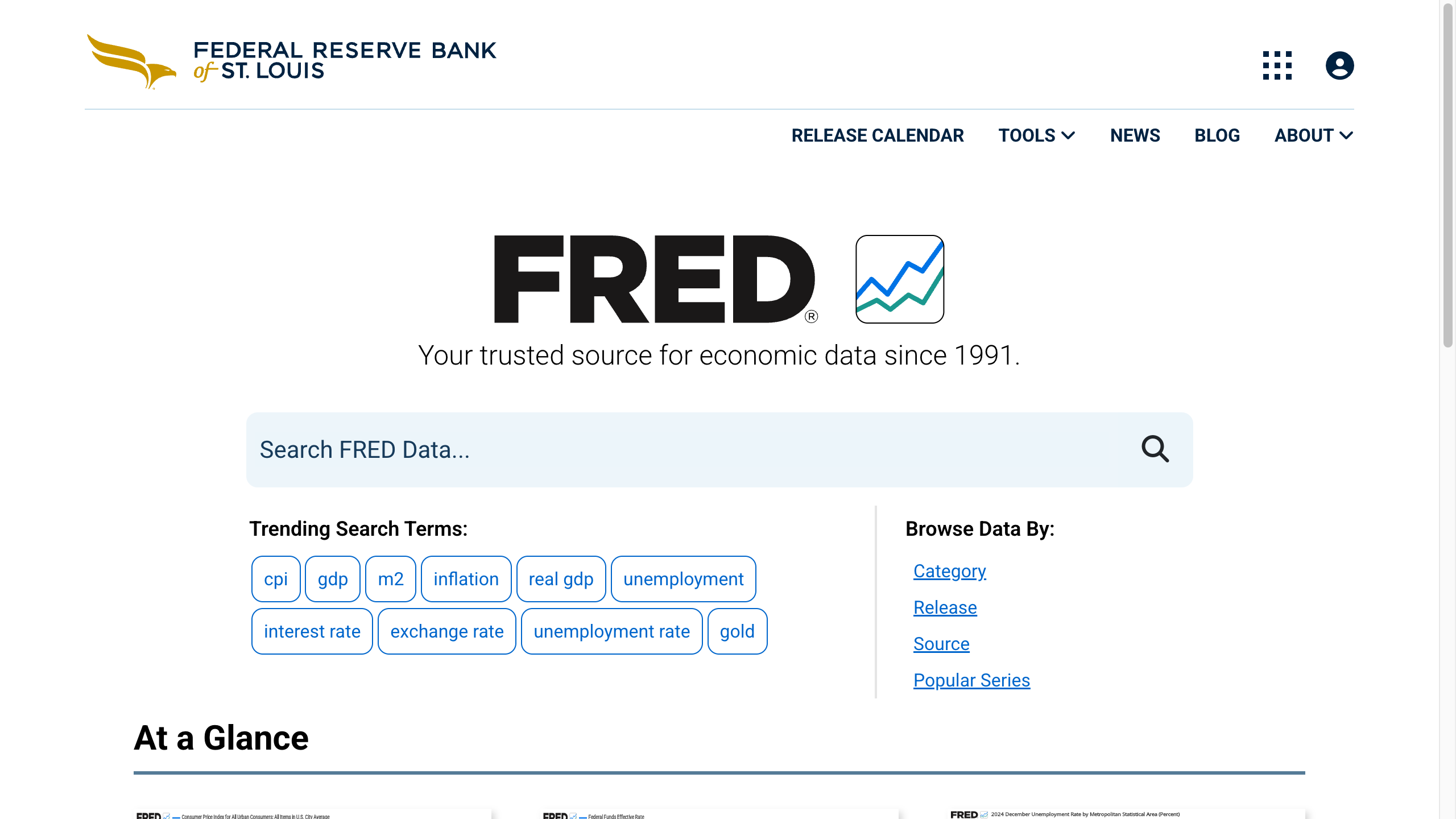}}
        \caption{Step 0: Open FRED website}
    \end{subfigure}
    \hfill
    \begin{subfigure}[t]{0.32\textwidth}
        \centering
        \fbox{\includegraphics[width=\linewidth]{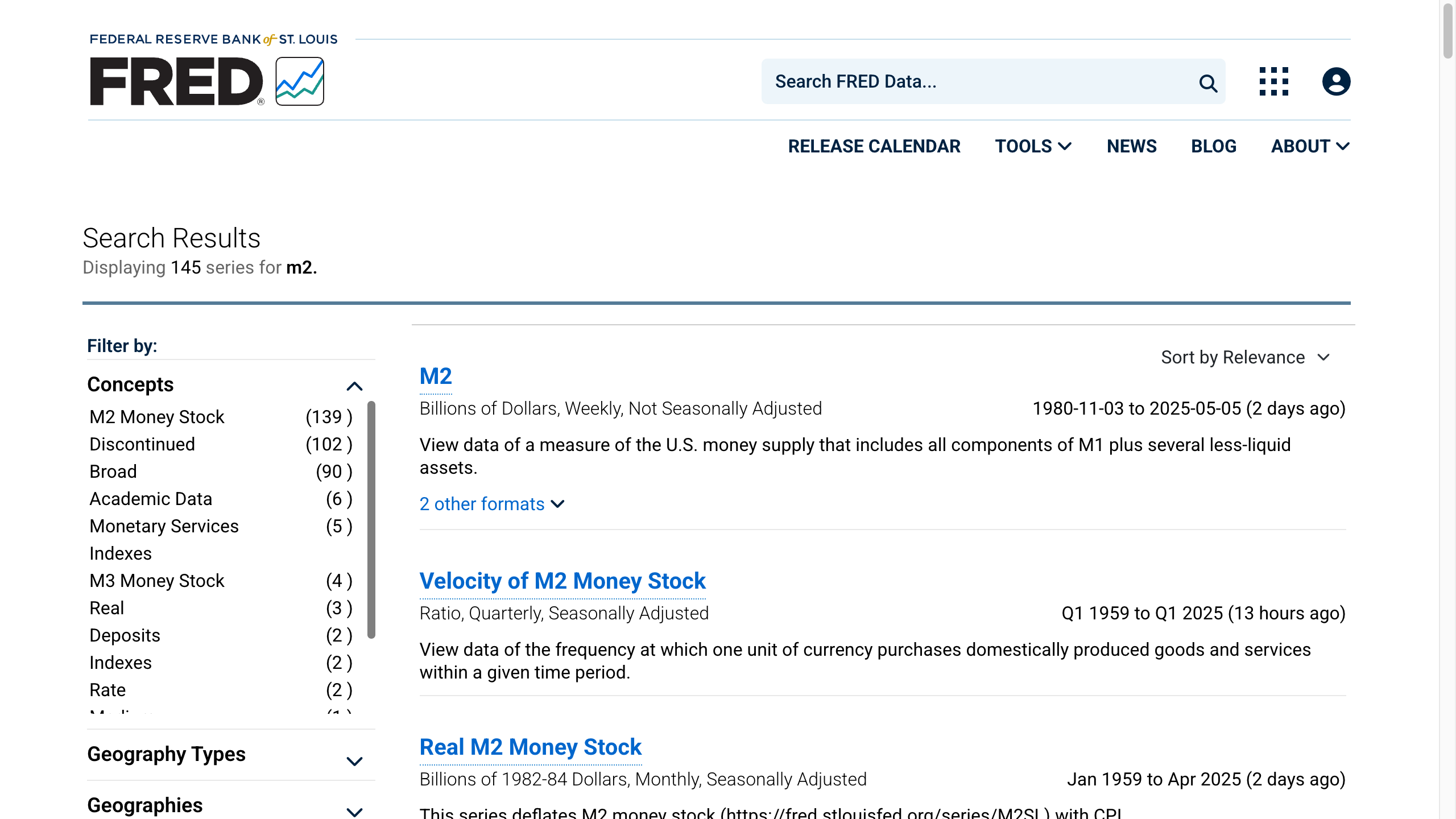}}
        \caption{Step 1: Search for M2}
    \end{subfigure}
    \hfill
    \begin{subfigure}[t]{0.32\textwidth}
        \centering
        \fbox{\includegraphics[width=\linewidth]{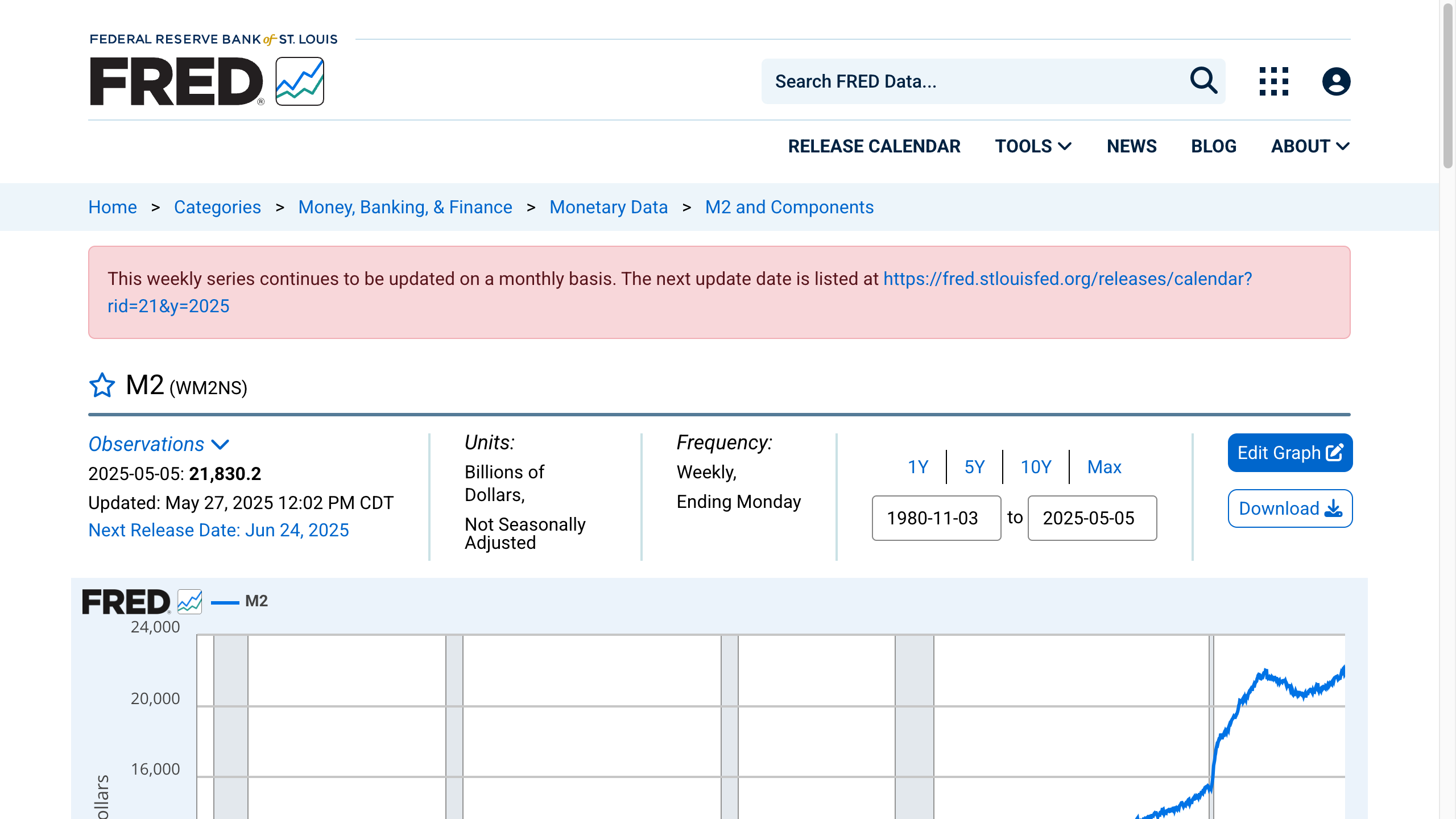}}
        \caption{Step 2: Click the M2 result}
    \end{subfigure}

    \vspace{1em}

    \begin{subfigure}[t]{0.32\textwidth}
        \centering
        \fbox{\includegraphics[width=\linewidth]{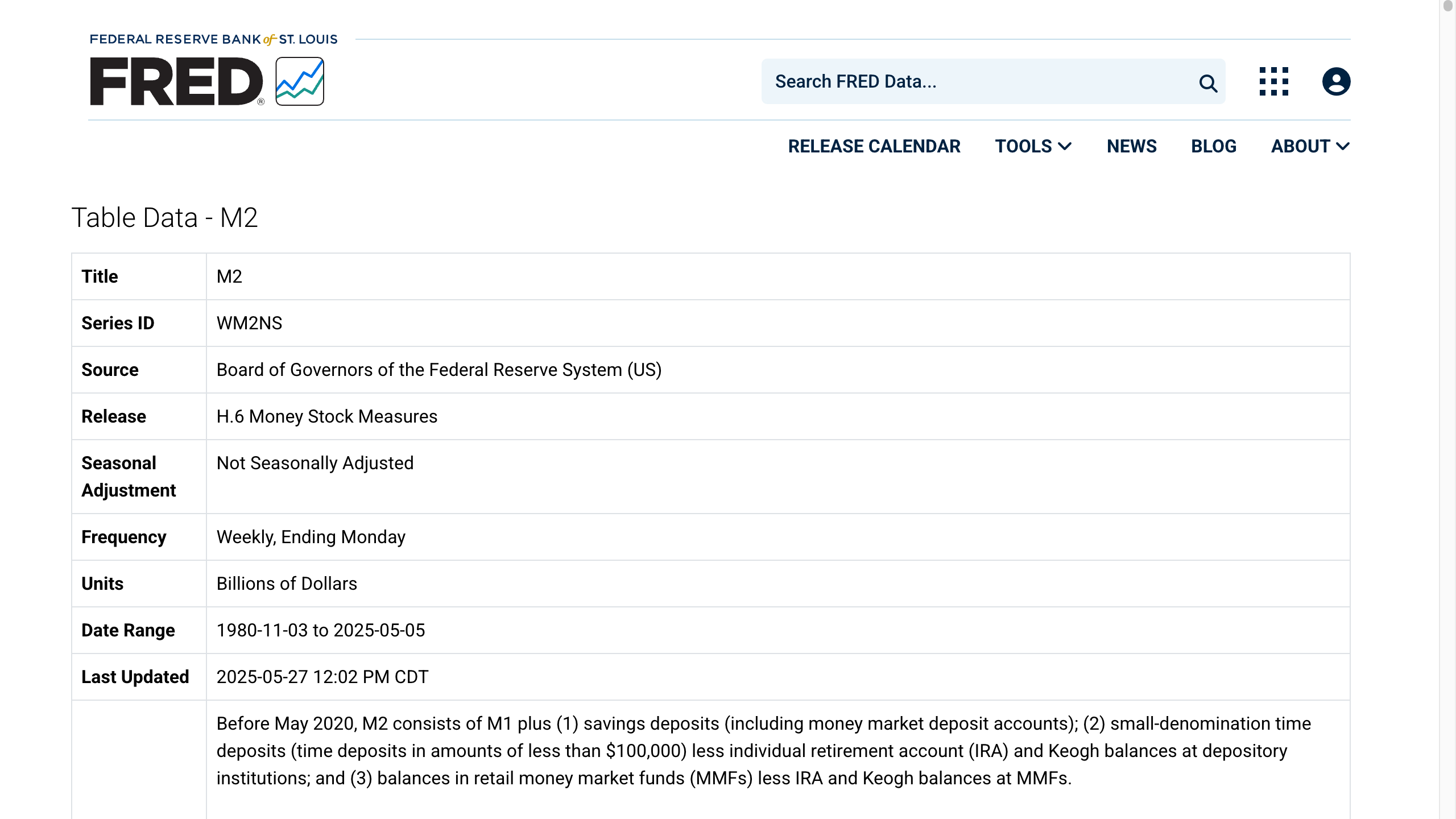}}
        \caption{Step 6: Switch to table view}
    \end{subfigure}
    \hfill
    \begin{subfigure}[t]{0.32\textwidth}
        \centering
        \fbox{\includegraphics[width=\linewidth]{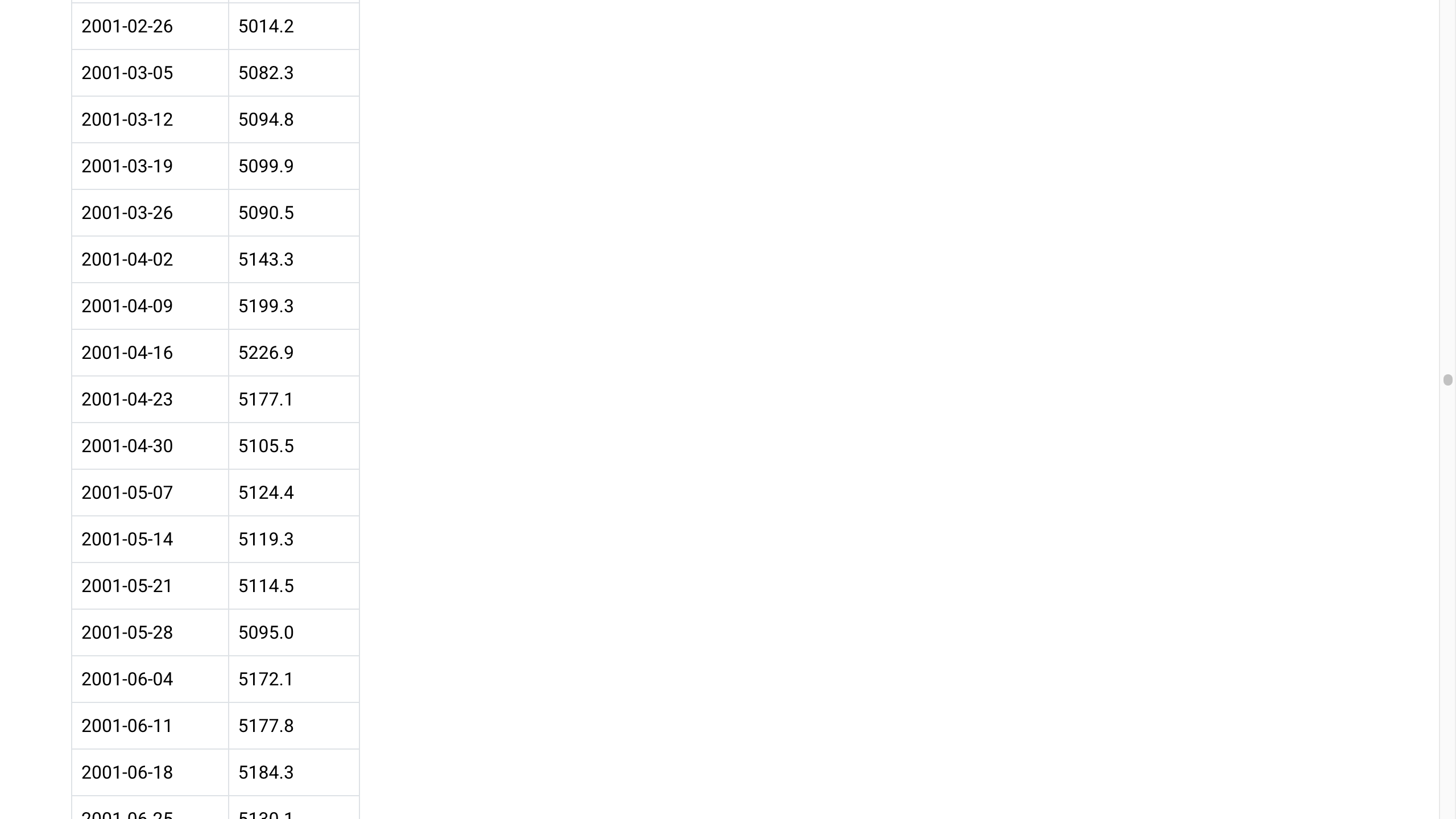}}
        \caption{Step 19: Scroll to locate the date}
    \end{subfigure}
    \hfill
    \begin{subfigure}[t]{0.32\textwidth}
        \centering
        \fbox{\includegraphics[width=\linewidth]{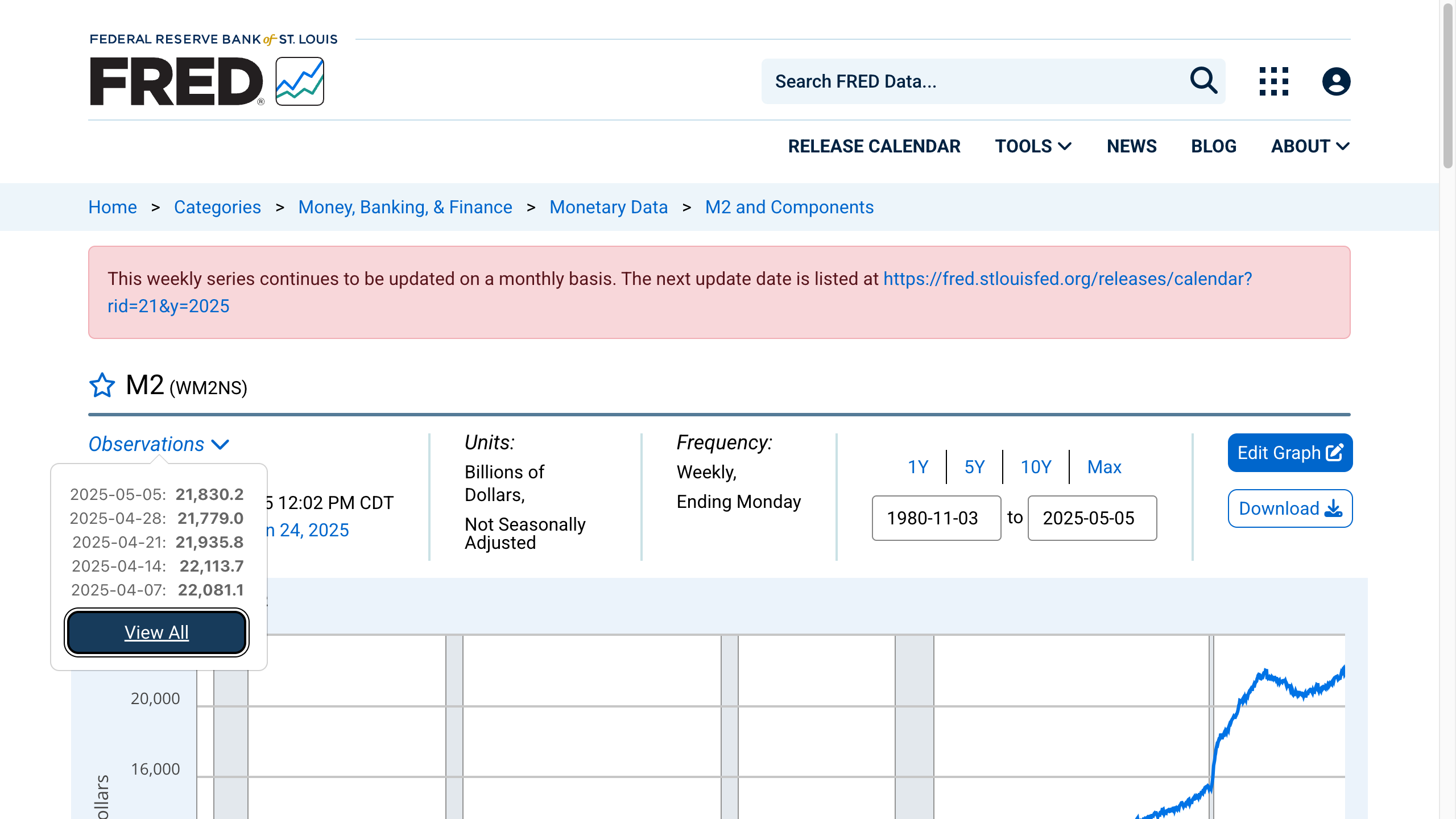}}
        \caption{Step 20: Attempt to filter data}
    \end{subfigure}

    \caption{Example of a data extraction error in Task 88 with o4-mini on retrieving the M2 money supply for February 1, 2025 from FRED. The agent reaches the correct page but fails to extract the target value despite repeated scrolling and filtering.}
    \label{fig:error-88}
\end{figure*}

\begin{figure*}[!h]
    \centering
    \setlength\fboxsep{1pt}
    \setlength\fboxrule{0.5pt}

    \begin{subfigure}[t]{0.32\textwidth}
        \centering
        \fbox{\includegraphics[width=\linewidth]{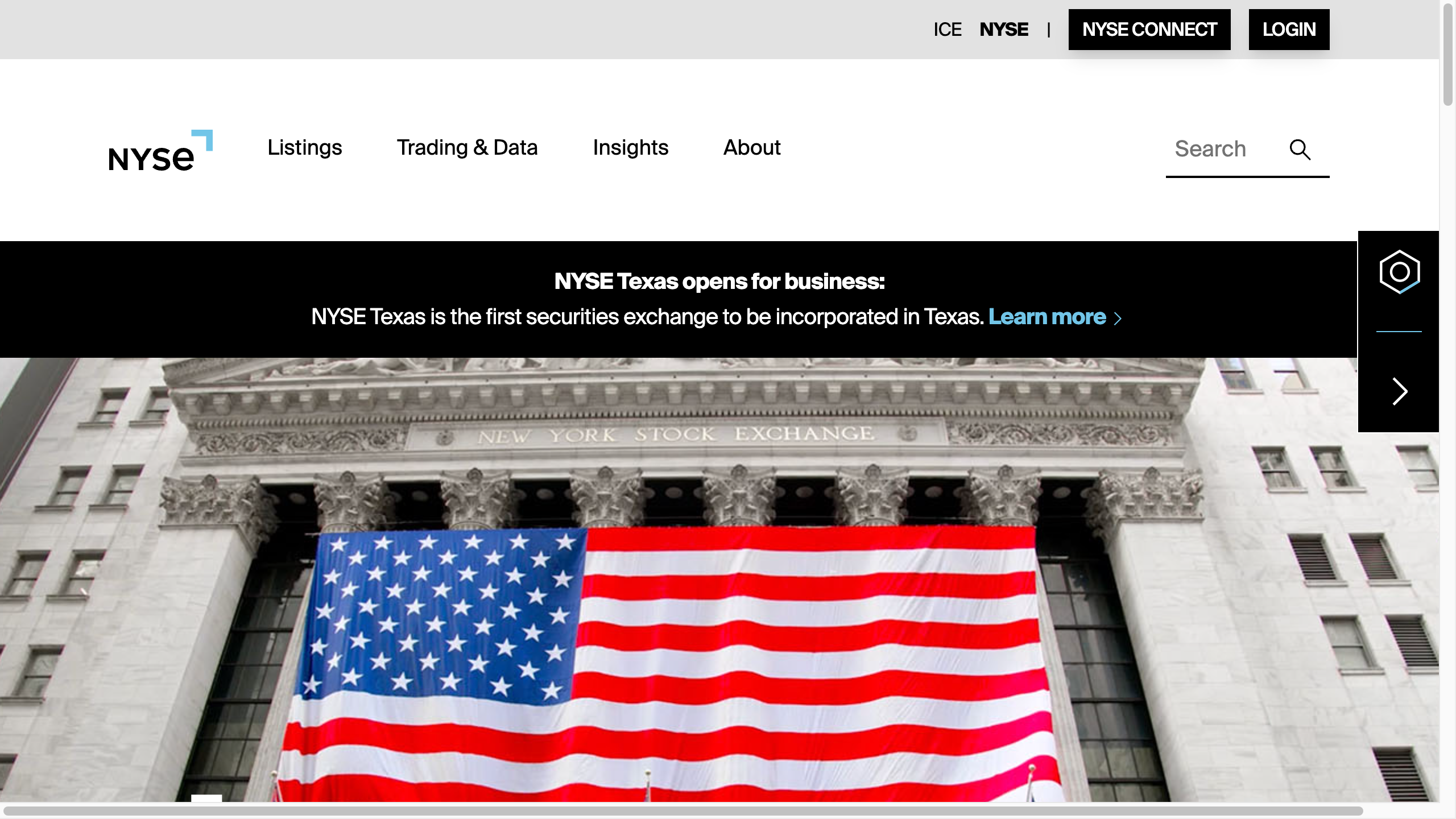}}
        \caption{Step 0: Open NYSE homepage}
    \end{subfigure}
    \hfill
    \begin{subfigure}[t]{0.32\textwidth}
        \centering
        \fbox{\includegraphics[width=\linewidth]{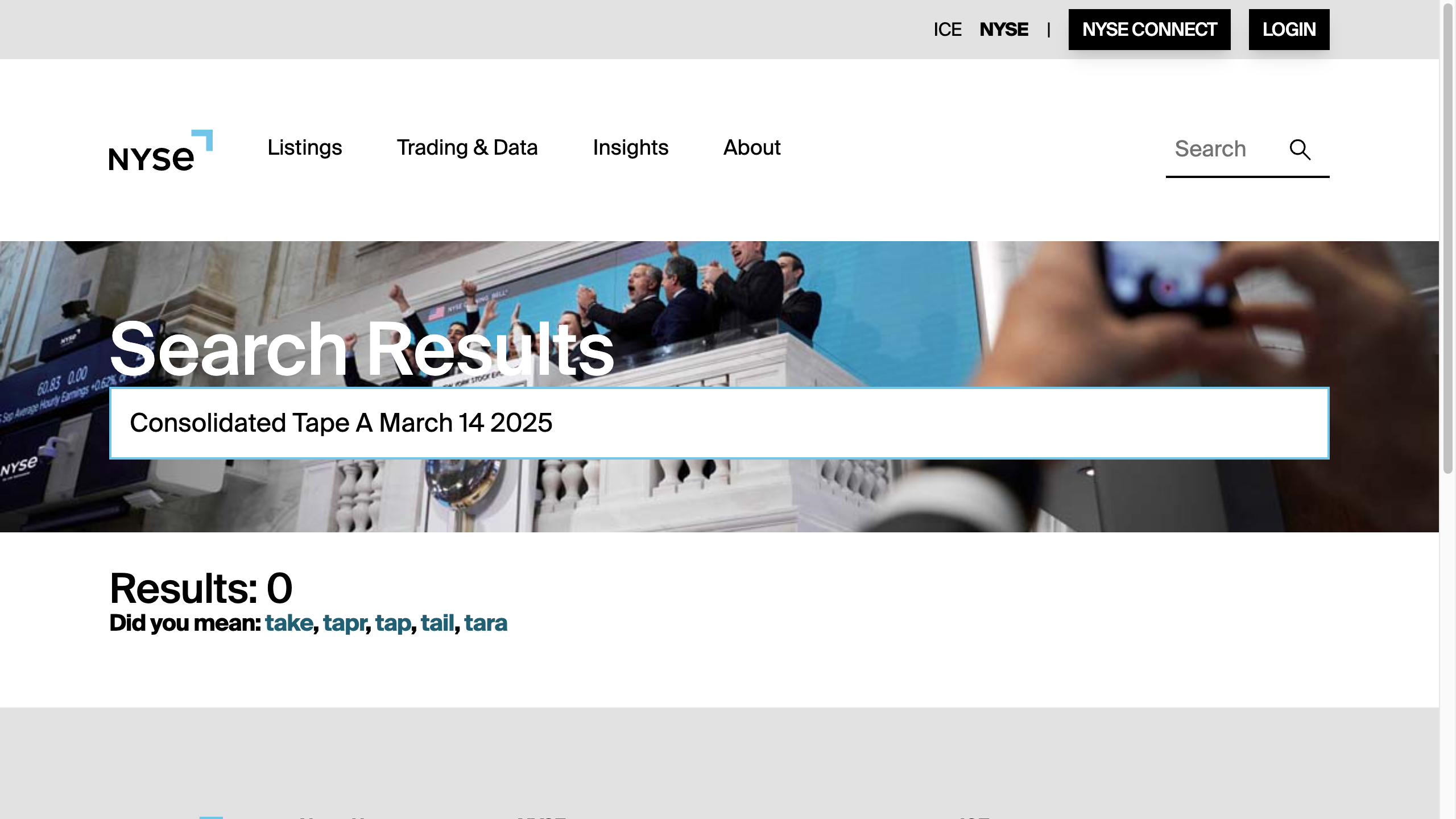}}
        \caption{Step 5: Search for "Consolidated Tap"}
    \end{subfigure}
    \hfill
    \begin{subfigure}[t]{0.32\textwidth}
        \centering
        \fbox{\includegraphics[width=\linewidth]{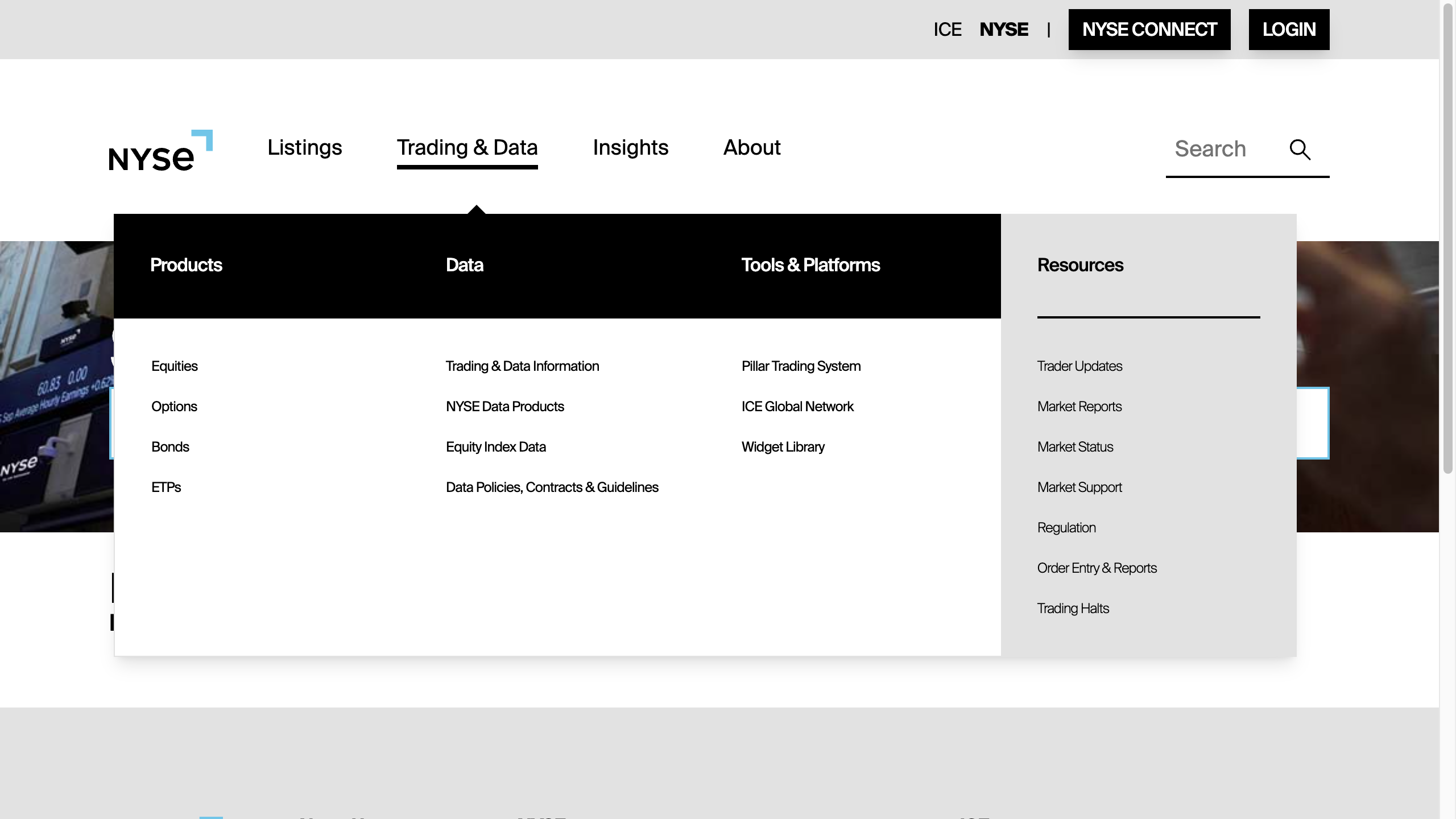}}
        \caption{Step 6: Open trading and data menu}
    \end{subfigure}

    \vspace{1em}

    \begin{subfigure}[t]{0.32\textwidth}
        \centering
        \fbox{\includegraphics[width=\linewidth]{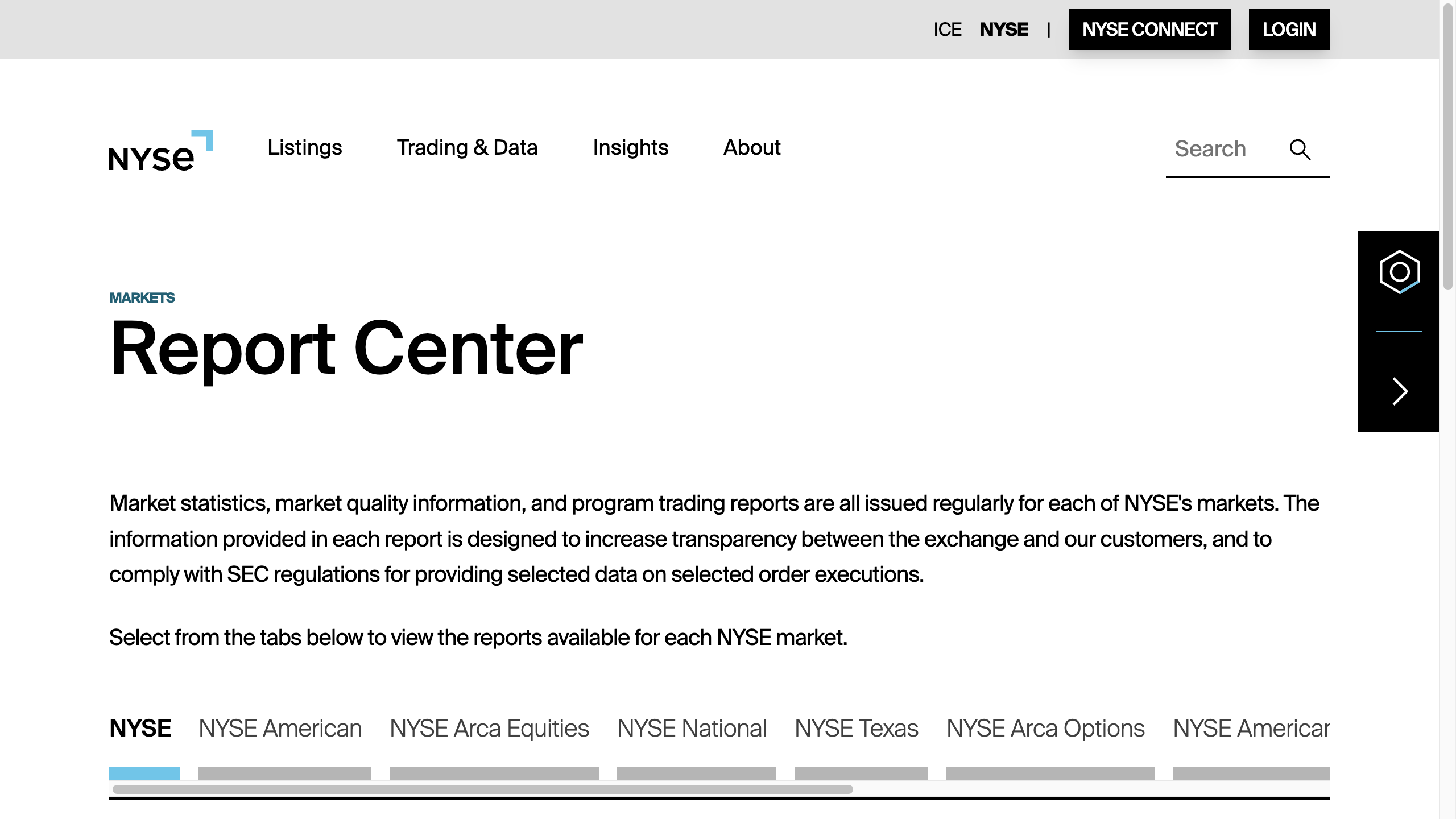}}
        \caption{Step 7: Navigate to market reports}
    \end{subfigure}
    \hfill
    \begin{subfigure}[t]{0.32\textwidth}
        \centering
        \fbox{\includegraphics[width=\linewidth]{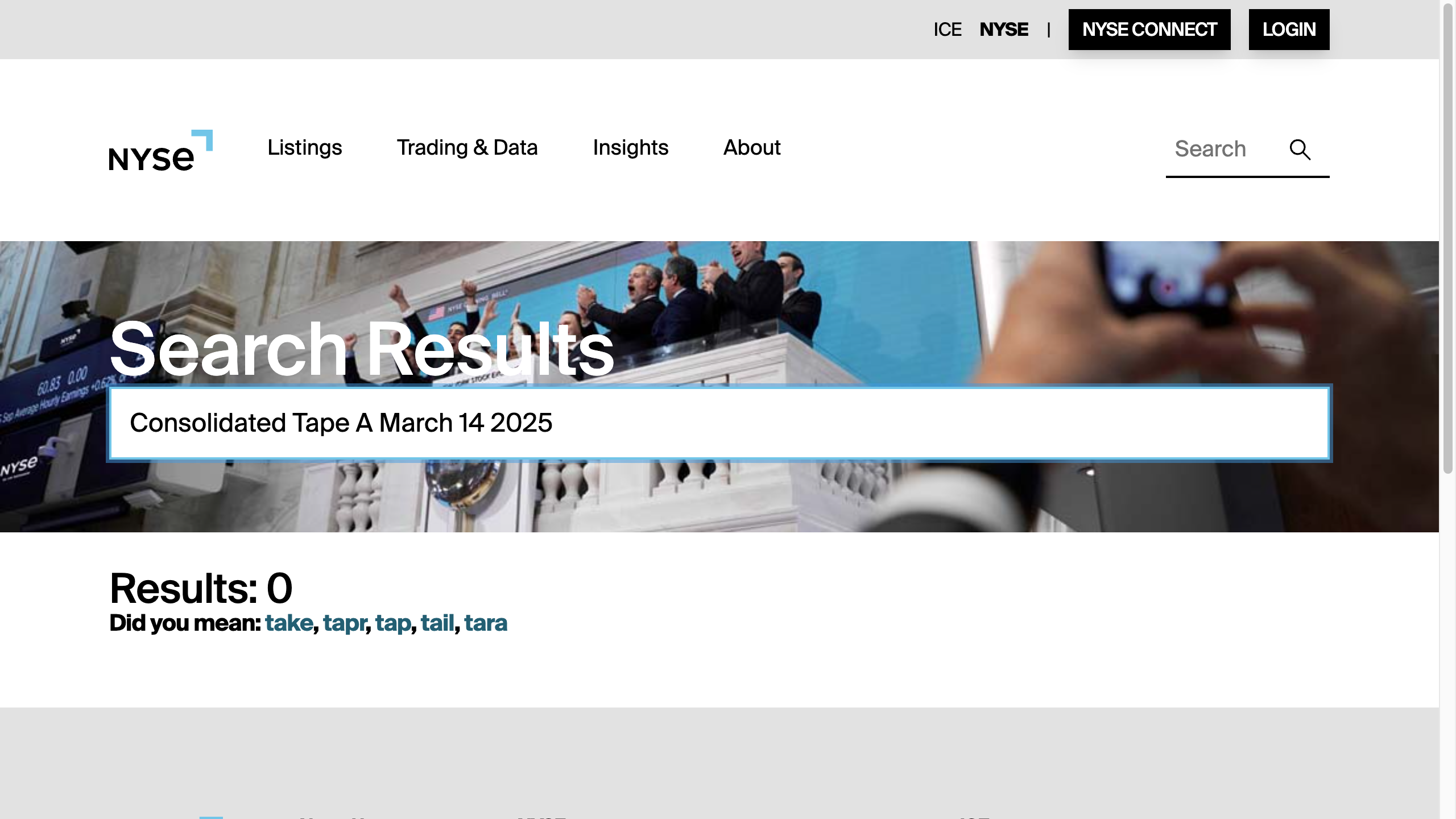}}
        \caption{Step 20: Repeat site search}
    \end{subfigure}
    \hfill
    \begin{subfigure}[t]{0.32\textwidth}
        \centering
        \fbox{\includegraphics[width=\linewidth]{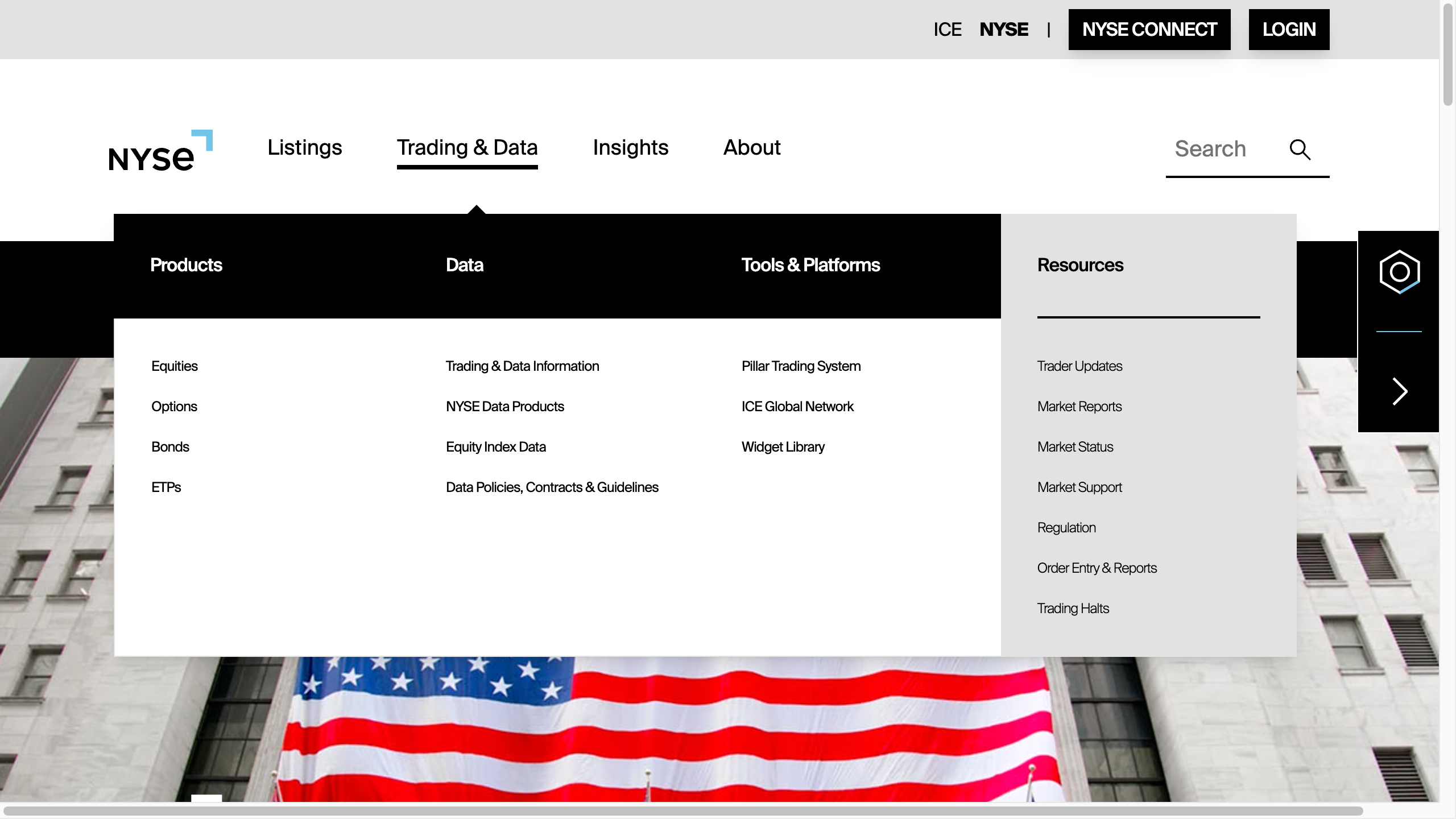}}
        \caption{Step 28: Return to the market reports}
    \end{subfigure}

    \caption{Example of a navigation failure in Task 157 with o4-mini on retrieving Consolidated Tape A trading volume for March 14, 2025 from the NYSE website. The agent repeatedly loops through menus and search pages but fails to reach the report containing the target value.}
    \label{fig:error-157}
\end{figure*}

\begin{figure*}[!h]
    \centering
    \setlength\fboxsep{1pt}
    \setlength\fboxrule{0.5pt}

    \begin{subfigure}[t]{0.32\textwidth}
        \centering
        \fbox{\includegraphics[width=\linewidth]{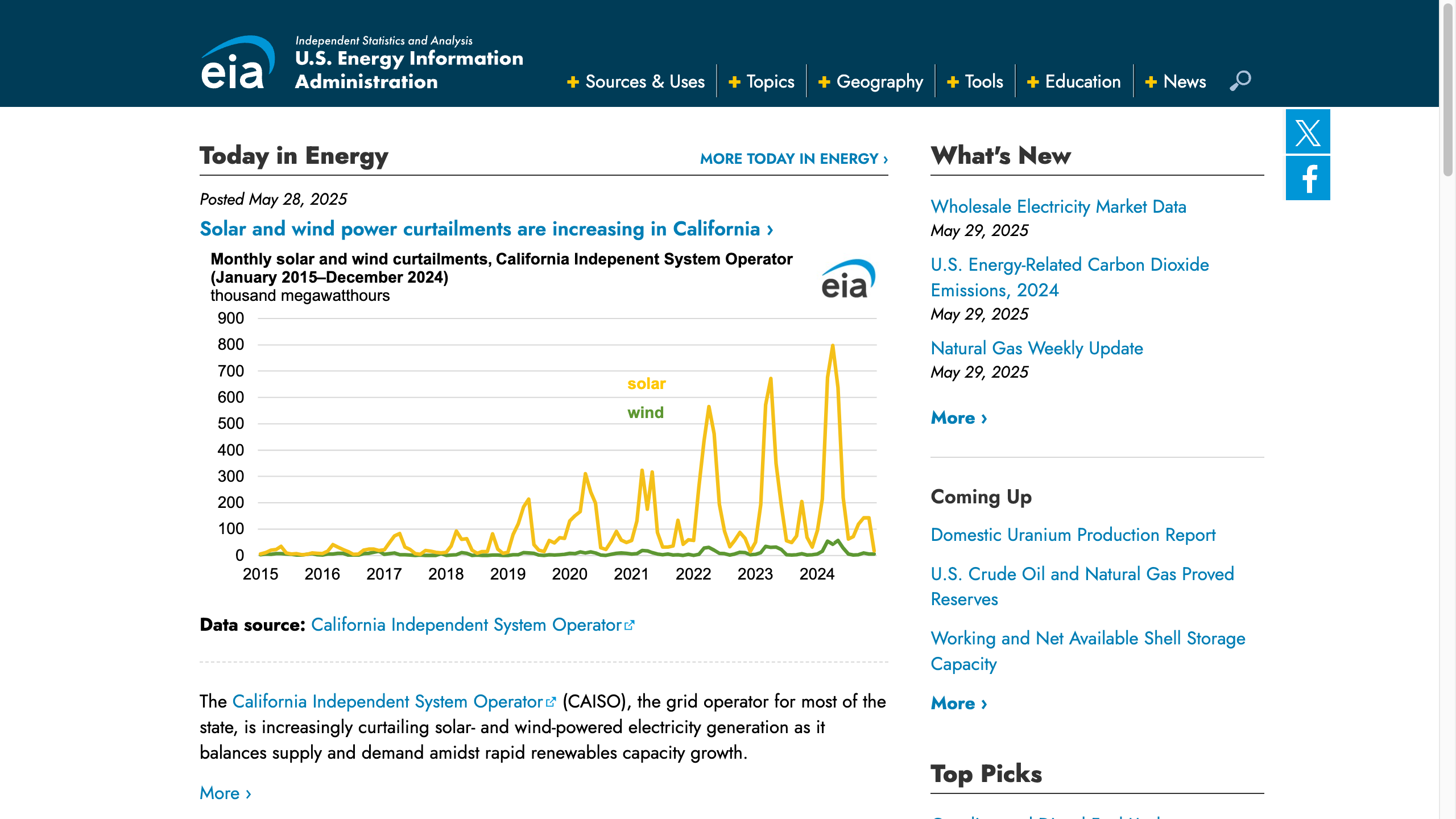}}
        \caption{Step 0: Open EIA homepage}
    \end{subfigure}
    \hfill
    \begin{subfigure}[t]{0.32\textwidth}
        \centering
        \fbox{\includegraphics[width=\linewidth]{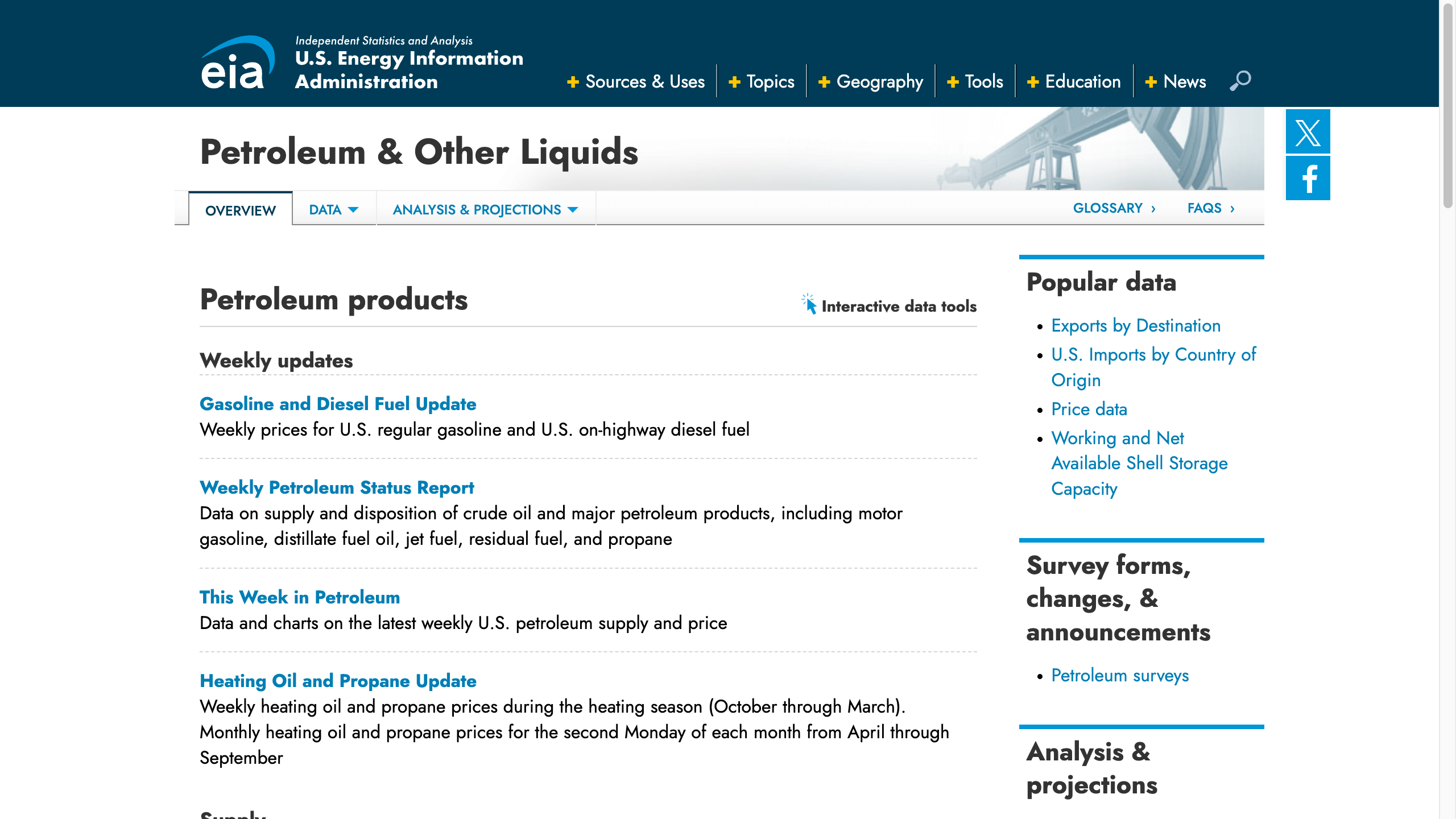}}
        \caption{Step 1: Navigate to price data}
    \end{subfigure}
    \hfill
    \begin{subfigure}[t]{0.32\textwidth}
        \centering
        \fbox{\includegraphics[width=\linewidth]{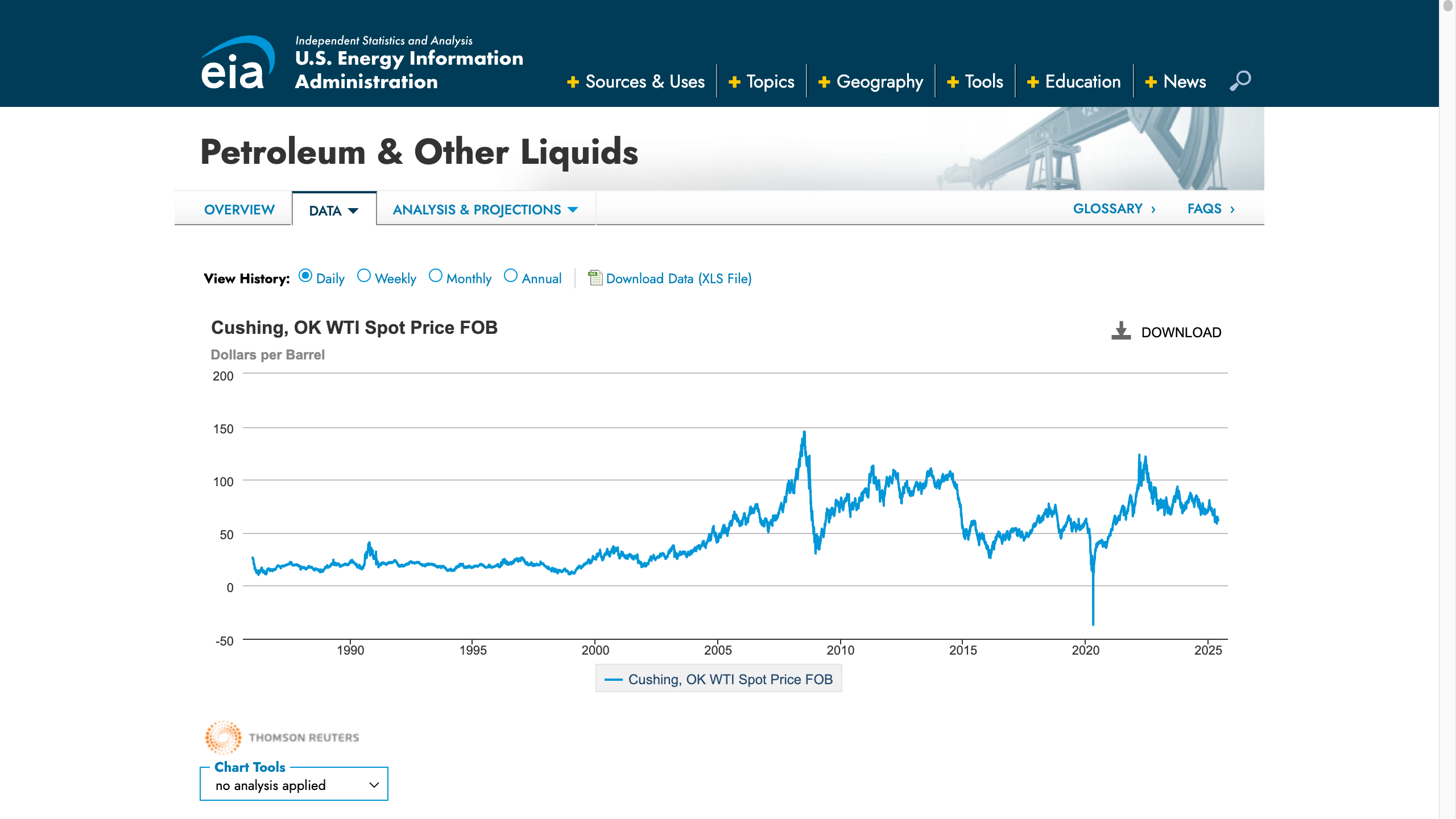}}
        \caption{Step 4: View daily price chart}
    \end{subfigure}

    \vspace{1em}

    \begin{subfigure}[t]{0.32\textwidth}
        \centering
        \fbox{\includegraphics[width=\linewidth]{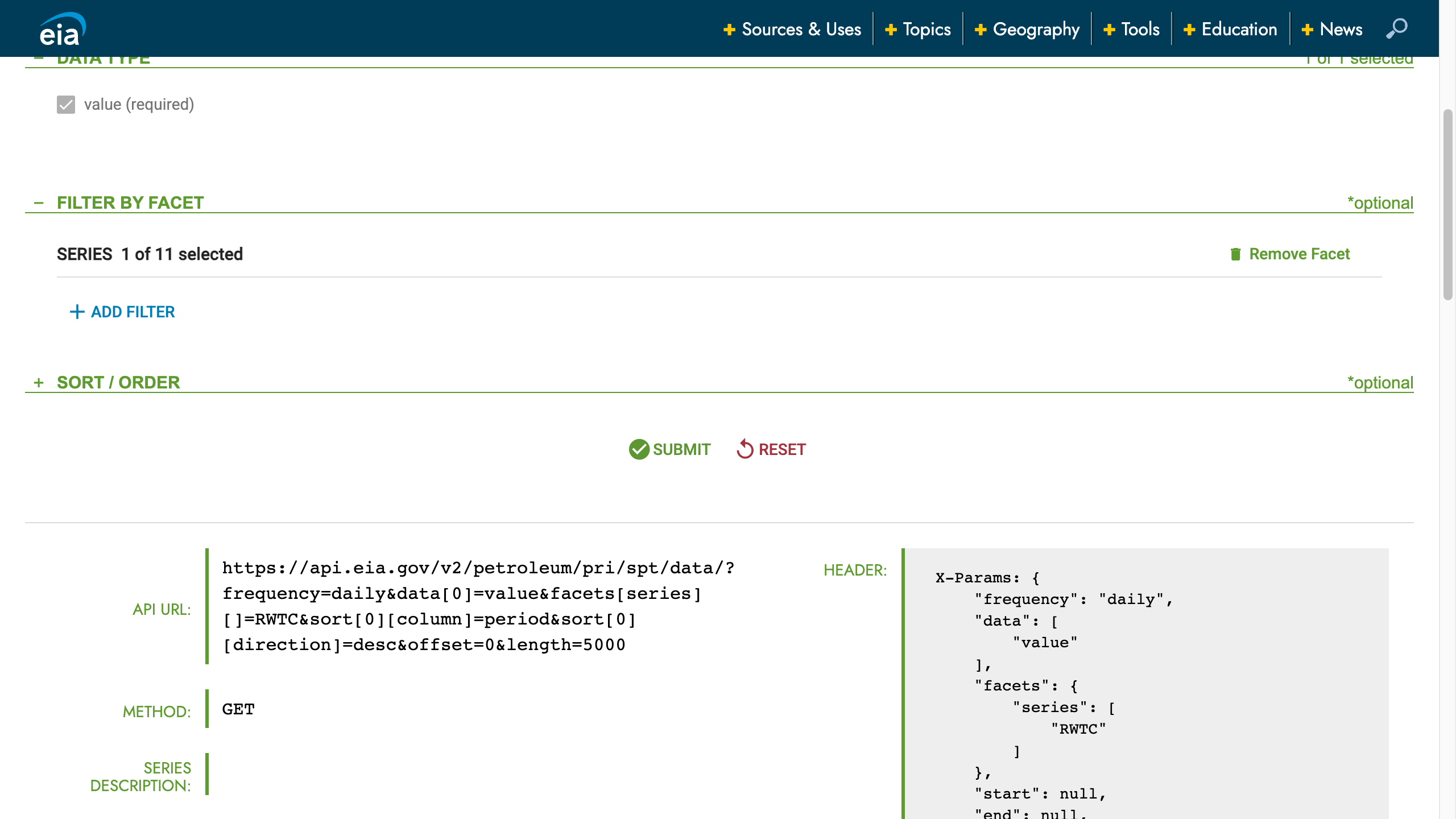}}
        \caption{Step 10: Try API request}
    \end{subfigure}
    \hfill
    \begin{subfigure}[t]{0.32\textwidth}
        \centering
        \fbox{\includegraphics[width=\linewidth]{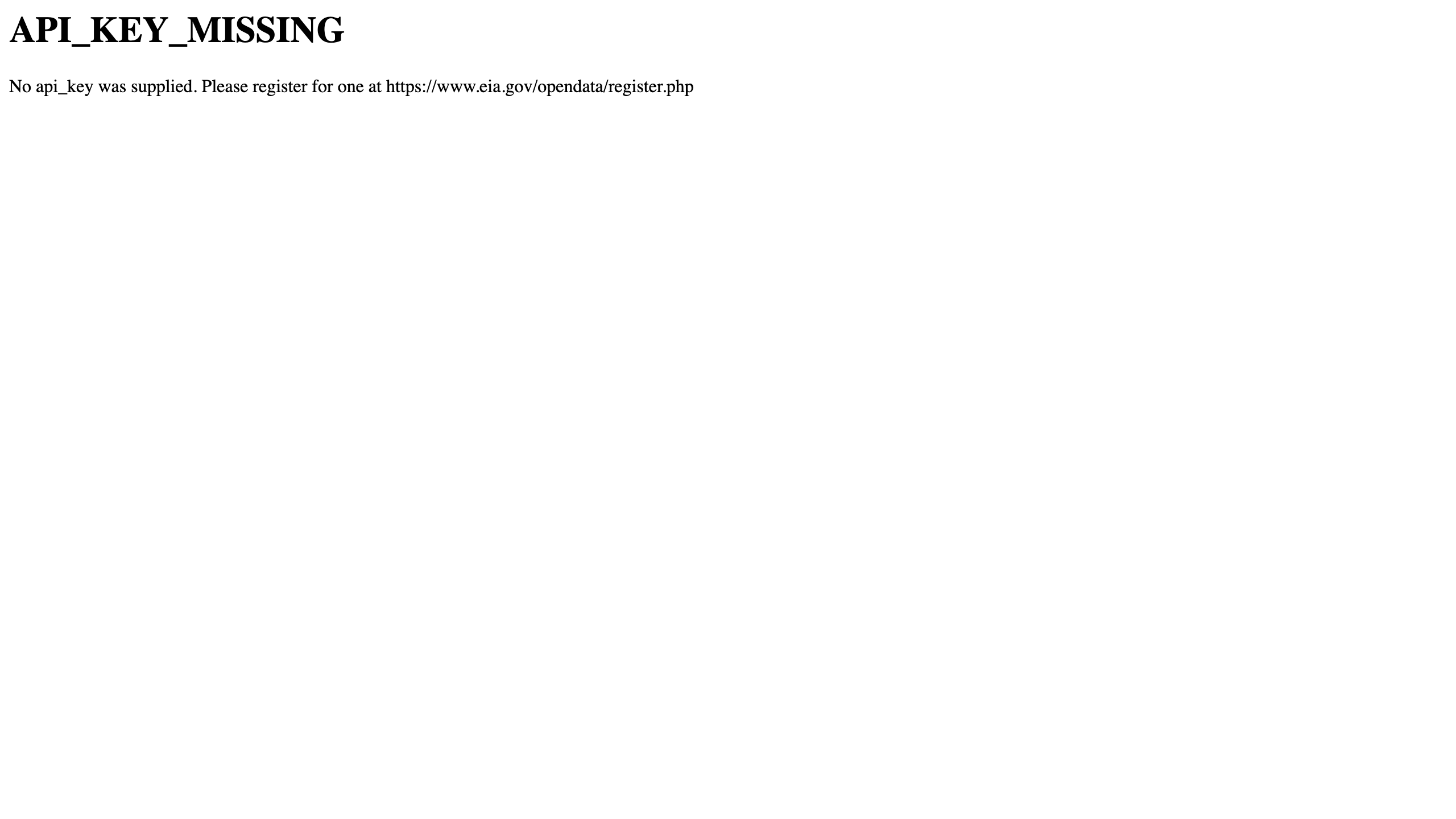}}
        \caption{Step 11: API blocked}
    \end{subfigure}
    \hfill
    \begin{subfigure}[t]{0.32\textwidth}
        \centering
        \fbox{\includegraphics[width=\linewidth]{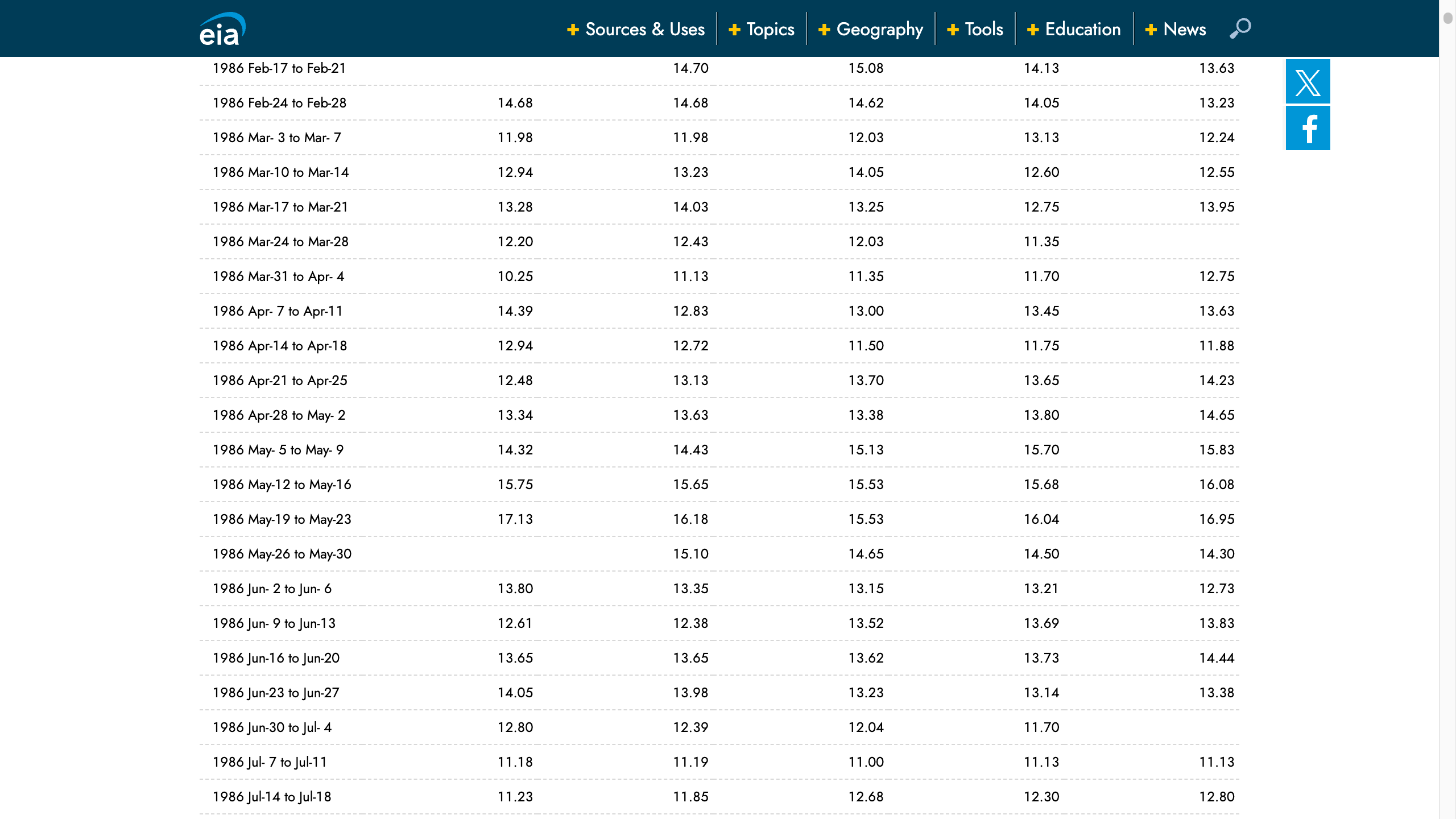}}
        \caption{Step 15: Fallback to HTML table}
    \end{subfigure}

    \caption{Example of a visual understanding failure in Task 283 with o4-mini for retrieving the WTI spot price from the U.S. Energy Information Administration (EIA) on March 10, 2025. The agent encounters multiple layout views, attempts API access, and struggles to correctly extract the value from the diagram or fallback table.}
    \label{fig:error-283}
\end{figure*}

\begin{figure*}[!h]
    \centering
    \setlength\fboxsep{1pt}
    \setlength\fboxrule{0.5pt}

    \begin{subfigure}[t]{0.32\textwidth}
        \centering
        \fbox{\includegraphics[width=\linewidth]{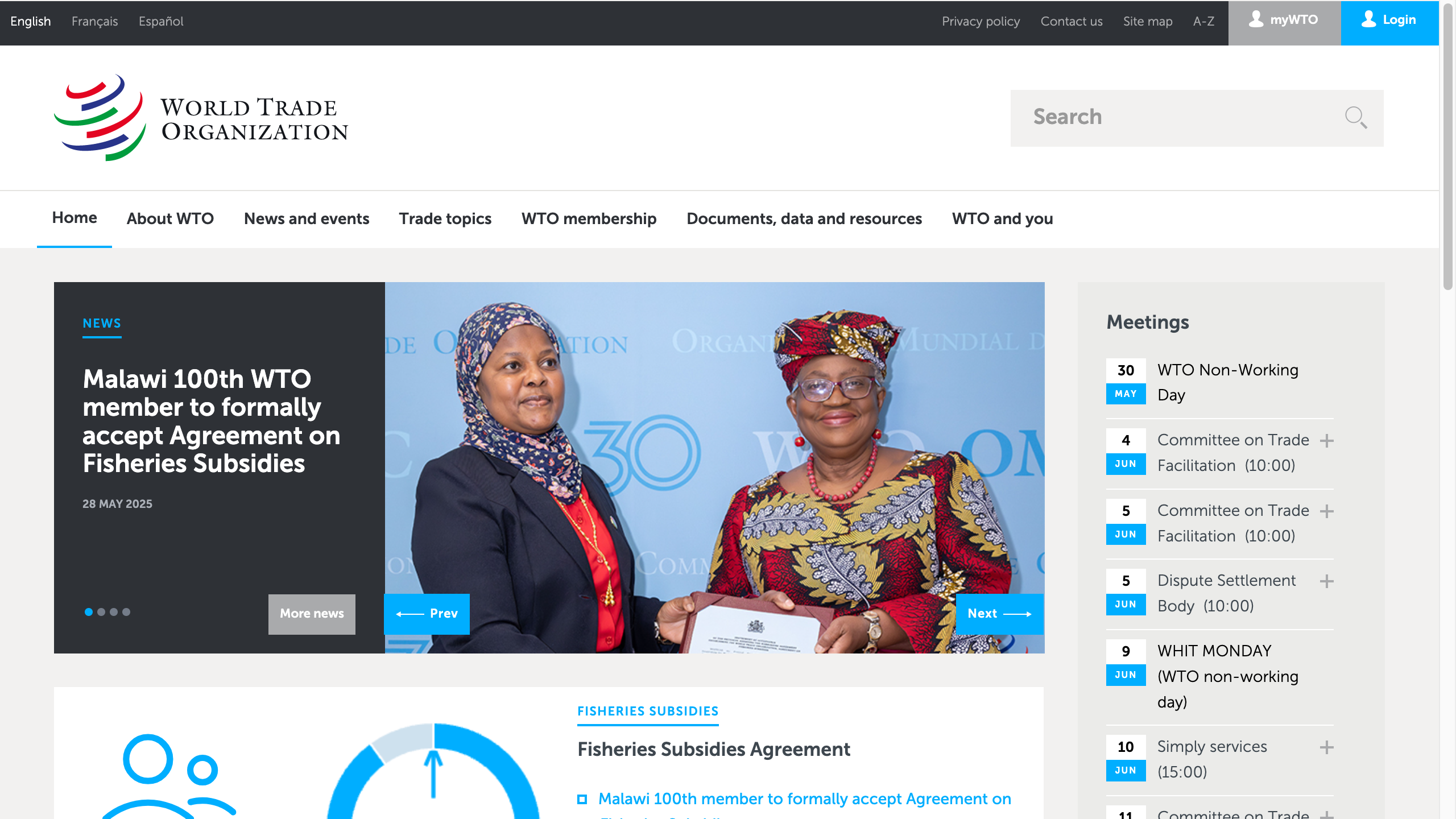}}
        \caption{Step 0: Open WTO homepage}
    \end{subfigure}
    \hfill
    \begin{subfigure}[t]{0.32\textwidth}
        \centering
        \fbox{\includegraphics[width=\linewidth]{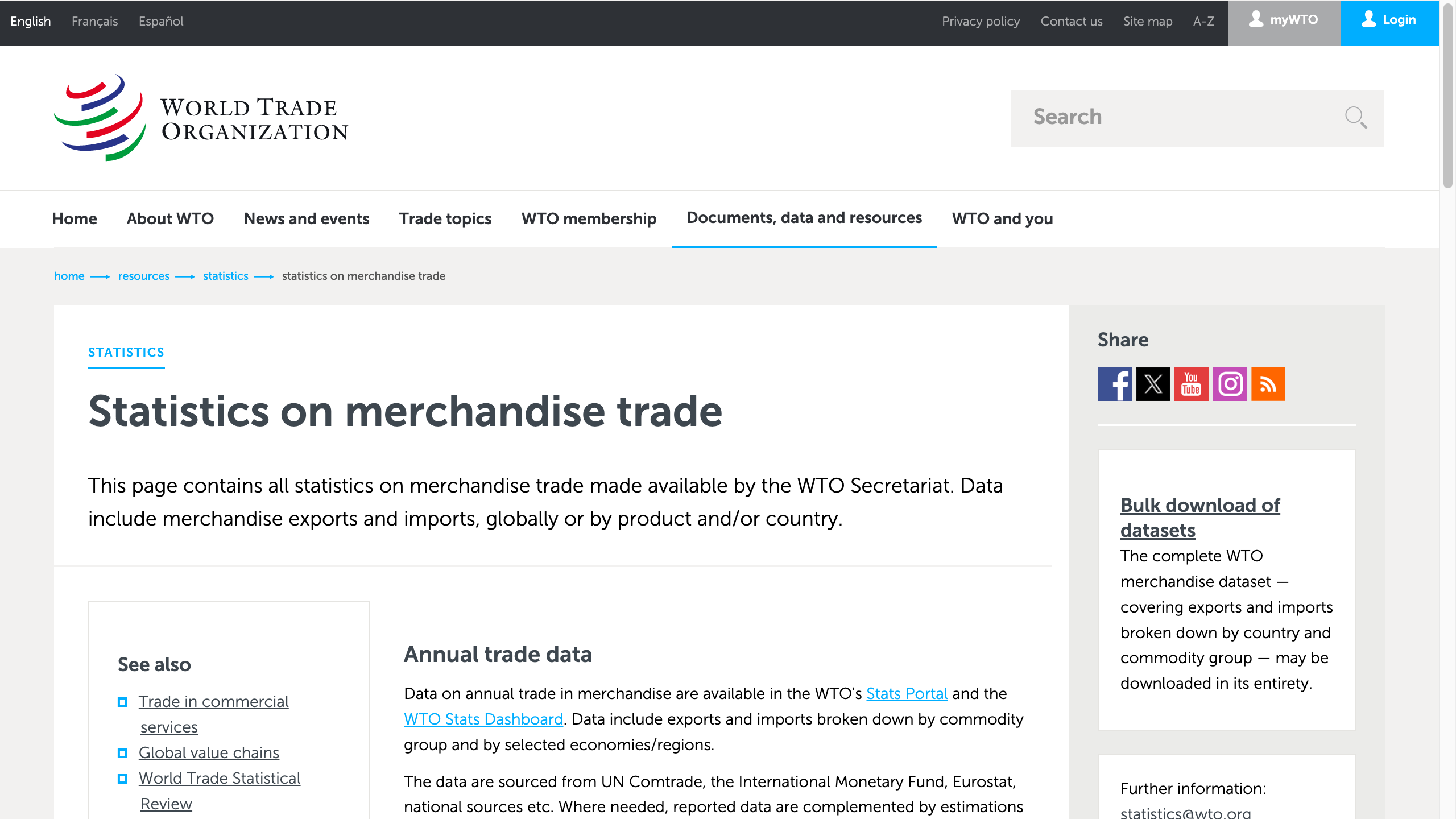}}
        \caption{Step 1: Navigate to statistics}
    \end{subfigure}
    \hfill
    \begin{subfigure}[t]{0.32\textwidth}
        \centering
        \fbox{\includegraphics[width=\linewidth]{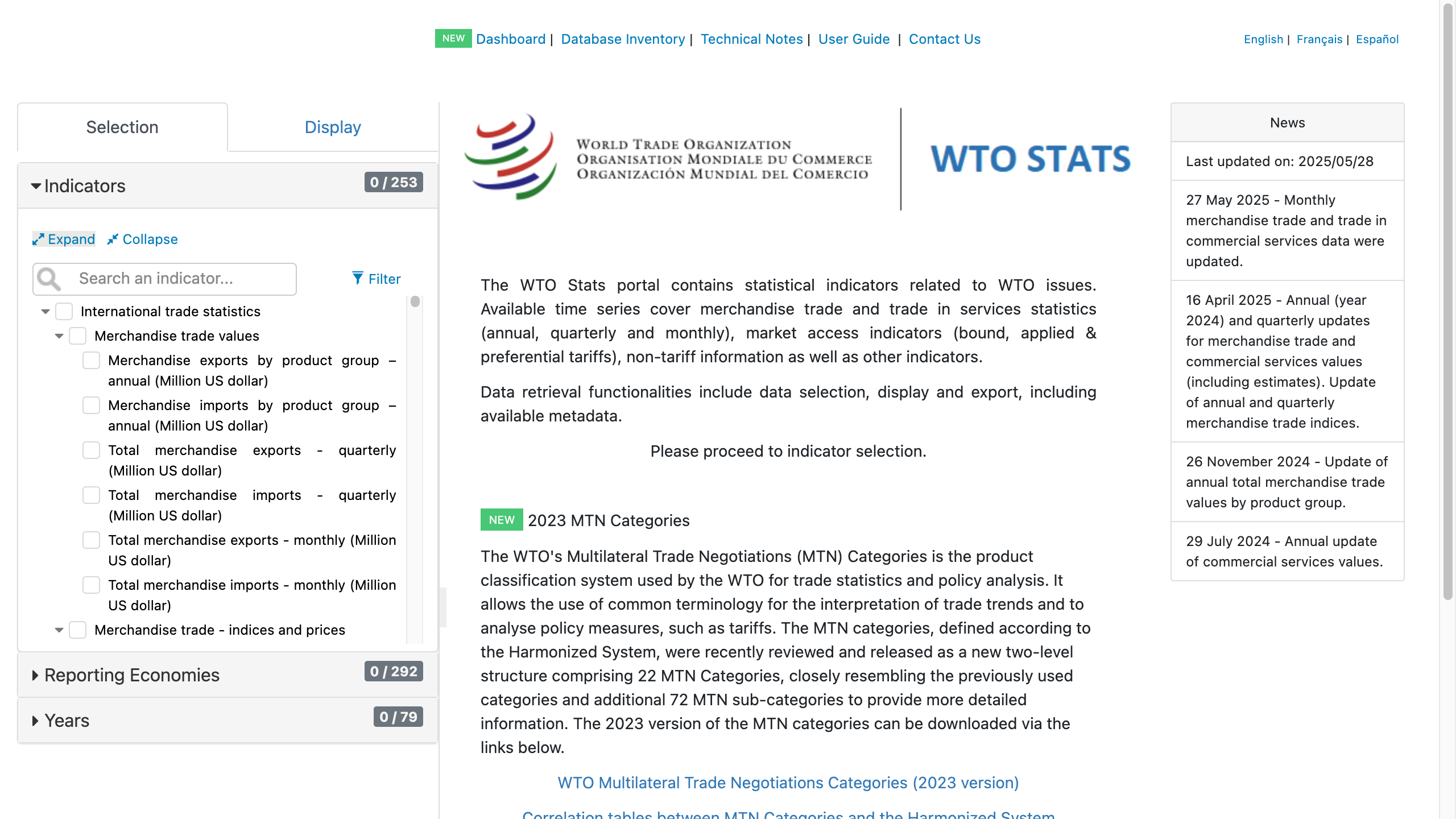}}
        \caption{Step 4: Go to WTO Stats portal}
    \end{subfigure}

    \vspace{1em}

    \begin{subfigure}[t]{0.32\textwidth}
        \centering
        \fbox{\includegraphics[width=\linewidth]{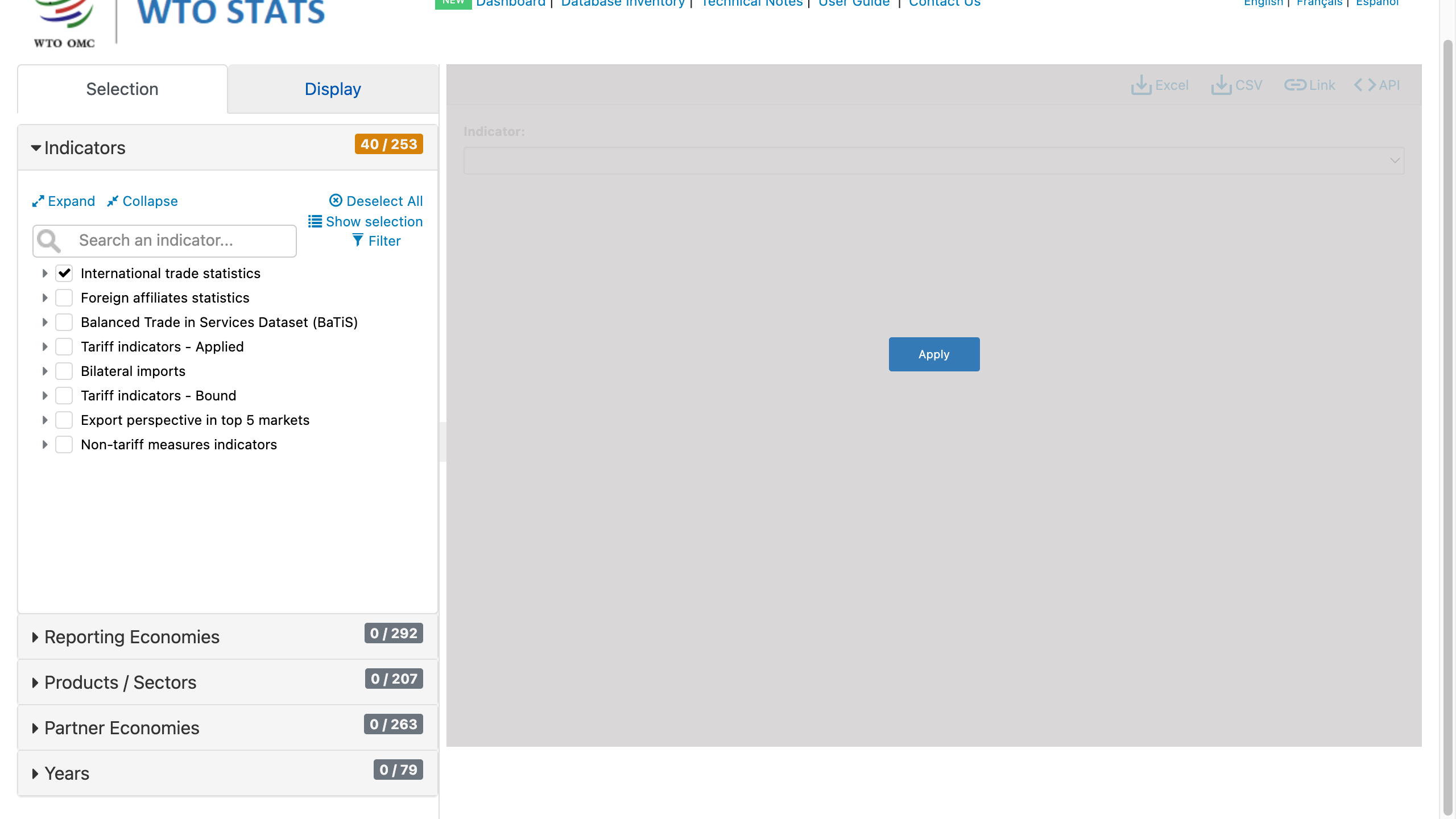}}
        \caption{Step 10: Expand indicators list}
    \end{subfigure}
    \hfill
    \begin{subfigure}[t]{0.32\textwidth}
        \centering
        \fbox{\includegraphics[width=\linewidth]{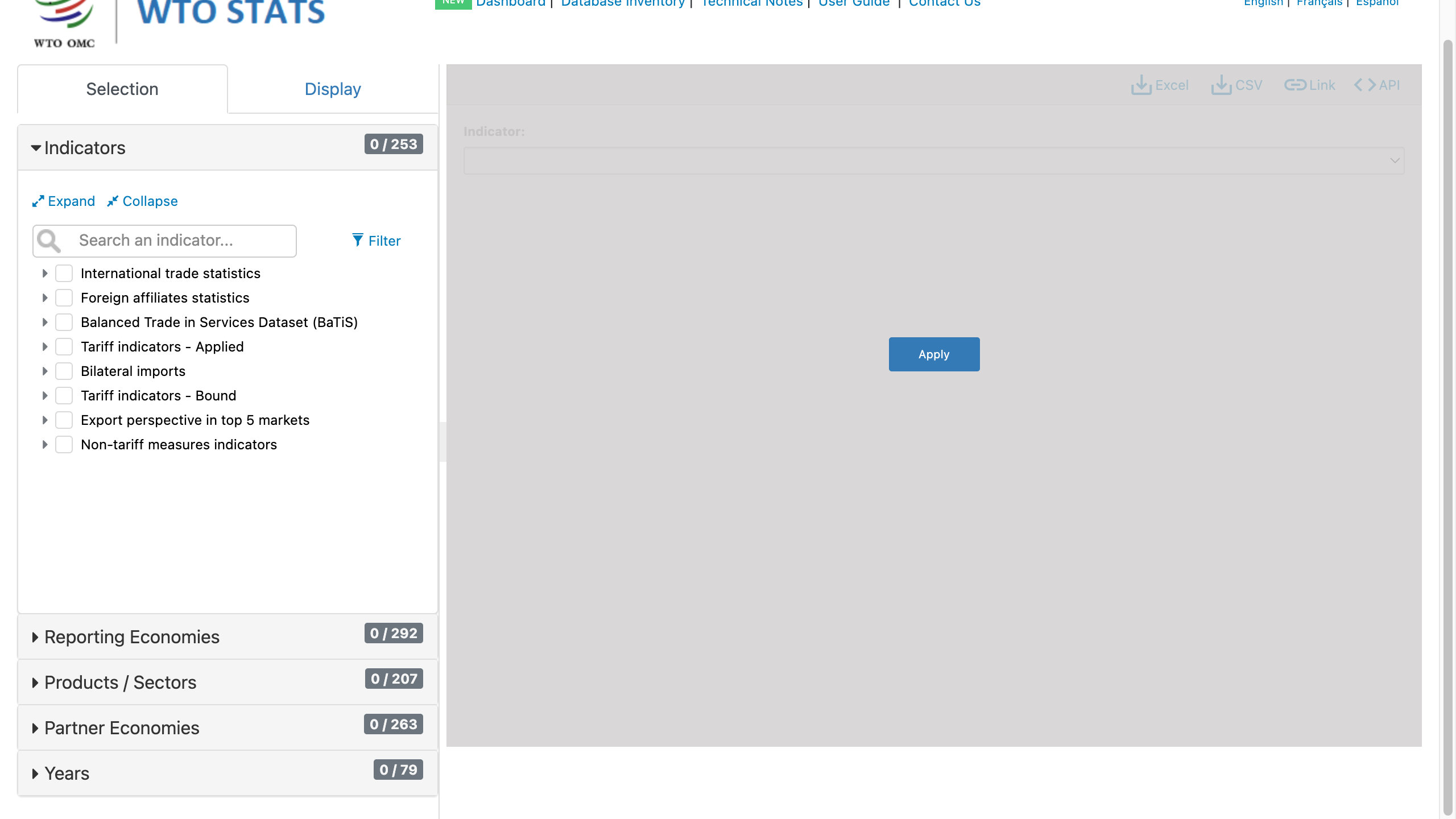}}
        \caption{Step 11: Struggles with selection}
    \end{subfigure}
    \hfill
    \begin{subfigure}[t]{0.32\textwidth}
        \centering
        \fbox{\includegraphics[width=\linewidth]{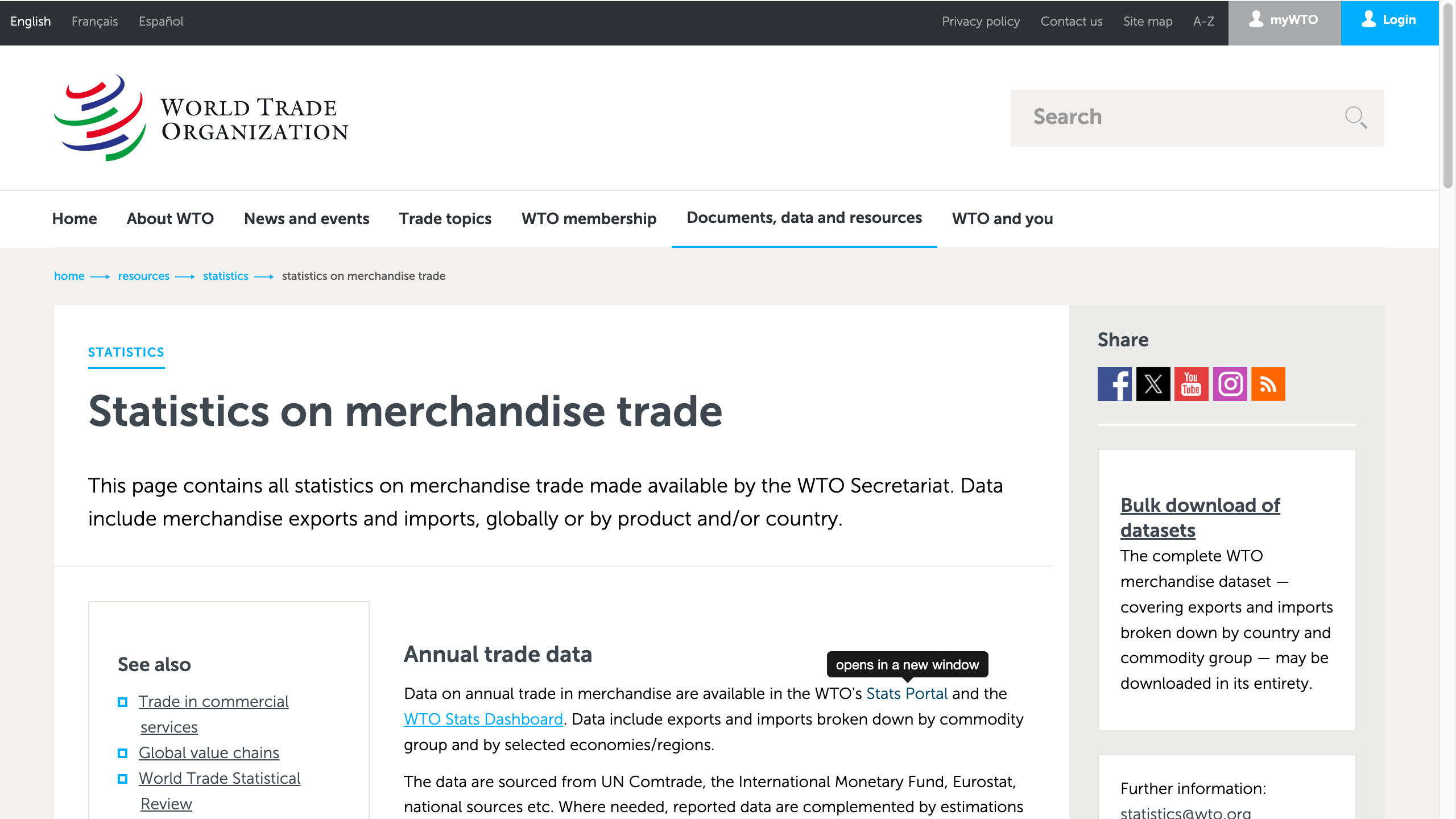}}
        \caption{Step 15: Returns to static page}
    \end{subfigure}

    \caption{Example of an interaction failure in Task 337 with o4-mini. The agent attempts to retrieve Egypt's 2024 merchandise export value from the WTO Stats portal but struggles to operate the interface, repeatedly failing to select the correct indicator due to the dynamic layout and nested checkbox menus.}
    \label{fig:error-337}
\end{figure*}

To supplement our quantitative error analysis, this appendix presents four representative examples that showcase the diverse failure modes observed in o4-mini's behavior. These include data extraction failures due to difficulty parsing large tables (Figure~\ref{fig:error-88}), navigation breakdowns across complex multi-page layouts (Figure~\ref{fig:error-157}), visual understanding issues in interpreting charts and diagrams (Figure~\ref{fig:error-283}), and interaction failures in dynamic UIs with layered filtering and selection logic (Figure~\ref{fig:error-337}). Together, these examples highlight the need for stronger grounding, multimodal reasoning, and interface robustness in economic web agents.

\end{document}